\documentclass{article}
\usepackage{arxiv}

\usepackage[utf8]{inputenc} % allow utf-8 input
\usepackage[T1]{fontenc}    % use 8-bit T1 fonts
\usepackage{hyperref}       % hyperlinks
\usepackage{url}            % simple URL typesetting
\usepackage{booktabs}       % professional-quality tables
\usepackage{amsfonts}       % blackboard math symbols
\usepackage{nicefrac}       % compact symbols for 1/2, etc.
\usepackage{microtype}      % microtypography

\usepackage{graphicx}

\usepackage{esvect}
\usepackage{amsmath}
\usepackage[tight,footnotesize]{subfigure}
\usepackage{appendix}
\usepackage{lscape}
\usepackage{textcomp}
\usepackage{amssymb}
\usepackage{lscape}
\usepackage[noend]{algpseudocode}
\usepackage[boxruled]{algorithm2e}
\usepackage{hhline}
\usepackage{slashbox}
\usepackage{multirow}
\usepackage{lettrine}
\usepackage{tabularx}

\title{IT2CFNN: An Interval Type-2 Correlation-Aware Fuzzy Neural Network to Construct Non-Separable Fuzzy Rules with Uncertain and Adaptive Shapes for Nonlinear Function Approximation}

\author{
  Armin Salimi-Badr \\
  Faculty of Computer Science and Engineering\\
  Shahid Beheshti University\\
  Tehran, Iran \\
  \texttt{a\_salimibadr@sbu.ac.ir} \\
}

\begin{document}
\maketitle

\begin{abstract}
In this paper, a new interval type-2 fuzzy neural network able to construct non-separable fuzzy rules with adaptive shapes is introduced. To reflect the uncertainty, the shape of fuzzy sets considered to be uncertain. Therefore, a new form of interval type-2 fuzzy sets based on a general Gaussian model able to construct different shapes (including triangular, bell-shaped, trapezoidal) is proposed. To consider the interactions among input variables, input vectors are transformed to new feature spaces with uncorrelated variables proper for defining each fuzzy rule. Next, the new features are fed to a fuzzification layer using proposed interval type-2 fuzzy sets with adaptive shape. Consequently, interval type-2 non-separable fuzzy rules with proper shapes, considering the local interactions of variables and the uncertainty are formed. For type reduction the contribution of the upper and lower firing strengths of each fuzzy rule are adaptively selected separately. To train different parameters of the network, the Levenberg-Marquadt optimization method is utilized. The performance of the proposed method is investigated on clean and noisy datasets to show the ability to consider the uncertainty. Moreover, the proposed paradigm, is successfully applied to real-world time-series predictions, regression problems, and nonlinear system identification. According to the experimental results, the performance of our proposed model outperforms other methods with a more parsimonious structure.\\

\\
The full version of this preprint is accepted for publication in Applied Soft Computing as: \\
\textit{A. Salimi-Badr, "IT2CFNN: An interval type-2 correlation-aware fuzzy neural network to construct non-separable fuzzy rules with uncertain and adaptive shapes for nonlinear function approximation," Applied Soft Computing, vol. 115, pp. 108258, 2022.
DOI : 10.1016/j.asoc.2021.108258}
\end{abstract}

%% main text
\section{Introduction}

Fuzzy Inference Systems (FISs) are rule based systems including a set of "IF-THEN" linguistic rules. Representing such an interpretable system in the form of an adaptive neural network constructs Fuzzy Neural Networks (FNNs) \cite{LUO2019,de2020fuzzy,Ebadzadeh2017,Salimi2020novel,Ebadzadeh15,Salimi-Badr2017,salimi2020backpropagation,ANFIS,SOFMLS,Das15,BAKLOUTI2018,ashrafi2020it2}. It is proved that FNNs are universal approximators \cite{Kosko94,Ying98,Zeng00} and their abilities including function approxiamtion, nonlinear system identification, time-series prediction, control, and classification in different applications like data mining, signal processing, system modeling, and robotics are successfully investigated \cite{KUO2021107711,VAN2021107186,WANG2021107471,MA2017296,Ebadzadeh08,Han14,Kim15,Khodabandelou2019,Chang2017novel,Han2019knowledge,AsadiEydivand2015,Salimi-Badr2017,HAN2021188,DECAMPOSSOUZA2021231}.

Although FNNs are able to learn fuzzy rules based on linguistic variables, they are unable to efficiently handle the uncertainty in data due to using type-1 fuzzy sets and considering membership degrees as precise values \cite{karnik1999type,liang2000interval}. To overcome this issue, Zadeh proposed the extension of fuzzy sets with fuzzy membership values (secondary membership values) known as type-2 fuzzy sets \cite{Zadeh75}. In comparison to an ordinary FIS that is constructed from fuzzifier, inference engine, and defuzzifier, a Type-2 FIS has an extra computing unit to reduce the type of fuzzy rules by extracting the embedded type-1 fuzzy rule before defuzzification (type-reduction) \cite{mendel2007}. Generally this type-reduction operation incurs additional computational cost \cite{Das15}. To decrease this computation cost, Liang and Mendel proposed interval type-2 fuzzy set theory by defining two bounds for the secondary membership function and considering it as a constant value between these bounds \cite{liang2000interval}. The FIS built upon these interval type-2 fuzzy sets is called Interval Type-2 FIS (IT2FIS) and its neural interpretation is called Interval Type-2 FNN (IT2FNN). Moreover, Karnik and Mendel proposed an iterative algorithm for type reduction known as Karnik-Mendel (KM) algorithm \cite{KM}.

Since proposing IT2FIS, various structures for realizing IT2FNNs have been presented. In \cite{CASTRO20092175} three IT2FNNs have been proposed. These architectures are based on the Takagi-Sugeno-Kang (TSK) FIS model (\cite{Takagi85,sugeno1988structure}) and both antecedent and consequent parts of fuzzy rules are defined based on type-2 fuzzy sets. They differ in the approach of their parameter learning. Moreover, in two proposed structures, KM algorithm is used as the type reduction method, while in the last one an adaptive type reduction method is proposed to adaptively average the upper and lower firing strength of fuzzy rules. This averaging method is more efficient than KM algorithm and similar adaptive weightening type reduction methods are proposed in other related studies for Mamdani and TSK IT2FNN \cite{kaynak2010,Lin2014,Das15,pratma2,Eyoh2018}.

There are two usual ways to define interval type-2 fuzzy sets: 1- uncertain mean, and 2- uncertain width. The first approach considers that the mean value of a fuzzy set is uncertain, while the second one assumes that the width of the fuzzy set is uncertain \cite{mendel2007}. Both approaches are utilized in previous studies \cite{rl2009,Das15,pratma2,Hantype22018,LUO2019}.

Different approaches are applied to construct IT2FNN and learn its different parameters. Using clustering methods like FCM was a popular approach to initialize type-1 FNNs \cite{Malek11,Ebadzadeh15,Ebadzadeh2017}. Different clustering methods to indicate the parameters of the IT2FIS like Interval Type-2 Fuzzy C-Means (IT2FCM) are investigated in the literature \cite{FAZELZARANDI2013346}. In \cite{REN2014121} subtractive clustering is used to initialize different parameters of fuzzy sets.

To handle varying data, networks with evolving structures are proposed based on online clustering \cite{SEIT2FNN,IT2FNN-SVR,Juang2015,Das15}. In \cite{SEIT2FNN}, to address uncertainty contained in data streams, expert knowledge, and noisy samples, an evolving type-2 fuzzy system (\textit{SEIT2FNN}) is presented. This method was able to online evolve its structure facing stream of data. It applies a Kalman filter based method for learning consequent parts' parameters, and an incremental gradient descent algorithm for antecedent parts' parameter training. Later, other approaches for evolving type-2 fuzzy neural networks are proposed like \textit{eT2FIS} \cite{eT2FIS}, McIT2FIS \cite{Das15}, \textit{eT2RFNN} \cite{pratma2}, and \textit{eRIT2IFNN} \cite{LUO2019} which expand the original approach in \cite{SEIT2FNN} based on novel concepts like recurrent units, applying Meta-cognitive learning, active learning for sample selection, dimension reduction, and handing cyclic drifts \cite{SKRJANC2019344}.

Some previous studies used Support Vector Machine (SVM) algorithm for parameter learning \cite{IT2FNN-SVR,Juang2015}. Another approach for learning network's parameters is to apply meta-heuristic optimization methods like Ant Colony Optimization \cite{rl2009} or Differential Evolution (DE) method \cite{ALIEV20111591}. To learn antecedent parts parameters, Gradient Descent and Backpropagation are used in some previous studies \cite{CASTRO20092175,Juangbp}.

In function approximation application (including subproblems like time-series prediction, regression problems, and nonlinear system identification), we can consider the surface of a function (function landscape) similar to a mountain landscape with many hills. In \cite{Ebadzadeh2017} it is proposed that to have efficient intuitive and interpretable fuzzy rules for function approximation, the fuzzy rules should be similar to the covered surface. Consequently the fuzzy rules should model the hills of the function landscape.

In \cite{Ebadzadeh2017} it is proposed that to realize the similarity between the fuzzy rules and the covered regions of the function, along with the fuzzy rules' centers and widthes, the local interactions among different variables and also the shapes of fuzzy sets should be determined properly. The local interactions in a region cause the rotation of the fuzzy rules and forming their correlated contours, like rotated ellipsoidal regions. However, the shapes of different hills in the function landscape are not always ellipsoidal. To have fuzzy rules with various shapes, it is necessary to have fuzzy sets with different shapes. In \cite{Ebadzadeh2017} a general form of Gaussian membership function is used that can form fuzzy sets with different and adaptive shapes (triangular, bell-shaped, and trapezoidal). Based on using the extension principle, the final shape of the fuzzy rules affected by the shape of involved fuzzy sets defined along each dimension. Therefore, by determining the shapes of different fuzzy sets involved in constructing different fuzzy rules, it is possible to form fuzzy rules with various shapes similar to the covered region. Defining such fuzzy rules can improve the accuracy of the method and also can make the final structure more parsimonious \cite{Ebadzadeh15,Ebadzadeh2017}.

Neither the interactions among variables, nor the adaptive shapes of fuzzy sets are considered by the most previous paradigms \cite{CASTRO20092175,Juangbp,IT2FNN-SVR,Juang2015,Das15,Eyoh2018,kaynak2010,Lin2014}. Recently, some type-1 correlation-aware fuzzy neural networks are presented \cite{Ebadzadeh15,Ebadzadeh2017,Pratama2014a,Pratama2014b}. In \cite{pratma2}, the interactions among input variables are considered by utilizing the Mahalanobis distance based on the covariance among input variables through each fuzzy rule. To consider uncertainty, it is considered that the means of fuzzy rules are uncertain.

In this paper, an interval type-2 correlation-aware fuzzy neural network (IT2CFNN) based on the Mamdani's FIS is proposed that can form interval type-2 fuzzy rules similar to the covered region of the target function by considering both the local interactions among input variables and also providing fuzzy sets with adaptive shapes. To define interval type-2 fuzzy rules, a new form based on uncertain shape is proposed which is more general than the usual uncertain mean and uncertain width. First, for each fuzzy rule, the input variables are mapped to a new feature space proper for defining that fuzzy rule considering the local interactions among input variables. Next, the extracted features are fed to interval type-2 fuzzy sets with uncertain and adaptive shapes. Afterward the upper and lower bounds of fuzzy rules firing strength are calculated based on the dot product as the t-norm operator. For type-reduction, an adaptive weightening formula based on the Nie-Tan operator is applied \cite{NieTan,NieTan2,Das15}. Finally the deffuzified output is calculated based on the weighted average of the rules' consequent parts. The parameters of the network are initialized by a K-Nearest Neighbor (KNN) based method and fine-tuned by hierarchical Levenberg-Marquadt (LM) optimization paradigm \cite{Marquardt63,Ebadzadeh15,Ebadzadeh2017,9170566}.

The contributions of the proposed method are summarized as follows:
\begin{enumerate}
  \item Introducing a novel concept of defining uncertainty as uncertain shape, proper for function approximation applications;
  \item Introducing a new form of interval type-2 fuzzy set with adaptive and uncertain shape which is more general than the uncertain mean and width models;
  \item Proposing a new interval type-2 fuzzy neural network which is able to learn interval type-2 fuzzy rules with uncertain shapes, similar to covered region of the target function by considering local interactions among input variables and also using interval type-2 fuzzy sets with adaptive and uncertain shapes;
  \item Initializing the architecture's parameters based on a clustering approach using the interpretability of FNNs;
  \item Fine-tuning the proposed FNN parameters by a hierarchical LM method.
\end{enumerate}

The rest of this article is organized as follows: first in section \ref{section2} the new form of uncertainty as uncertainty in shape is proposed. Next, the proposed interval type-2 fuzzy set with adaptive and uncertain shape is explained. Afterward, the architecture of the proposed method and the learning approach are presented. The experimental results are reported in section \ref{section3}. Finally, the conclusions are presented in section \ref{section4}.

\section{Proposed Model}
\label{section2}
In this section, first the new form of interval type-2 fuzzy set with adaptive and uncertain shape is introduced. Next, the advantages of this new form are explained in details. Afterward, the architecture of the proposed interval type-2 correlation-aware fuzzy neural network is presented. Next, the initialization method based on interpretability nature of FNNs is presented. Finally, the learning algorithm based on the hierarchically using the \textit{LM} optimization method is presented.

\subsection{Interval Type-2 Fuzzy Set with uncertain shape}
\label{section21}
\begin{figure}[!t]
  \centering
  \includegraphics[width=3in]{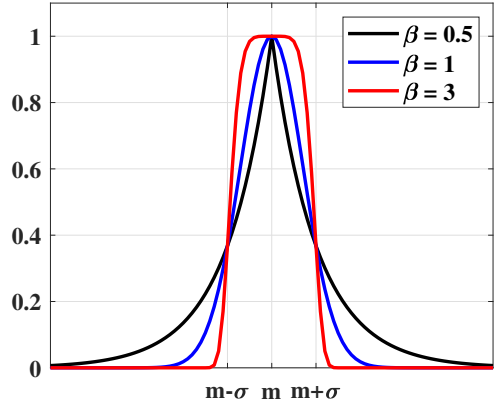}
  \caption{The proposed Type-1 membership function with adaptive shape. By changing $\beta$, different shapes of fuzzy sets are formed.}\label{fig1}
\end{figure}

In \cite{Ebadzadeh2017,AmirHaeri14} it is proposed that a general Gaussian function in the form of equation (\ref{eq1}) can model different shapes of fuzzy sets:

\begin{equation}\label{eq1}
  \mu(x;m,\sigma,\beta) = e^{(-\frac{1}{2}(\frac{x-m}{\sigma})^{2\beta^2})}
\end{equation}
where $m$ and $\sigma$ are mean and width values of the fuzzy set, and $\beta$ is a parameter that determines the shape of the fuzzy sets, which is called \textit{shape regulator}. If we select $\beta$ lower than one, the shape of the fuzzy sets become closer to triangular form and if it is selected greater than one, its shape become similar to trapezoidal form. To have Gaussian fuzzy set, we should select $\beta$ equal to one. Figure \ref{fig1} shows different shapes of fuzzy sets based on changing $\beta$.

To introduce an interval type-2 fuzzy set with adaptive and uncertain shape, the shape regulator is considered to be uncertain. Therefore, it is assumed that the shape regulator can varied between two bounds defined using extra parameter $\delta$ as follows:
\begin{equation}\label{eq2}
  \tilde\mu(x;m,\sigma,\beta,\delta) = e^{(-\frac{1}{2}\left((\frac{x-m}{\sigma})^2\right)^{(\beta^2\pm\delta^2)})}
\end{equation}

\begin{figure}[!b]
\centering
\begin{tabular}{cc}
    \subfigure[][]{\includegraphics[width = 2.7in]{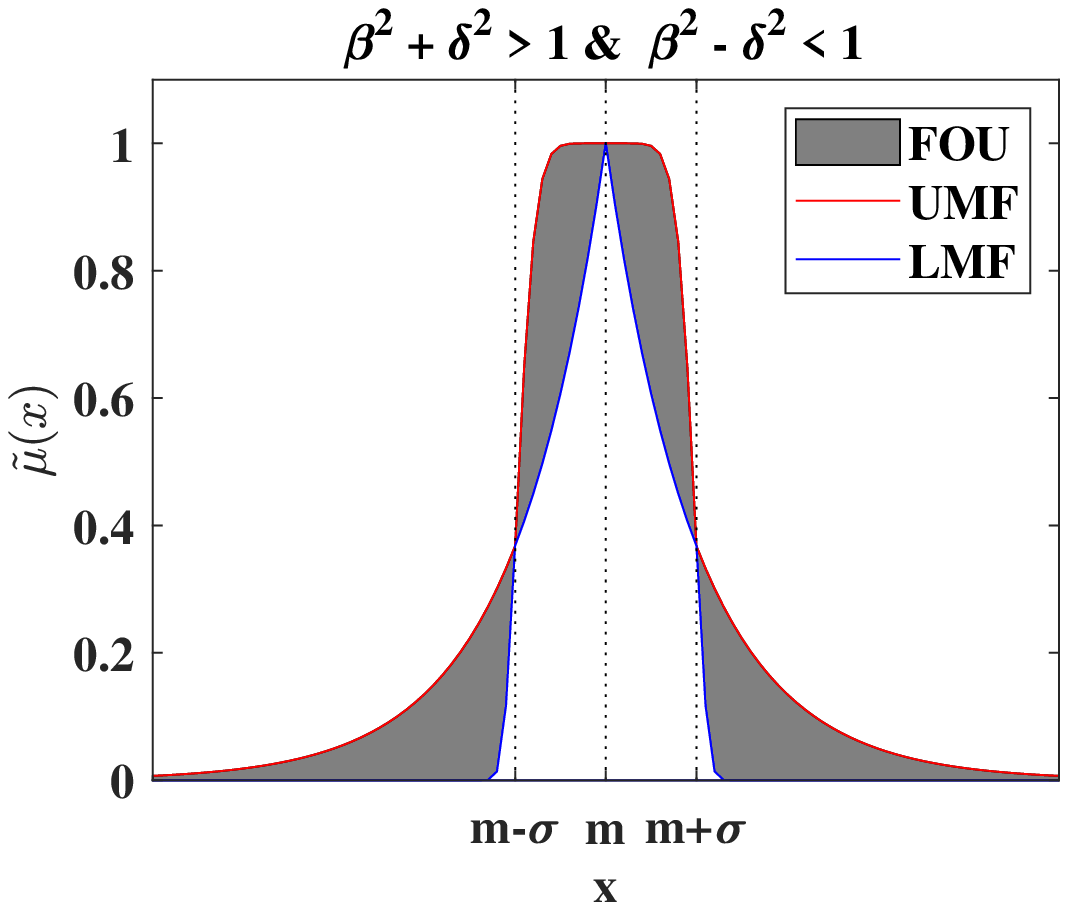}} &
    \subfigure[][]{\includegraphics[width = 2.7in]{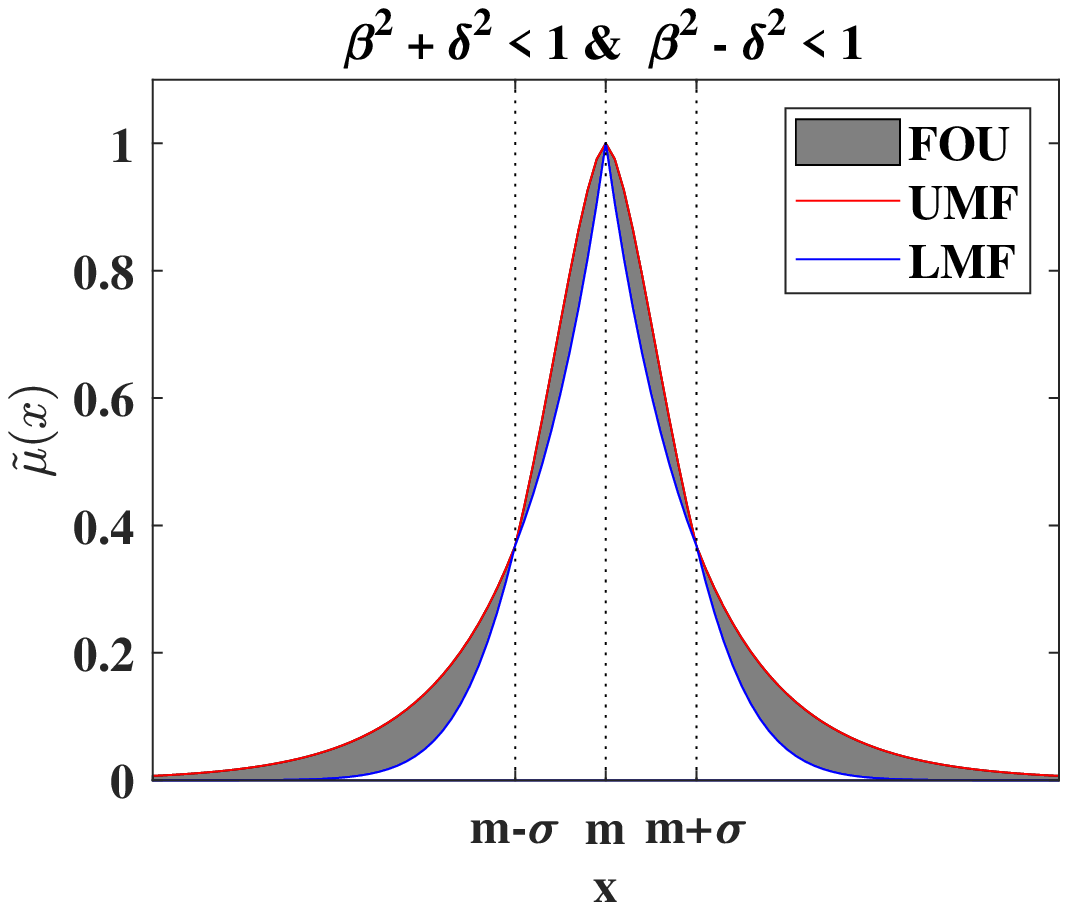}} \\
    \subfigure[][]{\includegraphics[width = 2.7in]{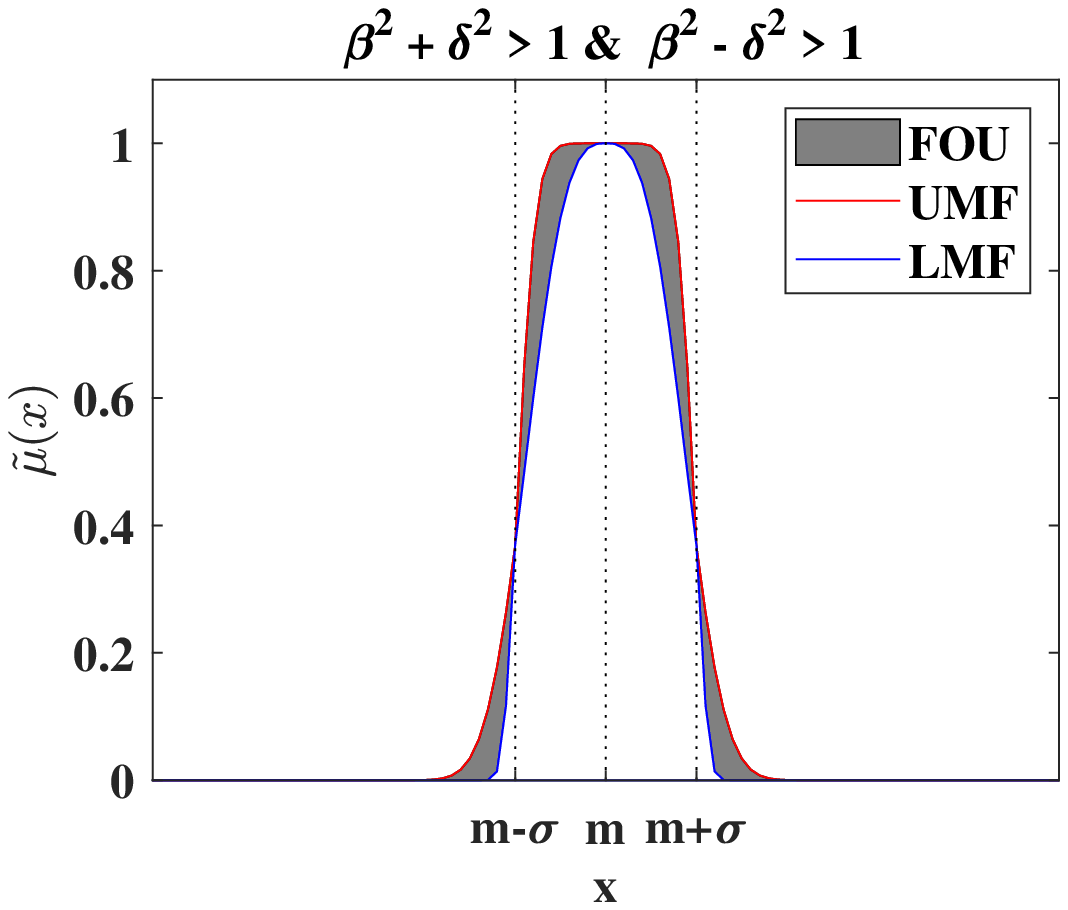}} &
    \subfigure[][]{\includegraphics[width = 2.7in]{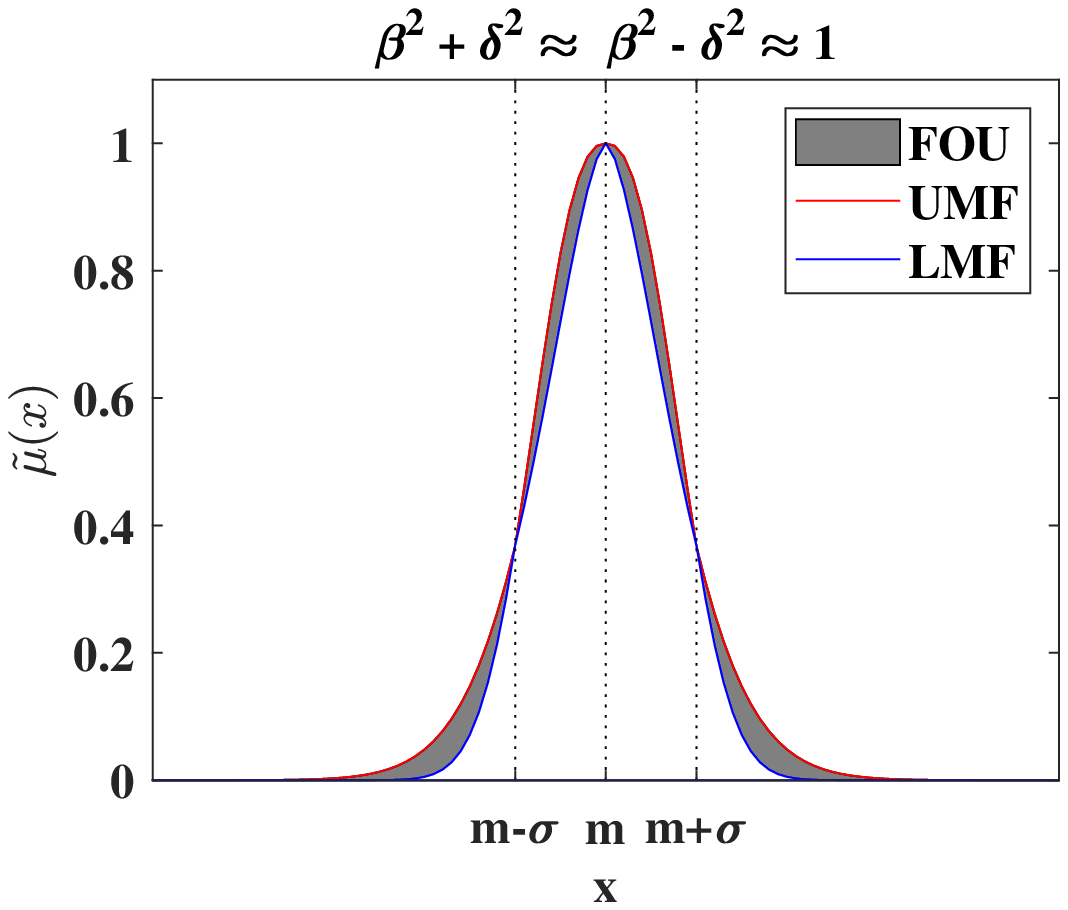}} \\
\end{tabular}
\caption{Different shapes of footprint of uncertainty (FOU) for the proposed interval type-2 fuzzy set. FOU: Footprint Of Uncertainty, UMF: Upper Membership Function, LMF: Lower Membership Function. }
\label{fig2}
\end{figure}

To determine the footprint of uncertainty (FOU), it is necessary to define the upper and lower membership functions (UMF and LMF). Here, based on the equation (\ref{eq2}) the FOU of the proposed interval type-2 fuzzy set is derived as follows :

\begin{equation}\label{eq3}
  \overline{\mu}(x;m,\sigma,\beta,\delta) = \left\{
  \begin{array}{cc}
    e^{(-\frac{1}{2}\left((\frac{x-m}{\sigma})^2\right)^{(\beta^2+\delta^2)})} &  |x-m| \leq \sigma  \\
     e^{(-\frac{1}{2}\left((\frac{x-m}{\sigma})^2\right)^{(\beta^2-\delta^2)})} & |x-m| > \sigma
  \end{array}
  \right.
\end{equation}

\begin{equation}\label{eq4}
  \underline{\mu}(x;m,\sigma,\beta,\delta) = \left\{
  \begin{array}{cc}
    e^{(-\frac{1}{2}\left((\frac{x-m}{\sigma})^2\right)^{(\beta^2-\delta^2)})} &  |x-m| \leq \sigma  \\
     e^{(-\frac{1}{2}\left((\frac{x-m}{\sigma})^2\right)^{(\beta^2+\delta^2)})} & |x-m| > \sigma
  \end{array}
  \right.
\end{equation}
where $\delta$ called "shape uncertainty regulator", and $\overline{\mu}$ and $\underline{\mu}$ are UMF and LMF, respectively. Figure \ref{fig2} shows different shapes of FOU based on the various conditions.

\begin{figure}[t]
\centering
\begin{tabular}{c}
    \subfigure[][]{\includegraphics[width = 3in]{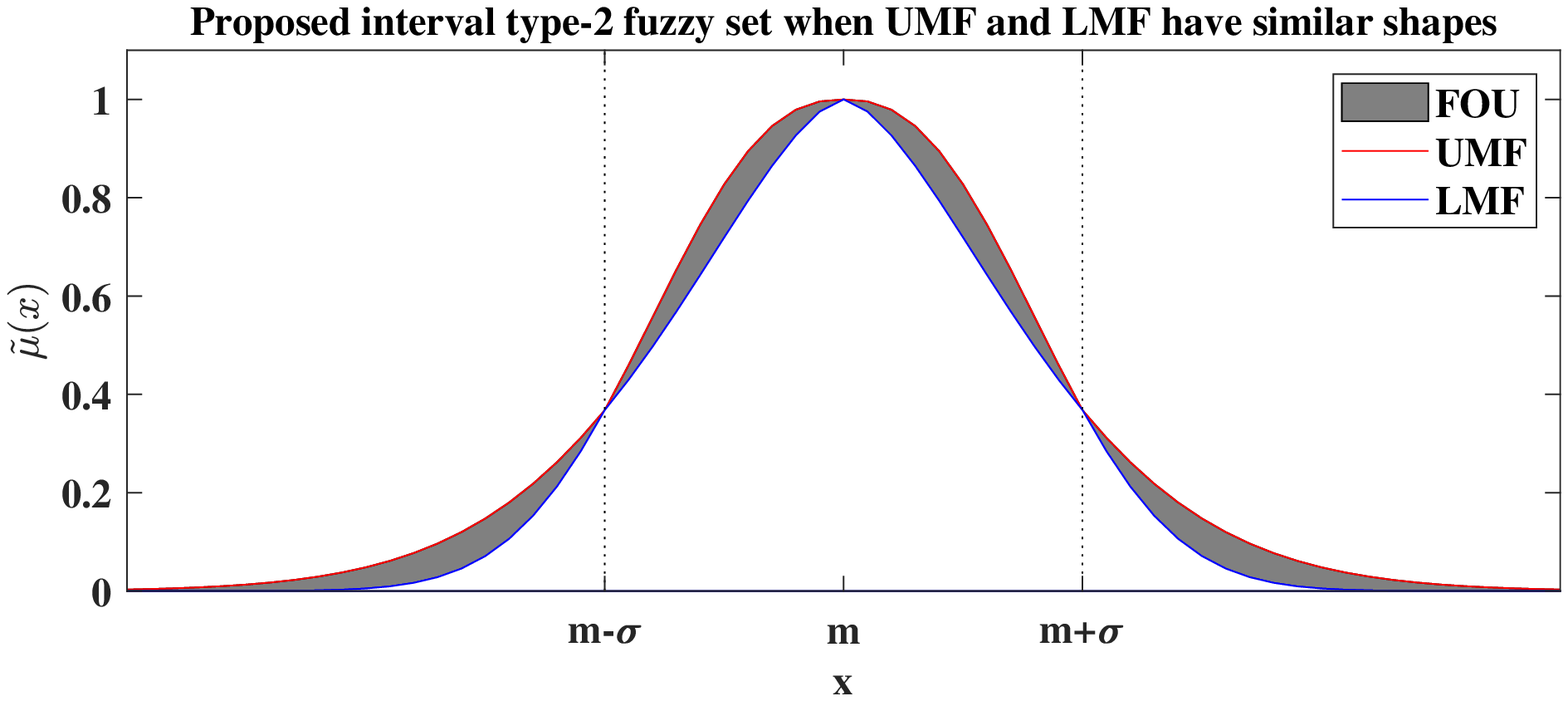}} \\
    \subfigure[][]{\includegraphics[width = 3in]{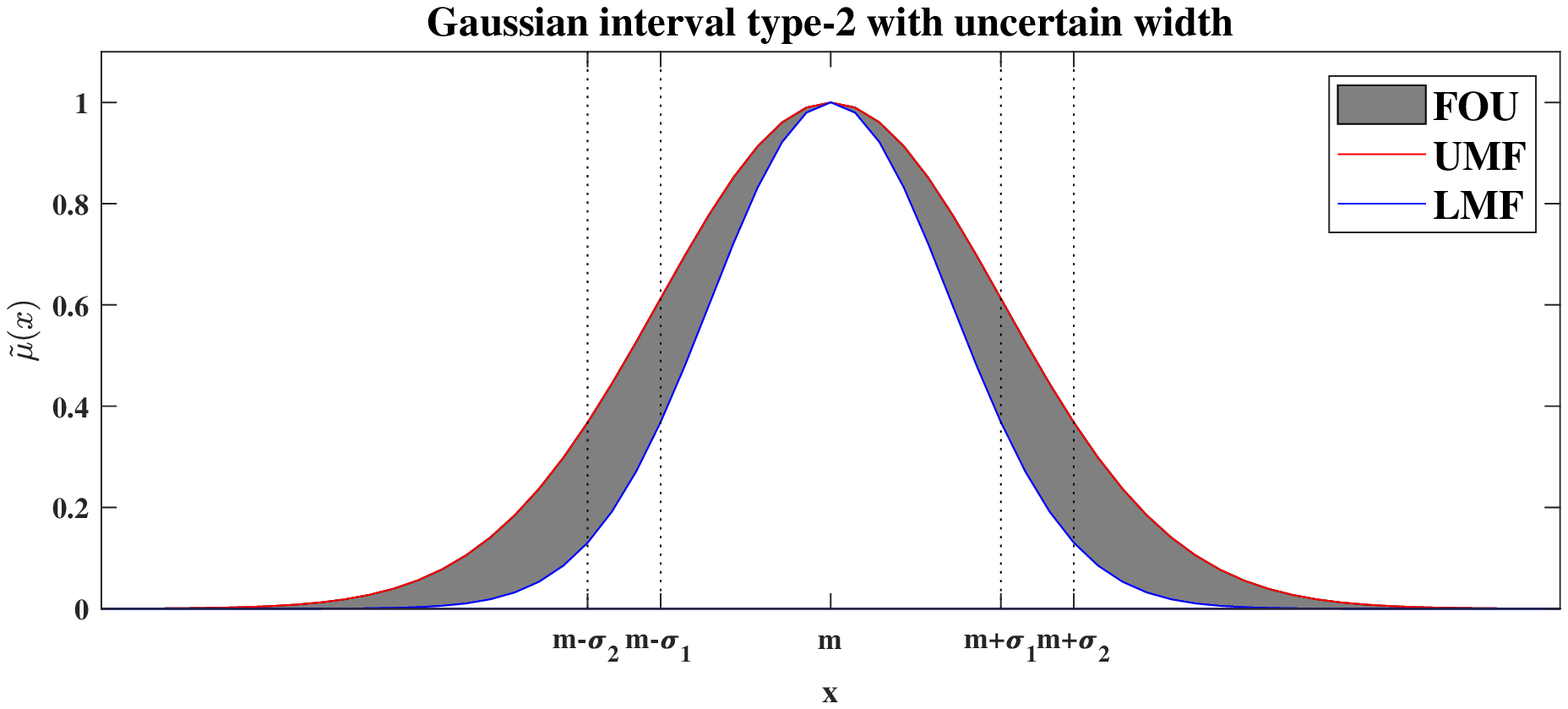}}
\end{tabular}
\caption{Comparison between the footprint of uncertainties (FOUs) of the proposed interval type-2 fuzzy set (a) with the Gaussian interval type-2 form with uncertain width (b). When both UMF and LMF of the proposed fuzzy set are similar, the created FOU is similar to the one created by the Gaussian with uncertain width. }
\label{fig3}
\end{figure}

\begin{figure}[t]
\centering
\begin{tabular}{c}
    \subfigure[][]{\includegraphics[width = 3in]{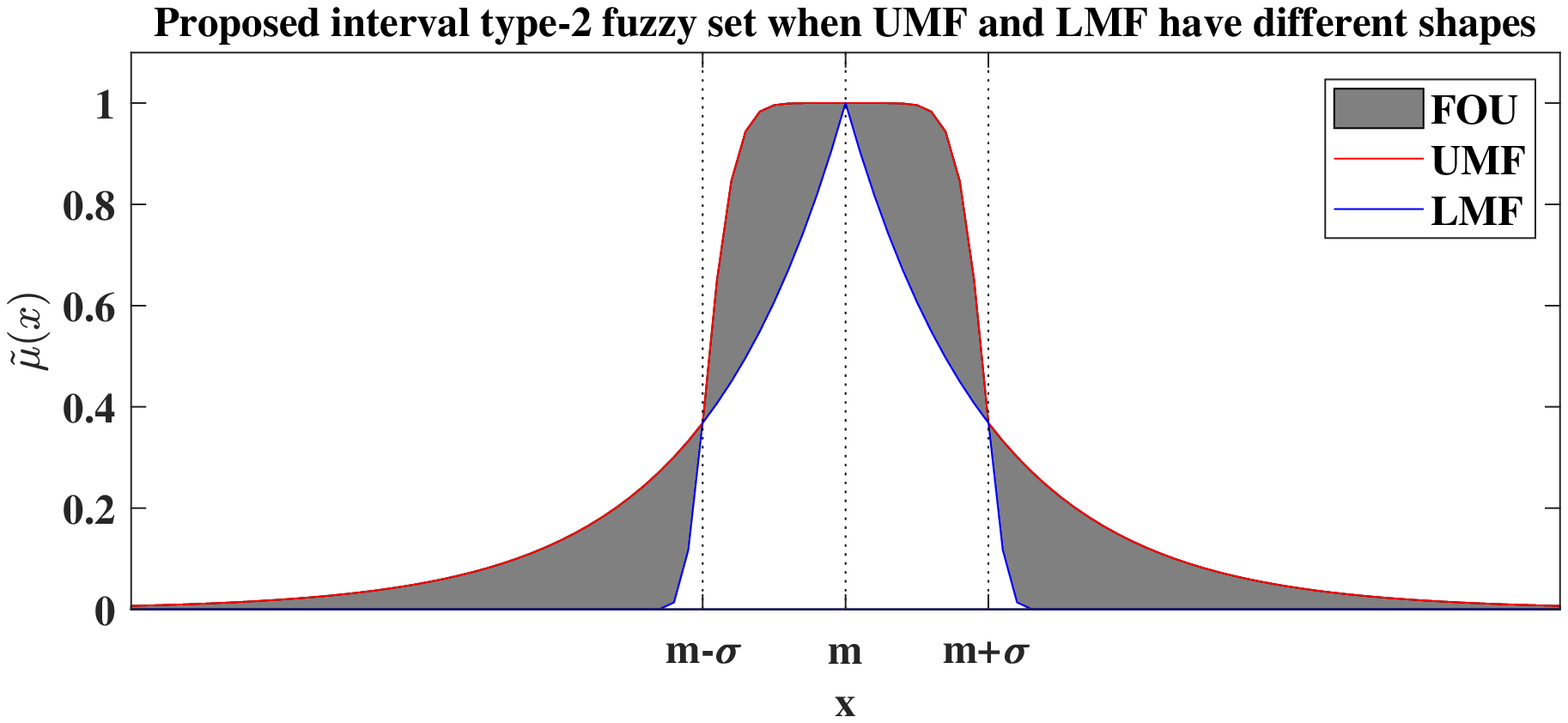}} \\
    \subfigure[][]{\includegraphics[width = 3in]{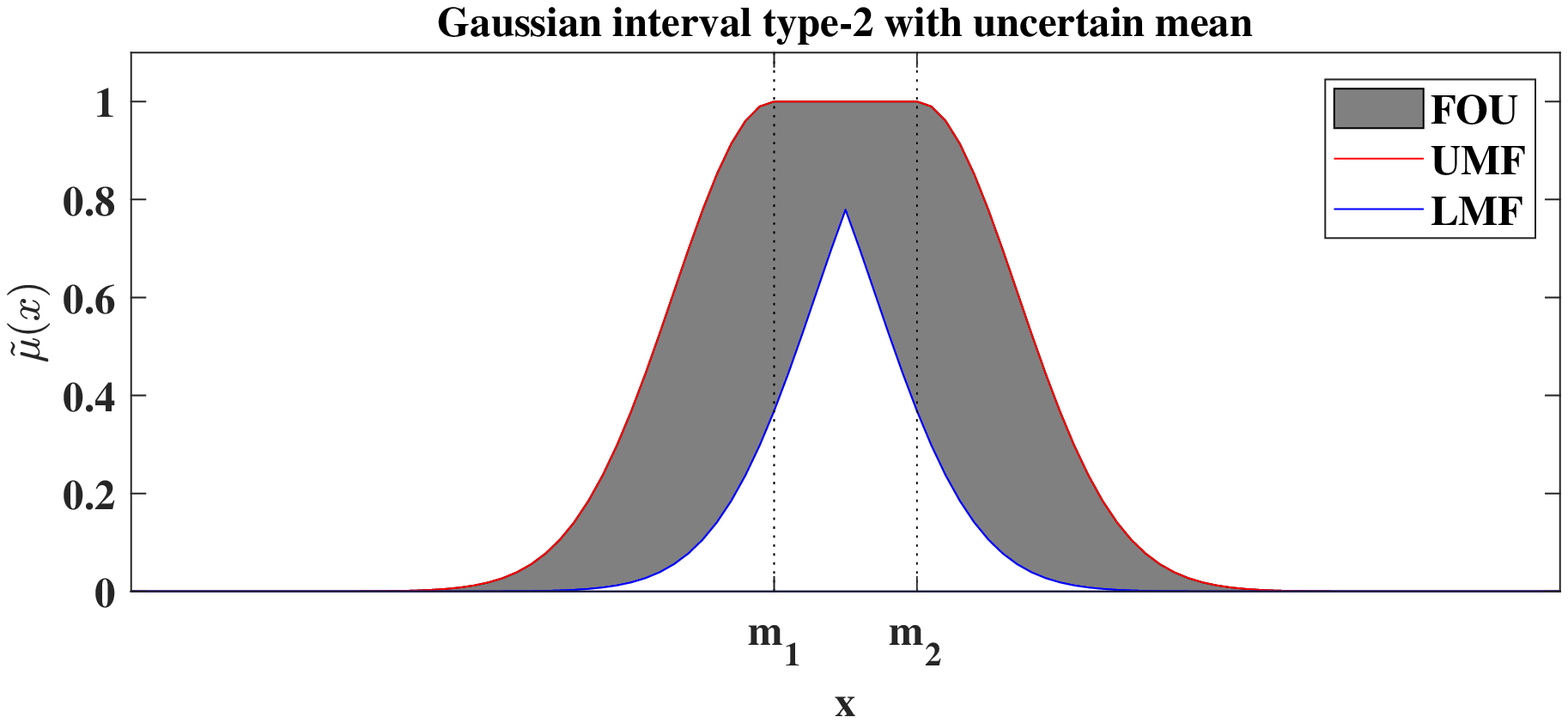}}
\end{tabular}
\caption{Comparison between the footprint of uncertainties (FOUs) of the proposed interval type-2 fuzzy set (a) with the Gaussian interval type-2 form with uncertain mean (b). When UMF and LMF of the proposed fuzzy set are not from the same category, the created FOU is similar to the one created by the Gaussian with uncertain mean. }
\label{fig4}
\end{figure}

Comparison between different shapes of created FOUs by the proposed interval type-2 fuzzy set and the FOUs by the usual Gaussian forms with uncertain mean or width reveals that:
\begin{itemize}
  \item When both UMF and LMF have the same shape category, the proposed form could cover the uncertain width approximately (see Figure \ref{fig3});
  \item When UMF and LMF have not the same shape, the proposed form could cover the uncertain mean approximately (see Figure \ref{fig4});
\end{itemize}
Consequently, the main advantage of the proposed interval type-2 fuzzy set is that its form of FOU is adaptive. Indeed, The shape of FOU is not fixed like the uncertain mean or uncertain width. The proper shape of FOU is determined based on the requirements of the problem in the learning process.

\begin{figure}[t]
  \centering
  \includegraphics[width=4in]{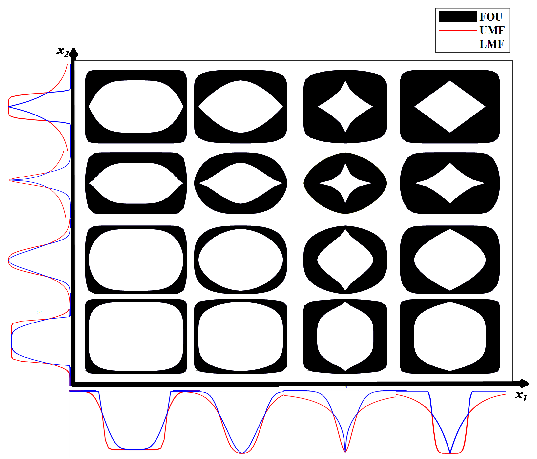}
  \caption{Different shapes of fuzzy sets along with different shapes of fuzzy rules' upper and lower firing strength and their footprint of uncertainty (FOU) based on applying the extension principle for a two dimension input sace.}\label{fig5}
\end{figure}

By defining rules' antecedent parts fuzzy sets based on the proposed membership function, fuzzy rules with different shapes and different forms of FOU are extracted. Figure \ref{fig5} compares contours of different extracted fuzzy rules' upper and lower firing strengthes along with their FOU for a two dimensional input space. It is obvious that by using the proposed interval type-2 fuzzy membership function, the shapes of fuzzy rules upper and lower strengthes and also their FOU are flexible and can be adjusted to be similar to the covered region of the target function.

\subsection{Interval Type-2 Correlation-Aware Architecture}
\label{section22}
The proposed interval type-2 correlation-ware fuzzy neural network (IT2CFNN) structure is composed of six layers including: 1- input layer, 2- transformation layer, 3- fuzzification layer, 4- fuzzy rules layer, 5- type reduction layer, and 6- output layer. The proposed architecture is shown in Figure \ref{fig6}. The proposed FNN is based on the Mamdani's FIS and the center of consequent parts fuzzy sets are provided as the outputs neuron weights. The details of each layer are explained as follows:

\begin{figure}[!b]
  \centering
  \includegraphics[width=6.5in]{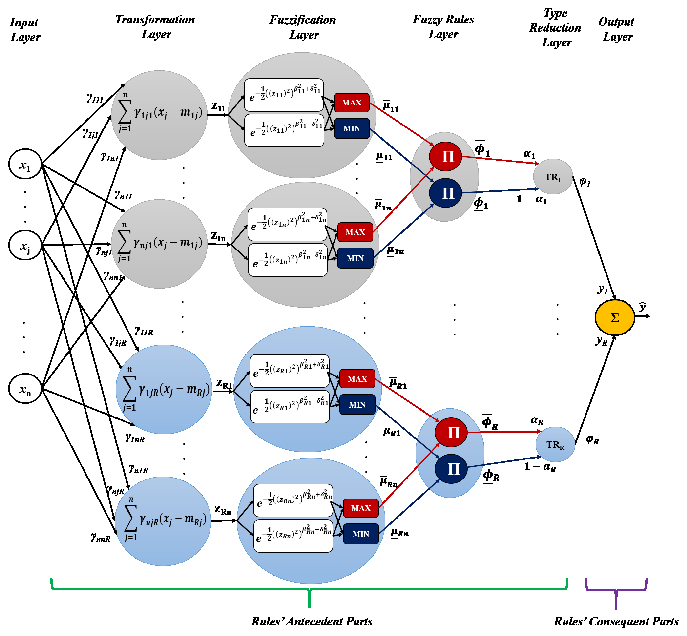}
  \caption{The proposed Architecture. In this sketch, two fuzzy rules (The first rule and the last one($R^{th}$ rule) are shown in different colors.}\label{fig6}
\end{figure}

\begin{enumerate}
  \item \textit{Input Layer}: The output vector of this layer is the input sample expressed as:
  \begin{equation}\label{eq5}
    o_1 = X = [x_1, x_2, \dots, x_n]^T
  \end{equation}
  where $o_1$ is the first layer's output vector, $X$ is the input vector, $T$ represents vector transpose, $n$ is the number of input space dimensions, and $x_i$ (i = 1,2, ...,n) is the $i^{th}$ input variable (the $i^{th}$ element of the input vector $X$).
  \item \textit{Transformation Layer}: Generally the input variables are interactive and correlated. Traditional separable fuzzy rules cannot consider these interactions. For defining proper fuzzy rule able to efficiently cover an input space with interactive variables, it is necessary to extract nonseparable fuzzy rules. Here, to define nonseparable fuzzy rules, the input variables are transformed by applying a linear transform to a new space with uncorrelated extracted features. Since these interactions vary in different regions of the input space (causing that the surface of the target function has different shapes in different regions of the input space) different transformations are applied for defining each fuzzy rule to cover each region of the target function. Next, the fuzzy sets and fuzzy rules are defined for these extracted feature in the uncorrelated new spaces. Indeed, for the $i^{th}$ fuzzy rule a new feature vector $Z_i$ is extracted as follows:
      \begin{equation}
       Z_i = \Gamma_i^T(X-M_i)
    \label{eq6}
    \end{equation}
    where $M_i$ is the center of the $i^{th}$ fuzzy rule defined as follows:
    \begin{equation}
        M_i = [m_{i,1},  m_{i,2},  \cdots, m_{i,n}]^T
    \label{eq7}
    \end{equation}
    and matrix $\Gamma_i$ is the transformation matrix for $i^{th}$ fuzzy rule to transform initial input space to non-interactive feature space proper for this rule as follows:
    \begin{equation}
        \forall i=1,2,...,R: \quad \Gamma_i = \left [ \begin{array}{cccccc}
                                                         \gamma_{1,1,i} & \gamma_{1,2,i} & \dots & \gamma_{1,j,i} & \dots &  \gamma_{1,n,i} \\
                                                         \gamma_{2,1,i} & \gamma_{2,2,i} & \dots & \gamma_{2,j,i} & \dots &  \gamma_{2,n,i} \\
                                                         \vdots & \vdots & \ddots & \vdots&\ddots & \vdots \\
                                                         \gamma_{l,1,i} & \gamma_{l,2,i} & \dots & \gamma_{l,j,i} & \dots &  \gamma_{l,n,i} \\
                                                         \vdots & \vdots & \ddots & \vdots&\ddots & \vdots \\
                                                         \gamma_{n,1,i} & \gamma_{n,2,i} & \dots & \gamma_{n,j,i} & \dots &  \gamma_{n,n,i} \\
                                                       \end{array} \right ]
    \label{eq8}
    \end{equation}
    Indeed, the element in $l^{th}$ row and $j$ column of matrix $\Gamma_i$ is shown as $\gamma_{l,j,i}$. To realize these transformations, each neuron of this layer extracts a feature for a fuzzy rule by applying a linear transformation. The output of the this layer's neuron for extracting the $j^{th}$ feature (l = 1,2, ..., n) for the $i^{th}$ fuzzy rule is calculated as follows:
    \begin{equation}
        z_{i,j} = \sum_{l=1}^{n}\left(\gamma_{j,l,i}.(x_l-m_{i,l})\right)
    \label{eq9}
    \end{equation}
    Therefore, the output vector of this layer is summarized as follows:
        \begin{equation}
        o_2 = [z_{11}, z_{12}, \dots, z_{1n}, z_{21}, \dots, z_{Rn}]^T
    \label{eq10}
    \end{equation}
    where $o_2$ is the output vector of the second layer and $R$ is the number of fuzzy rules.

  \item \textit{Fuzzification Layer}: Each neuron of this layer receives an extracted feature for a fuzzy rule from the previous layer and fuzzifies it using the proposed interval type-2 fuzzy membership function (see section \ref{section21}). Since the effect of rules' means and widthes are considered in the calculations of the second layer, here the mean and width of all fuzzy sets are assigned zero and one, respectively. Therefore, to calculate the FOU each extracted feature for each fuzzy rule, two neurons related to the upper and lower membership functions are required. Consequently, for $j^{th}$ extracted feature of the $i^{th}$ fuzzy rule ($z_{i,j}$), the upper and lower membership functions are calculated as follows:
      \begin{equation}\label{eq11}
  \overline{\mu}(z_{i,j}) = \left\{
  \begin{array}{cc}
    e^{(-\frac{1}{2}\left((z_{i,j})^2\right)^{(\beta_{i,j}^2+\delta_{i,j}^2)})} &  |z_{i,j}| \leq 1  \\
     e^{(-\frac{1}{2}\left((z_{i,j})^2\right)^{(\beta_{i,j}^2-\delta_{i,j}^2)})} & |z_{i,j}| > 1
  \end{array}
  \right.
\end{equation}

\begin{equation}\label{eq12}
  \underline{\mu}(z_{i,j}) = \left\{
  \begin{array}{cc}
    e^{(-\frac{1}{2}\left((z_{i,j})^2\right)^{(\beta_{i,j}^2-\delta_{i,j}^2)})} &  |z_{i,j}| \leq 1  \\
     e^{(-\frac{1}{2}\left((z_{i,j})^2\right)^{(\beta_{i,j}^2+\delta_{i,j}^2)})} & |z_{i,j}| > 1
  \end{array}
  \right.
\end{equation}
where $\beta_{i,j}$ and $\delta_{i,j}$ are \textit{"shape regulator"} and \textit{"shape uncertainty regulator"} parameters of the $j^{th}$ extracted feature for the $i^{th}$ fuzzy rule. The output vector of this layer can be represented as follows:
    \begin{equation}
        o_3 = [\underline{\mu}(z_{11}),\overline{\mu}(z_{11}), \underline{\mu}(z_{12}),\overline{\mu}(z_{12}), \dots, \underline{\mu}(z_{1n}),\overline{\mu}(z_{1n}), \underline{\mu}(z_{21}),\overline{\mu}(z_{21}), \dots, \underline{\mu}(z_{Rn}),\overline{\mu}(z_{Rn})]^T
    \label{eq13}
    \end{equation}

  \item \textit{Fuzzy Rules Layer}: Each neurons of this layer determines the interval of firing strength of an interval type-2 fuzzy rule. To calculate the lower firing strength of the $i^{th}$ fuzzy rule, the t-norm of the lower membership values of the extracted features for that rule ($\underline{\mu}(z_{i,j})s$, j=1,2, ..., n) is computed. Similarly, the t-norm of the upper membership values of the extracted features for the $i^{th}$ fuzzy rule ($\overline{\mu}(z_{i,j})s$, j=1,2, ..., n) determines the upper firing strength of the $i^{th}$ fuzzy rule. Therefore, by using dot product as the t-norm operator, the output of each neuron in this layer is summarized as follows:
    \begin{equation}
    \overline{\phi}_{i} = \Pi_{j=1}^{n}\overline{\mu}(z_{i,j})
    \label{eq14}
    \end{equation}

    \begin{equation}
    \underline{\phi}_{i} = \Pi_{j=1}^{n}\underline{\mu}(z_{i,j})
    \label{eq15}
    \end{equation}
    where $\underline{\phi}_{i}$ and $\overline{\phi}_{i}$ are lower and upper firing strengthes of the $i^{th}$ fuzzy rule. Consequently, the output vector of this layer is summarized as follows:
    \begin{equation}
        o_4 = [\underline{\phi}_{1},\overline{\phi}_{1},\underline{\phi}_{2},\overline{\phi}_{2}, \dots, \underline{\phi}_{R},\overline{\phi}_{R}]^T
    \label{eq16}
    \end{equation}
    where $R$ is the number of fuzzy rules.

  \item \textit{Type Reduction Layer}: Each neuron of this layer receives an interval as a rule's firing strength interval and extracts the embedded type-1 fuzzy rule by applying an adaptive version of \textit{Nie-Tan} operator (\cite{NieTan,NieTan2,Das15}) as follows:
          \begin{equation}
        \phi_i = (1-\alpha_i).\underline{\phi}_{i}+\alpha_i.\overline{\phi}_{i}, \quad 0 < \alpha_i < 1
    \label{eq17}
    \end{equation}
    where $\alpha_i \in (0,1)$ is a weight to determine the amount of importance of each interval bound. To ensure that $\alpha_i$ is always a positive number less than one, this weight is defined as follows:
              \begin{equation}
        \alpha_i = \frac{v_{i,1}^2}{v_{i,1}^2+v_{i,2}^2}  \Rightarrow 1-\alpha_i = \frac{v_{i,2}^2}{v_{i,1}^2+v_{i,2}^2}
    \label{eq18}
    \end{equation}
    where $v_{i,1}$ and $v_{i,2}$ are two adaptive weights. The output vector of this layer is as follows:
        \begin{equation}
        o_5 = [\phi_1,\phi_2, \dots, \phi_{R}]^T
    \label{eq19}
    \end{equation}

  \item \textit{Output Layer}: This layer is composed of a linear neuron which calculates the final output of the network. If the candidate output of the $i^{th}$ fuzzy rule is shown by $y_i$, the final output of the network represented by $\hat{y}$ is computed as follows:
        \begin{equation}
        \hat{y} = \sum_{i=1}^{R}\phi_i.y_i
    \label{eq20}
    \end{equation}
\end{enumerate}

Suppose the proposed structure has $R$ fuzzy rules to approximate a function with $n$ input variables. For $i^{th}$ fuzzy rule, such a structure  has $1$ consequent part parameters $y_i$, $n$ parameters of the center of the fuzzy rule $(M_i)$, $n$ \textit{shape regulator} parameter ($\beta_{ij}$), $n$ \textit{shape uncertainty regulator} parameter ($\delta_{ij}$), and $n^2$ parameters of the transformation matrix $(\Gamma_i)$. Consequently, the number of parameters, $p$ for the proposed structure is derived as follows:
\begin{equation}
    p = R(1 + n + n + n + n^2) = R(n^2 + 3n + 1)
    \label{eq21}
\end{equation}

\subsection{Initialization Method}
\label{section23}
\begin{figure}[t]
  \centering
  \includegraphics[width=3in]{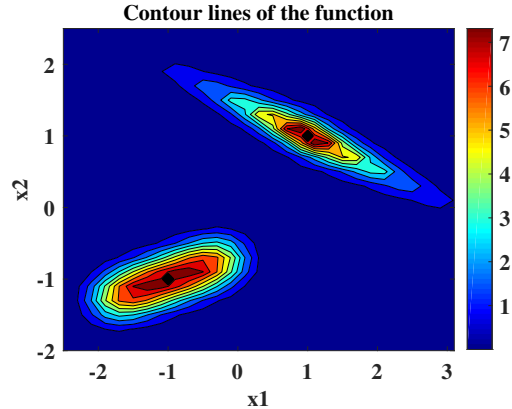}
  \caption{An illustrative example to show the proper fuzzy rules centers. In this figure, contours of a function with two input variables $x_1$ and $x_2$ are shown. The function has two hills located at $[1,1]^T$ and $[-1,-1]^T$. Therefore, two fuzzy rules similar to these hills are sufficient to properly cover the important regions of the target function. These centers are local optima of the target function. These centers of hills are proper initial values for the fuzzy rules. }\label{fig7}
\end{figure}

The interpretability of the FNN is utilized to initialize the parameters of the proposed architecture. If we look at the surface of a function as a mountain landscape (function landscape), each hill of this landscape determines an important region of the function formed based on the local interactions among input variables. In this paper it is proposed that extracted fuzzy rules should cover these hills properly. Therefore, center of each fuzzy rule should represent the center of the covered hill. The obvious feature of hill's centers is that they are local optima of the target function. Figure \ref{fig7} shows contours of a function with two input variables as an instance to show the proper centers for fuzzy rules (represented by black crosses).

To initialize the proposed FNN's parameters with $R$ fuzzy rules using $N$ training samples, first we find the $K = \frac{N}{R}$ nearest neighbors of each training sample, forming a K-Nearest Neighbor (KNN) cluster around each sample. Next, we compare the output value of each sample with the output values of its KNN. If the output value of a sample is greater or lower than the output values of all samples in its KNN, the sample will candidate for representing a fuzzy rule's center.

To choose $R$ centers from the candidate samples, the amount of KNN clusters density is investigated by calculating the mean distance of a candidate sample $X$ from its KNN as follows:
\begin{equation}
    D(X) = \frac{1}{K}\sum_{X_k \in KNN(X)}||X-X_k||_2^2
    \label{eq21_1}
\end{equation}

Lower mean distance represents denser cluster. Next, $R$ candidate samples with denser KNN clusters are selected as initial centers of the fuzzy rules ($M_is$). The output values of the extracted centers are used for initializing the consequent parts' parameters $y_i$s.

Afterward, to initialize the transformation matrices, we can utilize the correlations among input variables of the selected KNNs. Therefore, the sample covariance matrix of each selected KNN is  calculated as follows:

\begin{equation}
    \forall i=1,2, \dots, R: \quad \mathcal{Q}_i = \frac{1}{K}\sum_{X \in KNN(M_i)}\left((X - M_i)(X - M_i)^T\right)
    \label{eq22}
\end{equation}

By using the \textit{Mahalanobis distance}, rotated hyper ellipsoidal region around the center of $i^{th}$ fuzzy rule, formed based on the interactions among input variables has the following equation:
\begin{equation}
    (X-M_i)^T\mathcal{Q}^{-1}(X-M_i) = c
    \label{eq23}
\end{equation}
where $c$ represents different constant values to form different contours of the surface. On the other hand, $\mathcal{Q}_i$ can be decomposed in terms of its \textit{eigenvalues} and \textit{eigenvectors} as follows \cite{Fukunaga}:
\begin{equation}
    \begin{array}{lll}
    & \mathcal{Q}_i & = \Phi_i \Lambda_i\Phi_i^T \\
    \Rightarrow &  \mathcal{Q}_i^{-1} & = \Phi_i \Lambda_i^{-1}\Phi_i^T \\
    \Rightarrow &  \mathcal{Q}_i^{-1} & = \Phi_i \Lambda_i^{-\frac{1}{2}}\Lambda_i^{-\frac{1}{2}}\Phi_i^T \\
    \Rightarrow &  \mathcal{Q}_i^{-1} & = (\Lambda_i^{-\frac{1}{2}}\Phi_i^T)^T(\Lambda_i^{-\frac{1}{2}}\Phi_i^T) \\
    \Rightarrow &  \mathcal{Q}_i^{-1} & = \Gamma_i^T\Gamma_i
    \end{array}
        \label{eq24}
\end{equation}
where $\Phi_i$ is the \textit{eigenvector matrix} in which each column is equivalent to an \textit{eigenvector} of $\mathcal{Q}_i$, $\Lambda_i$ is the \textit{eigenvalues matrix}, which is a diagonal matrix that has the \textit{eigenvalues} of $\mathcal{Q}_i$, on its diagonal, and $\Gamma_i$ is a matrix as follows:
\begin{equation}
\left\{
\begin{array}{ll}
     \Gamma_i = &\Lambda_i^{-\frac{1}{2}}\Phi_i^T  \\
     \mathcal{Q}_i^{-1} = &\Gamma_i^{T}\Gamma_i
\end{array}
\right.
\label{eq25}
\end{equation}

By substituting equation (\ref{eq25}) in equation (\ref{eq23}) we have:
\begin{equation}
\begin{array}{ll}
     &(X-M_i)^T\mathcal{Q}^{-1}(X-M_i) = c \\
     \Rightarrow &(X-M_i)^T\Gamma_i^{T}\Gamma_i(X-M_i) = c \\
     \Rightarrow &(\Gamma_i(X-M_i))^T(\Gamma_i(X-M_i)) = c \\
\end{array}
    \label{eq26}
\end{equation}
Therefore, if we define a new feature space for the $i^{th}$ fuzzy rule as:
\begin{equation}
    Z_i = (\Gamma_i(X-M_i))
    \label{eq27}
\end{equation}
based on the equations (\ref{eq26}) and (\ref{eq27}) we have the following relation for contours of the surface in the new feature space:
\begin{equation}
    Z_i^T Z_i = c
    \label{eq28}
\end{equation}
which represents circular contours instead of the ellipsoidal ones. Indeed, the local correlations among input variables in the region of a hill are considered in this transformation and the consequent new feature space has uncorrelated dimensions. Therefore, the transformation matrix of $i^{th}$ fuzzy rule is initialized by this obtained $\Gamma_i$ (see equation (\ref{eq25})).

To initialize \textit{shape regulators} and \textit{shape uncertainty regulators} ($\beta_{ij}s$ and $\delta_{i,j}s$), it is assumed that the shape of all fuzzy sets are initially near Gaussian. Therefore, we set $\beta_{ij} = 1 \quad (i = 1,2, \dots, R \, \& j = 1, 2, \dots, n)$ and $\delta_{ij} = \epsilon \quad (i = 1,2, \dots, R \, \& j = 1, 2, \dots, n)$, where $\epsilon$ is a small positive number close to one. Finally, we consider $\alpha_i s = 0.5$ to weight two frontiers of the FOUs equally. Consequently, we set $v_{i1} = v_{i2} \quad (i = 1,2, \dots, R)$. The initialization algorithm is summarized in Algorithm \ref{alg1}.

\begin{algorithm}[t]
\scriptsize
  \textbf{Inputs:} \\
  $\quad$ \text{training samples $X$ along with their desired output values $Y^*$} \\
  $\quad$ \text{number of rules ($R$), number of samples ($N$)} \\
  \textbf{Outputs:}\\
   $\quad$ \text{initial values of center of rules ($M$), transformation matrices ($\Gamma$), shape regulators ($\beta$),}\\
    $\quad\,$\text{shape uncertainty regulators ($\delta$), consequent parts' parameters ($y_i$), and type reduction weights ($v$)}\\
  $K \gets N/R$ \;
   $\forall i=1,2, \dots, R \quad \& \quad j=1,2, \dots, n$: \text{$\beta_{ij} \gets 1$ and $\delta_{ij} \gets \epsilon$} \;
   $\forall i=1,2, \dots, R$: \text{$v_{i1} \gets 0.5$ and $v_{i2} \gets 0.5$} \;
   $CandidateCenters \gets \emptyset$ \;
   $Centers \gets \emptyset$ \;
   $CentersOutputs \gets \emptyset$ \;
    \For {$k \gets 1$ \KwTo $N$}{
        \text{put K nearest neighbors of $X_k$ in $KNN(X_k)$} \;
        \If {\text{$y_k$ greater or lower than all samples in $KNN(X_k)$}}{
        \text{add $X_k$ to CandidateCenters} \;
        calculate $D(X_k)$ based on eq. (\ref{eq21_1}) \;
        }
    }
    \text{find $R$ samples in CandidateCenters with lowest amount of $D$ as Centers} \;
    \text{put output values of centers in CentersOutputs}\;
     \For {$i \gets 1$ \KwTo $R$}{
     $M_i \gets Centers_i$ \;
     $y_i \gets CentersOutputs_i$ \;
     }

  \For{$i \gets 1$ \KwTo $R$}{
  $\mathcal{Q}_i = 0$ \;
    \For{$X \in KNN(M_i)$}{
    $\mathcal{Q}_i \gets \mathcal{Q}_i + (X-M_i)(X-M_i)^T$ \;
    }
    $\mathcal{Q}_i \gets \frac{1}{K}.\mathcal{Q}_i$ \;
    \text{extract \textit{eigenvector matrix} and \textit{eigenvalues matrix} of $\mathcal{Q}_i$ as $\Phi_i$ and $\Lambda_i$, respectively} \;
    $\Gamma_i \gets \Lambda_i^{-\frac{1}{2}}\Phi_i^T$
\;  }
 \caption{Initialization algorithm}
 \label{alg1}
 \end{algorithm}

\subsection{Hierarchical Fine-Tuning Method}
\label{section24}
After initializing the parameters based on the proposed algorithm in section \ref{section23}, a hierarchical Levenberg-Marquadt (LM) optimization technique is applied to fine-tune these parameters based on minimizing the squared error between the network's prediction and the desired output values. Since there are different groups of parameters with different domains, the LM method is applied hierarchically to fine-tune parameters belong to each group separately \cite{Ebadzadeh2017}. The efficiency of the hierarchical method versus the classic non-hierarchical (changing all parameters at once) is already investigated in our previous study \cite{Ebadzadeh2017}.

Different groups of parameters to learn are: 1- fuzzy rules centers ($M_i$ i=1,2, ..., R), 2- transformation matrices ($\Gamma_i$ i=1,2, ..., R), 3- shape regulators ($\beta_{ij}$, i=1,2,..., R and j=1,2, ..., n), 4- shape uncertainty regulators ($\delta_{ij}$, i=1,2,..., R and j=1,2, ..., n), 5- type reduction weights ($v_{i1}$ and $v_{i2}$, i=1,2, ..., R), and 6- consequent parts' parameters ($y_i$, i=1,2, ..., R). These parameters are learned separately and iteratively. Therefore, the parameter vectors $\rho_1$ to $\rho_6$ related to these 6 mentioned categories of adaptive parameters, are defined as follows:

\begin{eqnarray}
\rho_1 & = & \left[\gamma_{111}, \gamma_{121}, \cdots, \gamma_{211}, \cdots \gamma_{jli}, \cdots \gamma_{nnR},\right]^T
\label{eq29}\\
\rho_2 & = & \left[m_{11}, m_{12}, \cdots m_{1n}, \cdots, m_{ij}, \cdots m_{Rn}\right]^T
\label{eq30}\\
\rho_3 & = & \left[y_1, y_2, \cdots, y_i, \cdots, y_R\right]^T
\label{eq31}\\
\rho_4 & = & \left[\beta_{11}, \beta_{12}, \cdots \beta_{1n}, \cdots \beta_{ij}, \cdots, \beta_{Rn}\right]^T
\label{eq32}\\
\rho_5 & = & \left[\delta_{11}, \delta_{12}, \cdots \delta_{1n}, \cdots, \delta_{ij}, \cdots, \delta_{Rn}\right]^T
\label{eq322}\\
\rho_6 & = & \left[v_{11}, v_{12}, v_{21}, v_{22}, \cdots, v_{i1}, v_{i2}, \cdots, v_{R1}, v_{R2}\right]^T
\label{eq33}
\end{eqnarray}

We define $\hat{Y} = [\hat{y}_1, \cdots, \hat{y}_k, \cdots, \hat{y}_N]^T$ as the vector containing network's predictions for $N$ training samples and $Y^* = [y_1^*, \cdots, y_k^*, \cdots, y_N^*]^T$ as the desired output of these training instances. The error vector $e$ is defined as follows:
\begin{equation}\label{eq34}
  e = \hat{Y}-Y^*
\end{equation}
Moreover, the \textit{Jacobian} matrix for the $c^{th}$ category of parameters (c=1,2,...,6) $J_c$ is defined as follows:
\begin{equation}
    J_c = \frac{\partial \hat{Y}}{\partial \alpha_c} =     \left[\begin{array}{ccc}
        \frac{\partial \hat{y}_1}{\partial \alpha_{c}^1} & \cdots & \frac{\partial \hat{y}_1}{\partial \alpha_{c}^{p_c}} \\
        \vdots &  \ddots & \vdots \\
        \frac{\partial \hat{y}_N}{\partial \alpha_{c}^{1}} & \cdots & \frac{\partial \hat{y}_N}{\partial \alpha_{c}^{p_c}}
    \end{array}\right]
    \label{eq35}
\end{equation}
where $p_c$ is the number of parameters of $c^{th}$ category and based on equations (\ref{eq29}) to (\ref{eq33}) we have
$p_1 = Rn^2$, $p_2 = Rn$, $p_3 = R$, $p_4 = Rn$, $p_5 = Rn$ and $p_6 = 2R$.

According to $LM$ method, parameter vector $\rho_c$ (c=1,2,...,6) is updated iteratively as follows \cite{Marquardt63,Menhaj94,Ebadzadeh15,Han16,Ebadzadeh2017}:

\begin{eqnarray}
\rho_c^{new} & = & \rho_c^{old} + \Delta\rho_c
\label{eq361}\\
\Delta\rho_c & = & -(J_c^TJ_c + \lambda I)^{-1}J_c^Te
\label{eq36}
\end{eqnarray}
where, $I$ is the Identity matrix, $e$ is error vector calculated in equation (\ref{eq33}), $J_c$ is the Jacobian matrix of $c^{th}$ parameter category defined in equation (\ref{eq34}), $\lambda$ is a scaler related to the trust region's size and plays the role of a learning rate \cite{Conn00}. If $\lambda$ is very large, $LM$ performs like \textit{Gradient Descent} method. In the case of using very small value for $\lambda$, $LM$ method approaches \textit{Gauss-Newton} which is a second order optimization method. Therefore, this parameter can be justified to combine proper features of these algorithms: the local search ability of \textit{Gauss-Newton} with the global properties of \textit{Gradient Descent} \cite{Ampazis02,Menhaj94}. Generally. to adjust the value of $\lambda$, its value is changed in each epoch based on the following rule \cite{Ampazis02,Menhaj94}:
\begin{itemize}
  \item If the current error is lower than the previous step, decrease $\lambda$ by dividing its value by $\eta$ ($\eta > 1$). Consequently the behavior of $LM$ approaches the \textit{Gauss-Newton} method;
  \item If the current error is greater than the previous step, increase $\lambda$ by multiplying its value to $\eta$ ($\eta > 1$). Consequently the behavior of $LM$ approaches the \textit{Gradient Descent} method.
\end{itemize}
we can summarize these rules as follows \cite{Ebadzadeh15}:
\begin{equation}
    \left\{\begin{array}{lll}
            \text{if }E(\rho + \Delta\rho) < E(\rho) & then & \lambda \gets \lambda/\eta \\
            \text{if }E(\rho + \Delta\rho) > E(\rho) & then & \lambda \gets \lambda\times \eta \\
            \text{if }E(\rho + \Delta\rho) = E(\rho) & then & \lambda \gets \lambda
          \end{array}
    \right.
    \label{eq37}
\end{equation}
where, $E$ is the sum of squared error, and $c$ is a constant greater than one.

Therefore, to learn parameters based on the $LM$ method it is necessary to derive elements of $J_c$ (c=1,2,...,6). To calculate these elements, first the network's output for $k^{th}$ sample, $\hat{y}_k$ is defined as a function of network parameters ($\rho$) and $k^{th}$ training sample vector ($X_k$) based on equations (\ref{eq5}) to (\ref{eq20}) as follow:
\begin{equation}
    \begin{aligned}
        \hat{y}_k &= f(X_k,\rho)\\
        & = \sum_{i=1}^R\left(y_i.\phi_i^{(k)}\right)\\
        & = \sum_{i=1}^R\left(y_i.((1-\alpha_i).\underline{\phi}_i^{(k)} + \alpha_i.\overline{\phi}_i^{(k)})\right)\\
        & = \sum_{i=1}^R\left(y_i.(\frac{v_{i1}^2}{v_{i1}^2+v_{i2}^2}.\Pi_{j=1}^n\underline{\mu}(z^{(k)}_{i,j}) + \frac{v_{i2}^2}{v_{i1}^2+v_{i2}^2}.\Pi_{j=1}^n\overline{\mu}(z^{(k)}_{i,j}))\right) \\
        & = \sum_{i=1}^R\left(y_i.\left(\frac{v_{i1}^2}{v_{i1}^2+v_{i2}^2}.\Pi_{j=1}^n\underline{\mu}\left(\sum_{l=1}^{n}\left(\gamma_{j,l,i}.(x^{(k)}_l-m_{i,l})\right)\right) + \frac{v_{i2}^2}{v_{i1}^2+v_{i2}^2}.\Pi_{j=1}^n\overline{\mu}\left(\sum_{l=1}^{n}\left(\gamma_{j,l,i}.\psi^{(k)}_{ij}\right)\right)\right)\right)\\
    \end{aligned}
    \label{eq38}
\end{equation}
where, $\phi_i^{(k)}$ is firing strength of $i^{th}$ fuzzy rule based on the $k^{th}$ sample ($X_k$), $x^{(k)}_l$ is the $l^{th}$ dimension of the $k^{th}$ training instance, $z^{(k)}_{i,j}$ is the $j^{th}$ feature extracted for defining $R^{th}$ fuzzy rule based on the $k^{th}$ training sample, and finally $\psi^{(k)}_{ij}$ is an auxiliary variable as follows:
\begin{equation}\label{eqaux}
  \psi^{(k)}_{ij} = x^{(k)}-m_{ij}
\end{equation}

According to equations (\ref{eq29}) - (\ref{eq33}), and definition of Jacobian matrix in equation (\ref{eq34}), to obtain the \textit{Jacobian} matrices, $\frac{\partial \hat{y}_k}{\partial y_i}$, $\frac{\partial \hat{y}_k}{\partial v_{i1}}$, $\frac{\partial \hat{y}_k}{\partial v_{i2}}$, $\frac{\partial \hat{y}_k}{\partial \gamma_{jli}}$, $\frac{\partial \hat{y}_k}{\partial \beta_{ij}}$, $\frac{\partial \hat{y}_k}{\partial \delta_{ij}}$, and $\frac{\partial \hat{y}_k}{\partial m_{ij}}$ must be derived. Using network's function derived in equation (\ref{eq38}) and the \textit{Chain Rule} these derivatives are extracted as follows:

\begin{equation}
    \frac{\partial  \hat{y}_k}{\partial y_i} = \phi_{i}^{(k)}
    \label{eq39}
\end{equation}

\begin{equation}
    \frac{\partial  \hat{y}_k}{\partial v_{i1}} = y_i.\left(\frac{2v_{i2}^2v_{i1}}{(v_{i1}^2+v_{i2}^2)^2}\underline{\phi}_{i}^{(k)}+\frac{-2v_{i2}^2v_{i1}}{(v_{i1}^2+v_{i2}^2)^2}\overline{\phi}_{i}^{(k)}\right)
    \label{eq40}
\end{equation}

\begin{equation}
    \frac{\partial  \hat{y}_k}{\partial v_{i2}} = y_i.\left(\frac{-2v_{i1}^2v_{i2}}{(v_{i1}^2+v_{i2}^2)^2}\underline{\phi}_{i}^{(k)}+\frac{2v_{i1}^2v_{i2}}{(v_{i1}^2+v_{i2}^2)^2}\overline{\phi}_{i}^{(k)}\right)
    \label{eq41}
\end{equation}

\begin{equation}
    \begin{aligned}
    \frac{\partial  \hat{y}_k}{\partial \beta_{ij}} &= y_i.\left(\frac{v_{i1}^2}{v_{i1}^2+v_{i2}^2}.\frac{\partial \underline{\phi}_{i}^{(k)}}{\partial \beta_{ij}}+\frac{v_{i2}^2}{v_{i1}^2+v_{i2}^2}.\frac{\partial \overline{\phi}_{i}^{(k)}}{\partial \beta_{ij}}\right) \\
    &=\left\{\begin{array}{cc}
               -y_i.ln((z^{(k)}_{ij})^2).\beta_{ij}.\left(\frac{v_{i1}^2}{v_{i1}^2+v_{i2}^2}.\underline{\phi}_i^{(k)}.(z^{(k)}_{ij})^{2(\beta_{ij}^2-\delta_{ij}^2)}
               +\frac{v_{i2}^2}{v_{i1}^2+v_{i2}^2}.\overline{\phi}_i^{(k)}.(z^{(k)}_{ij})^{2(\beta_{ij}^2+\delta_{ij}^2)}\right) &  |z_{ij}^{(k)}| \leq 1\\
               -y_i.ln((z^{(k)}_{ij})^2).\beta_{ij}.\left(\frac{v_{i1}^2}{v_{i1}^2+v_{i2}^2}.\underline{\phi}_i^{(k)}.(z^{(k)}_{ij})^{2(\beta_{ij}^2+\delta_{ij}^2)}
               +\frac{v_{i2}^2}{v_{i1}^2+v_{i2}^2}.\overline{\phi}_i^{(k)}.(z^{(k)}_{ij})^{2(\beta_{ij}^2-\delta_{ij}^2)}\right) &  |z_{ij}^{(k)}| > 1
             \end{array}
             \right.
    \end{aligned}
    \label{eq342}
\end{equation}

\begin{equation}
    \begin{aligned}
    \frac{\partial  \hat{y}_k}{\partial \delta_{ij}} &= y_i.\left(\frac{v_{i1}^2}{v_{i1}^2+v_{i2}^2}.\frac{\partial \underline{\phi}_{i}^{(k)}}{\partial \delta_{ij}}+\frac{v_{i2}^2}{v_{i1}^2+v_{i2}^2}.\frac{\partial \overline{\phi}_{i}^{(k)}}{\partial \delta_{ij}}\right) \\
    &=\left\{\begin{array}{cc}
               -y_i.ln((z^{(k)}_{ij})^2).\delta_{ij}.\left(\frac{v_{i2}^2}{v_{i1}^2+v_{i2}^2}.\overline{\phi}_i^{(k)}.(z^{(k)}_{ij})^{2(\beta_{ij}^2+\delta_{ij}^2)} -\frac{v_{i1}^2}{v_{i1}^2+v_{i2}^2}.\underline{\phi}_i^{(k)}.(z^{(k)}_{ij})^{2(\beta_{ij}^2-\delta_{ij}^2)}\right)
                &  |z_{ij}^{(k)}| \leq 1\\
               -y_i.ln((z^{(k)}_{ij})^2).\delta_{ij}.\left(\frac{v_{i1}^2}{v_{i1}^2+v_{i2}^2}.\underline{\phi}_i^{(k)}.(z^{(k)}_{ij})^{2(\beta_{ij}^2+\delta_{ij}^2)}
               -\frac{v_{i2}^2}{v_{i1}^2+v_{i2}^2}.\overline{\phi}_i^{(k)}.(z^{(k)}_{ij})^{2(\beta_{ij}^2-\delta_{ij}^2)}\right) &  |z_{ij}^{(k)}| > 1
             \end{array}
             \right.
    \end{aligned}
    \label{eq43}
\end{equation}

\begin{equation}
    \begin{aligned}
    \frac{\partial  \hat{y}_k}{\partial m_{ij}} &= y_i.\left(\frac{v_{i1}^2}{v_{i1}^2+v_{i2}^2}.\frac{\partial \underline{\phi}_{i}^{(k)}}{\partial m_{ij}}+\frac{v_{i2}^2}{v_{i1}^2+v_{i2}^2}.\frac{\partial \overline{\phi}_{i}^{(k)}}{\partial m_{ij}}\right) \\
    &= y_i.\left(\frac{v_{i1}^2}{v_{i1}^2+v_{i2}^2}.\sum_{l=1}^{n}\left(\frac{\partial \underline{\phi}_{i}^{(k)}}{\partial z_{il}}.\frac{\partial z_{il}}{\partial m_{ij}}\right)+\frac{v_{i2}^2}{v_{i1}^2+v_{i2}^2}.\sum_{l=1}^{n}\left(\frac{\partial \overline{\phi}_{i}^{(k)}}{\partial z_{il}}.\frac{\partial z_{il}}{\partial m_{ij}}\right)\right) \\
    &=\left\{\begin{array}{cc}
                y_i.\sum_{l=1}^{n}\gamma_{lji}\left(\frac{v_{i1}^2}{v_{i1}^2+v_{i2}^2}.\underline{\phi}_i^{(k)}.(\beta_{il}^2-\delta_{il}^2).(z^{(k)}_{il})^{(2\beta_{il}^2-2\delta_{il}^2-1)}+\frac{v_{i2}^2}{v_{i1}^2+v_{i2}^2}.\overline{\phi}_i^{(k)}(\beta_{il}^2+\delta_{il}^2).(z^{(k)}_{il})^{(2\beta_{il}^2+2\delta_{il}^2-1)}\right)
               &|z_{ij}^{(k)}| \leq 1\\
                y_i.\sum_{l=1}^{n}\gamma_{lji}\left(\frac{v_{i1}^2}{v_{i1}^2+v_{i2}^2}.\underline{\phi}_i^{(k)}.(\beta_{il}^2+\delta_{il}^2).(z^{(k)}_{il})^{(2\beta_{il}^2+2\delta_{il}^2-1)}+\frac{v_{i2}^2}{v_{i1}^2+v_{i2}^2}.\overline{\phi}_i^{(k)}(\beta_{il}^2-\delta_{il}^2).(z^{(k)}_{il})^{(2\beta_{il}^2-2\delta_{il}^2-1)}\right)
               &|z_{ij}^{(k)}| > 1
             \end{array}
             \right.
    \end{aligned}
    \label{eq44}
\end{equation}

\begin{equation}
    \begin{aligned}
    \frac{\partial  \hat{y}_k}{\partial \gamma_{jli}} &= y_i.\left(\frac{v_{i1}^2}{v_{i1}^2+v_{i2}^2}.\frac{\partial \underline{\phi}_{i}^{(k)}}{\partial \gamma_{jli}}+\frac{v_{i2}^2}{v_{i1}^2+v_{i2}^2}.\frac{\partial \overline{\phi}_{i}^{(k)}}{\partial \gamma_{jli}}\right) \\
    &= y_i.\left(\frac{v_{i1}^2}{v_{i1}^2+v_{i2}^2}.\left(\frac{\partial \underline{\phi}_{i}^{(k)}}{\partial z_{ij}}.\frac{\partial z_{ij}}{\partial \gamma_{jli}}\right)+\frac{v_{i2}^2}{v_{i1}^2+v_{i2}^2}.\left(\frac{\partial \overline{\phi}_{i}^{(k)}}{\partial z_{ij}}.\frac{\partial z_{ij}}{\partial \gamma_{jli}}\right)\right) \\
    &=\left\{\begin{array}{cc}
                -y_i.\psi^{(k)}_{il}\left(\frac{v_{i1}^2}{v_{i1}^2+v_{i2}^2}.\underline{\phi}_i^{(k)}.(\beta_{il}^2-\delta_{il}^2).(z^{(k)}_{il})^{(2\beta_{il}^2-2\delta_{il}^2-1)}+\frac{v_{i2}^2}{v_{i1}^2+v_{i2}^2}.\overline{\phi}_i^{(k)}(\beta_{il}^2+\delta_{il}^2).(z^{(k)}_{il})^{(2\beta_{il}^2+2\delta_{il}^2-1)}\right)
               &|z_{ij}^{(k)}| \leq 1\\
               - y_i.\psi^{(k)}_{il}\left(\frac{v_{i1}^2}{v_{i1}^2+v_{i2}^2}.\underline{\phi}_i^{(k)}.(\beta_{il}^2+\delta_{il}^2).(z^{(k)}_{il})^{(2\beta_{il}^2+2\delta_{il}^2-1)}+\frac{v_{i2}^2}{v_{i1}^2+v_{i2}^2}.\overline{\phi}_i^{(k)}(\beta_{il}^2-\delta_{il}^2).(z^{(k)}_{il})^{(2\beta_{il}^2-2\delta_{il}^2-1)}\right)
               &|z_{ij}^{(k)}| > 1
             \end{array}
             \right.
    \end{aligned}
    \label{eq45}
\end{equation}

The proposed fine-tuning algorithm is summarized in Algorithm \ref{alg2}.

\begin{algorithm}[!t]
\scriptsize
  \textbf{Inputs:} \\
  $\quad$ \text{number of rules ($R$), number of samples ($N$), inital value of $\lambda$} \\
  $\quad$ \text{network's initial parameters calculated by alg. \ref{alg1}} \\
  $\quad$ \text{training samples $X$ along with their desired output values $Y^*$} \\
  \textbf{Outputs:}\\
   $\quad$ \text{final values of center of rules ($M$), transformation matrices ($\Gamma$), shape regulators ($\beta$),}\\
    $\quad\,$\text{shape uncertainty regulators ($\delta$), consequent parts' parameters ($y_i$), and type reduction weights ($v$)}\\
\caption{Fine-tuning Algorithm}\label{alg2}
\While{Validation data error is decreasing}{
    \tcc{General loop:}
    \For{$c \gets 1 \quad \KwTo \quad 6$}{
        \text{Make parameter vector $\rho_c$ (eqns. (\ref{eq29}) \KwTo (\ref{eq33}))}\;
        \While{Validation data error is decreasing}{
            \tcc{Local loop:}
            \text{Make error vector $e$}\;
            \text{Make Jacobian matrices $J_c$ (eqn. (\ref{eq35}) along with eqns. (\ref{eq39}) \KwTo (\ref{eq45}))}\;
            \text{Calculate $\Delta\rho_c$ (eqn. (\ref{eq36}))}\;
            $\rho_c \gets \rho_c + \Delta\rho_c$\;
            \text{Update parameters based on $\rho_c$}\;
            \text{Update $\lambda$ (eqn. (\ref{eq37}))}\;
        }
    }
}
\end{algorithm}

\section{Experimental Results}
\label{section3}
In this section the performance of the proposed method is studied and compared to the other type-1 and type-2 methods in the literature. First, the ability of the network to extract fuzzy rules similar to the covered region of the target function is evaluated. To show this ability the synthetic nonlinear function used in \cite{Ebadzadeh2017} is utilized. In this experiment the effect of adding noise to training data on the footprint of uncertainty is investigated. Moreover, the process of extracting non-separable interval type-2 fuzzy rules is demonstrated.

Next, the ability of the proposed model to encounter noisy data is evaluated on noisy Mackey-Glass time-series data similar to experiments performed in previous studies  \cite{Das15,IT2FNN-SVR, BAKLOUTI2018,LUO2019}, and the results of the proposed method are compared to the results of some previous type-1 and interval type-2 fuzzy neural networks.

Afterward, the performance of the proposed method in some real-world problems including financial time-series prediction, time-series prediction, nonlinear system identification, is studied.

In order to measure the precision and compare with the past proposed methods, \textit{Root Mean Squared Error} (RMSE) (\cite{SOFMLS,Das15,Ebadzadeh15,Rubio15,Mansouri16}) is used which is defined as follows:
\begin{equation}
    RMSE = \sqrt{\frac{1}{N}\sum_{k=1}^N\left(y_k^*-\hat{y}_k\right)^2}
    \label{eq46}
\end{equation}
where $N$ is the number of training samples, $y_k^*$ and $\hat{y}_k$ are the desired and predicted output values for the $k^{th}$ training sample.

\subsection{Experiment 1: Noisy Synthetic Function Approximation}
\label{section31}
To show the ability of the proposed model to extract fuzzy rules similar to the surface of the covered region of the target function, a synthetic function with highly correlated input variables is used \cite{Ebadzadeh2017}. If $x_1$ and $x_2$ are input dimensions defined in $[-2 \,\, 3]$ and $X = [x_1 \, x_2]^T$ is the input vector, two vectors $Z_1 = [z_{11} \, \, z_{12}]^T$ and $Z_2 = [z_{21} \, \, z_{22}]^T$ are defined as follows:

\begin{equation}
    \left\{
    \begin{array}{cc}
      Z_1 &= A_1(X-[-0.7 \,\, 1.3]^T) \\
      Z_2 &= A_2(X-[1.2 \,\, -0.6]^T) \\
    \end{array}
    \right.
    \label{eq47}
\end{equation}
where $A_1$ and $A_2$ are defined as follows (more details are presented in \cite{Ebadzadeh2017}):

\begin{equation}
    \left\{
    \begin{array}{cc}
      A_1 &=\left[\begin{array}{cc}
              -4.5721 & -2.1415 \\
              -0.3855 & 0.8230
            \end{array}\right] \\\\
      A_2 &=\left[\begin{array}{cc}
              -2.4801 & 0.8700 \\
              -0.3149 & 0.8976
            \end{array}\right] \\
    \end{array}
    \right.
    \label{eq48}
\end{equation}

\begin{figure}[t]
\centering
\begin{tabular}{cc}
    \subfigure[][]{\includegraphics[width = 2.25in]{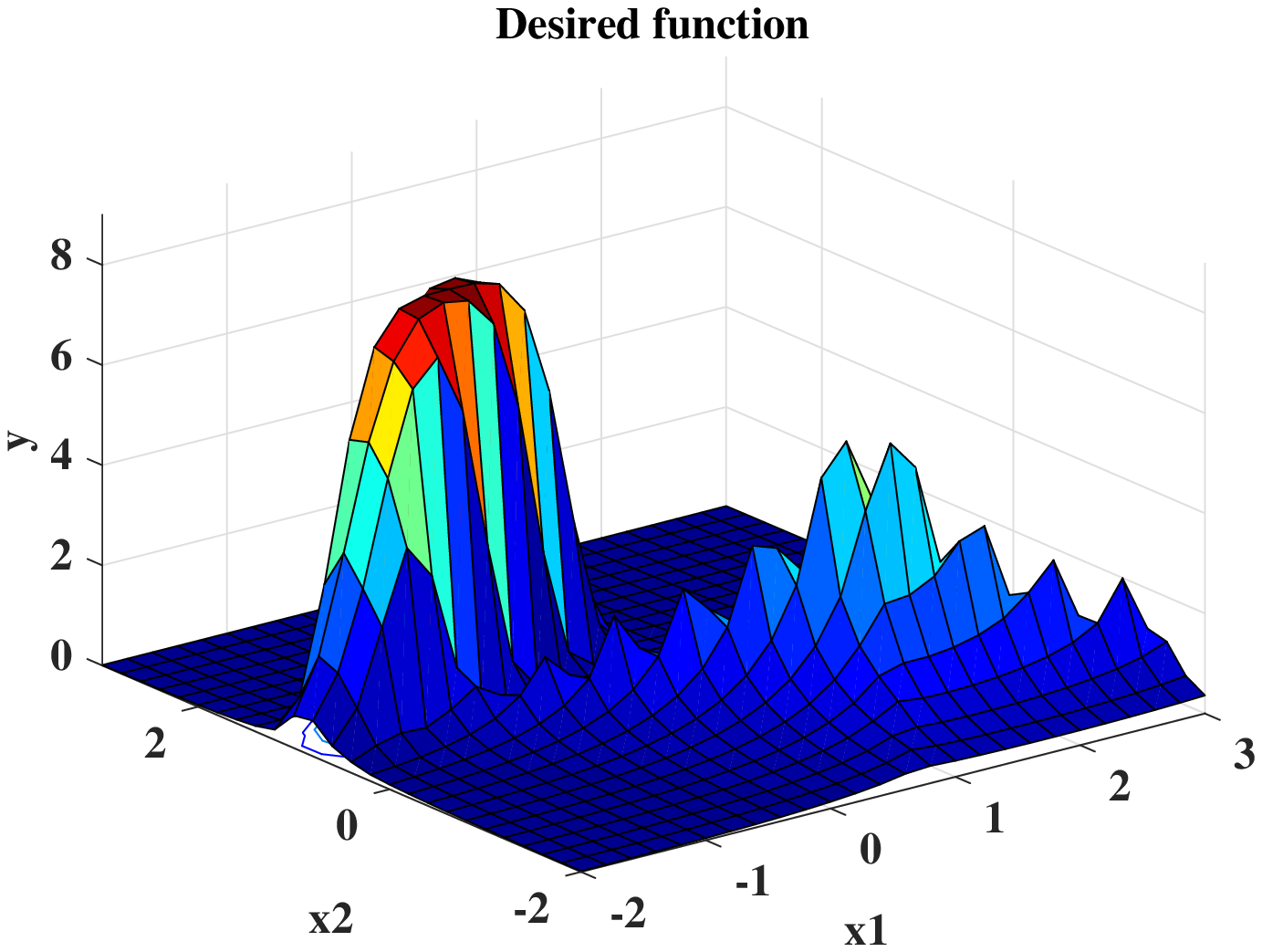}} &
    \subfigure[][]{\includegraphics[width = 2.25in]{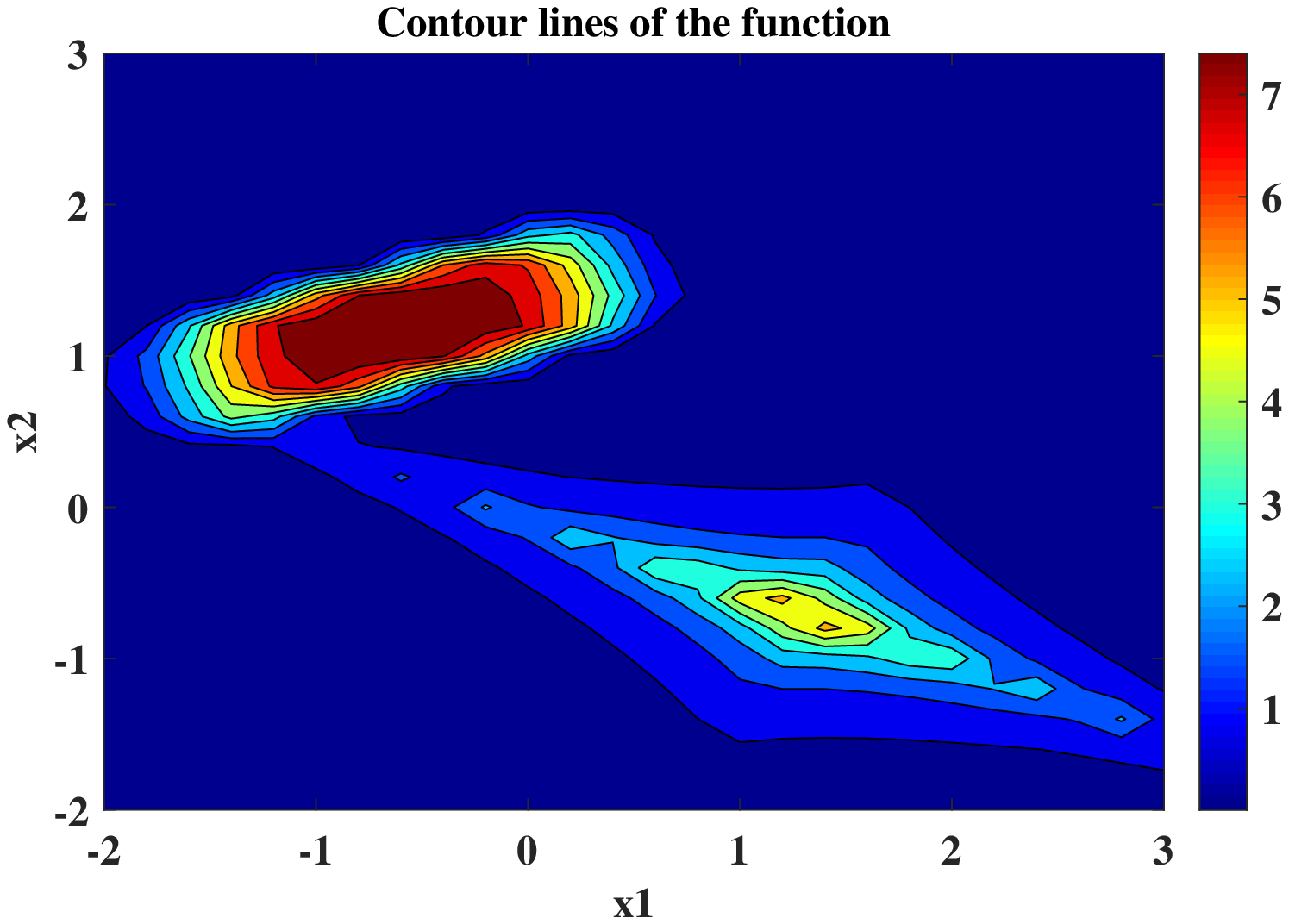}}
\end{tabular}
\caption{Target function of experiment 1.}
\label{fig_exp1}
\end{figure}

The target function is defined as follow:

\begin{equation}
    y = 10.e^{-(z_{11}^{0.6}+z_{12}^{0.8})}+8.e^{-(z_{21}^6+z_{22}^4)}
    \label{eq49}
\end{equation}
The surface of this function along with its contours are shown in Figure \ref{fig_exp1}. It is observed that this function composed of two regions with different local interactions among input variables. Therefore, ideally two fuzzy rules similar to these regions are sufficient for approximating it.

\begin{figure}[!t]
\centering
\begin{tabular}{ccc}
    \multicolumn{3}{c}{\subfigure[][]{\includegraphics[width = 2in]{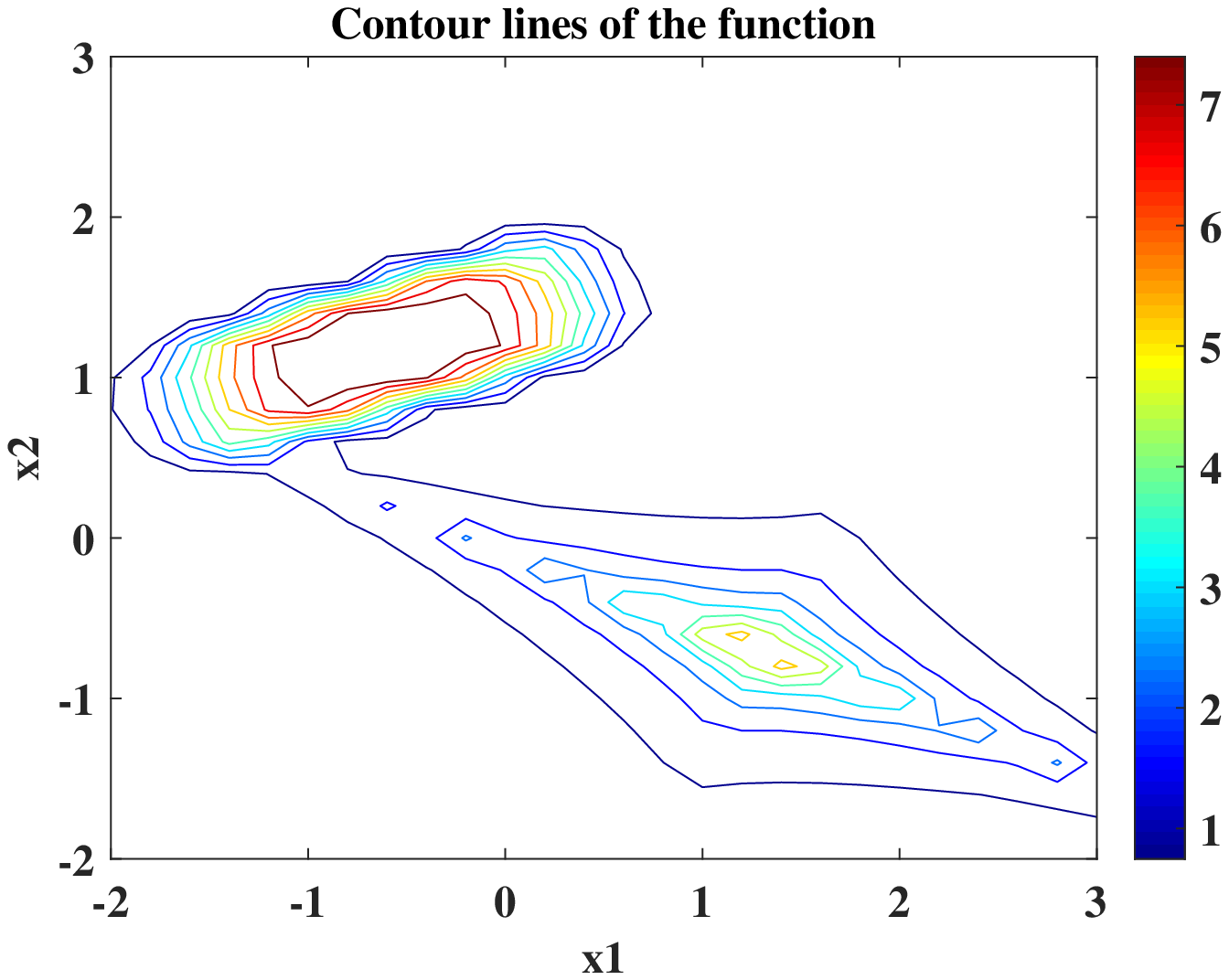}}}\\
    \subfigure[][]{\includegraphics[width = 2in]{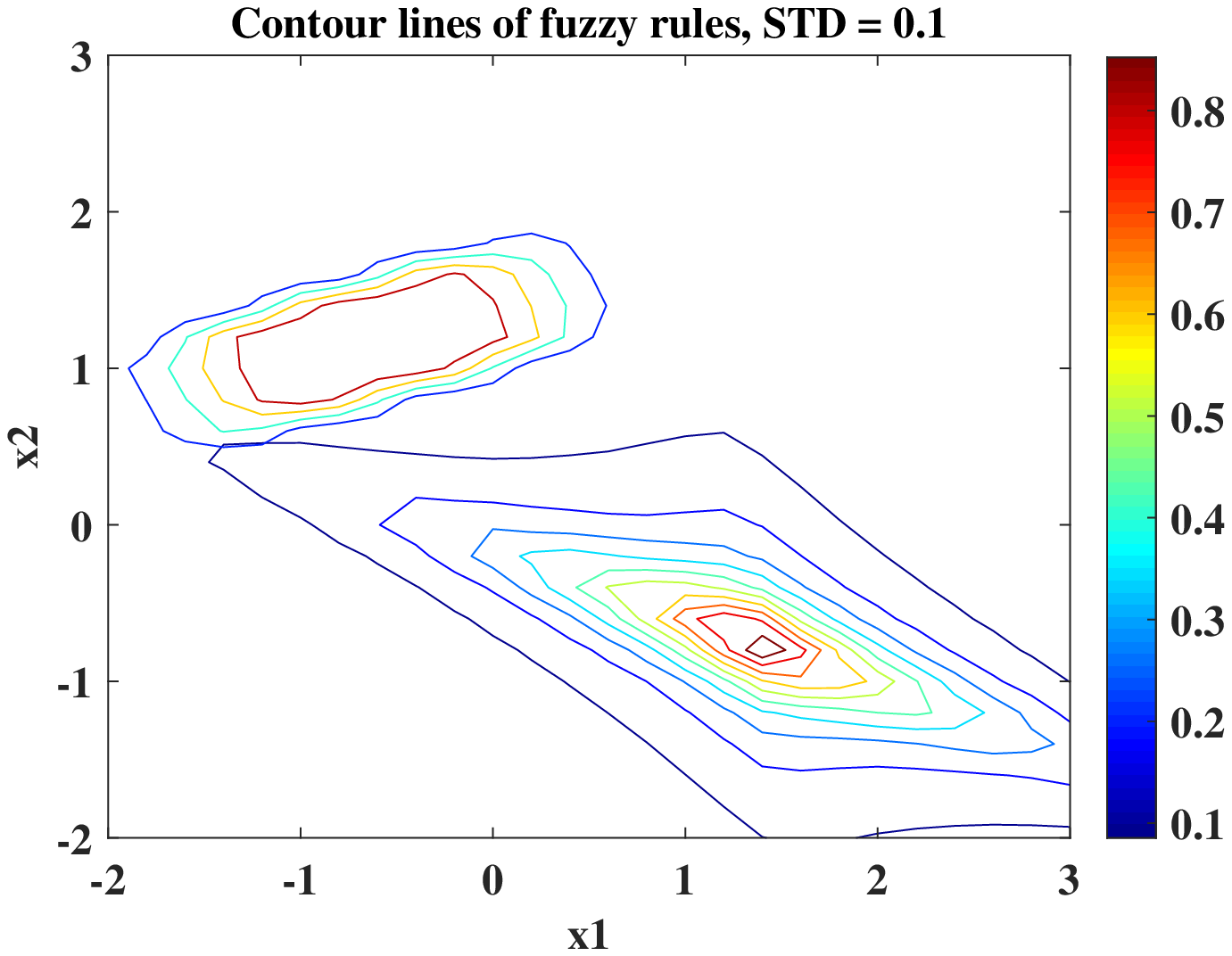}} &
    \subfigure[][]{\includegraphics[width = 2in]{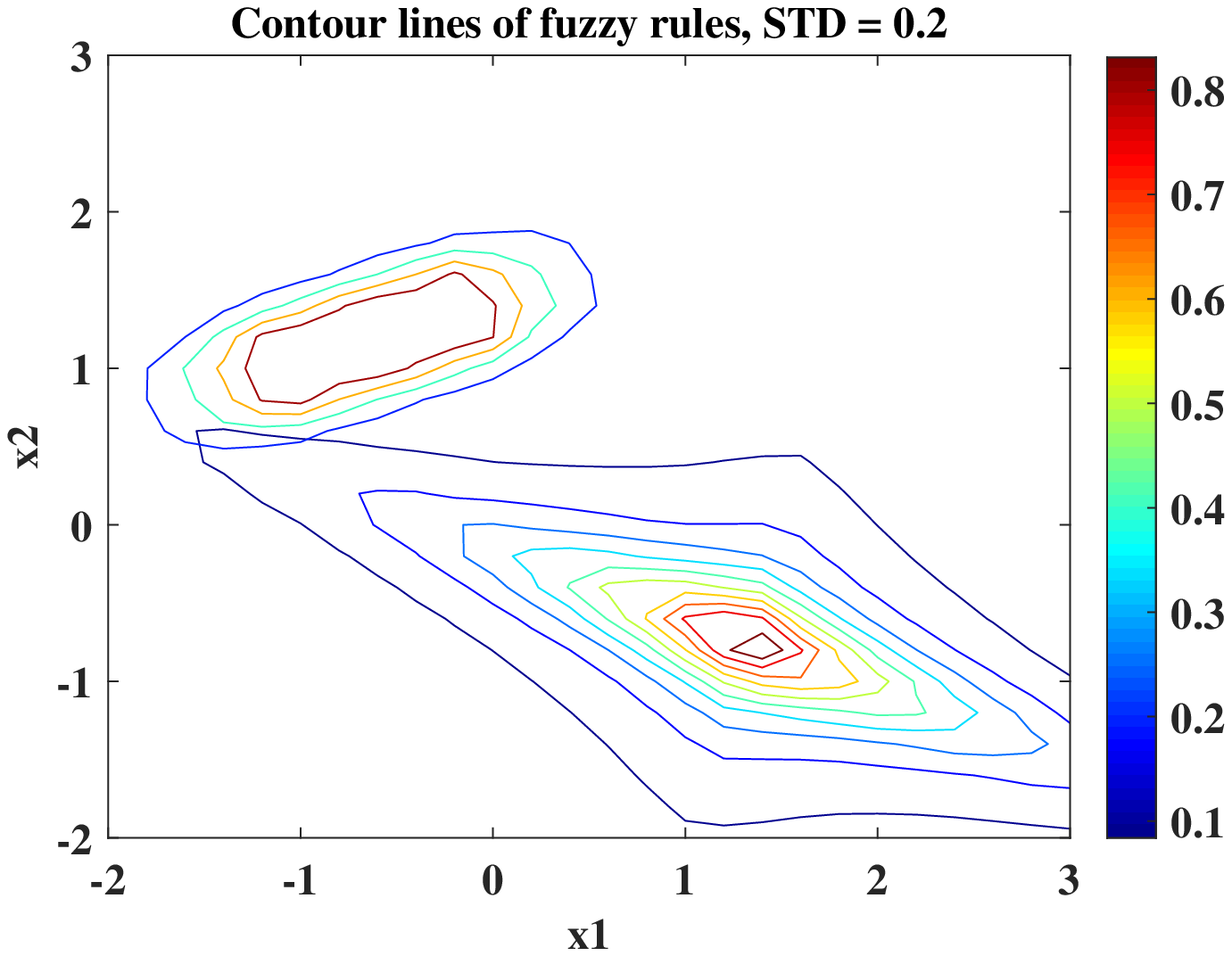}} &
    \subfigure[][]{\includegraphics[width = 2in]{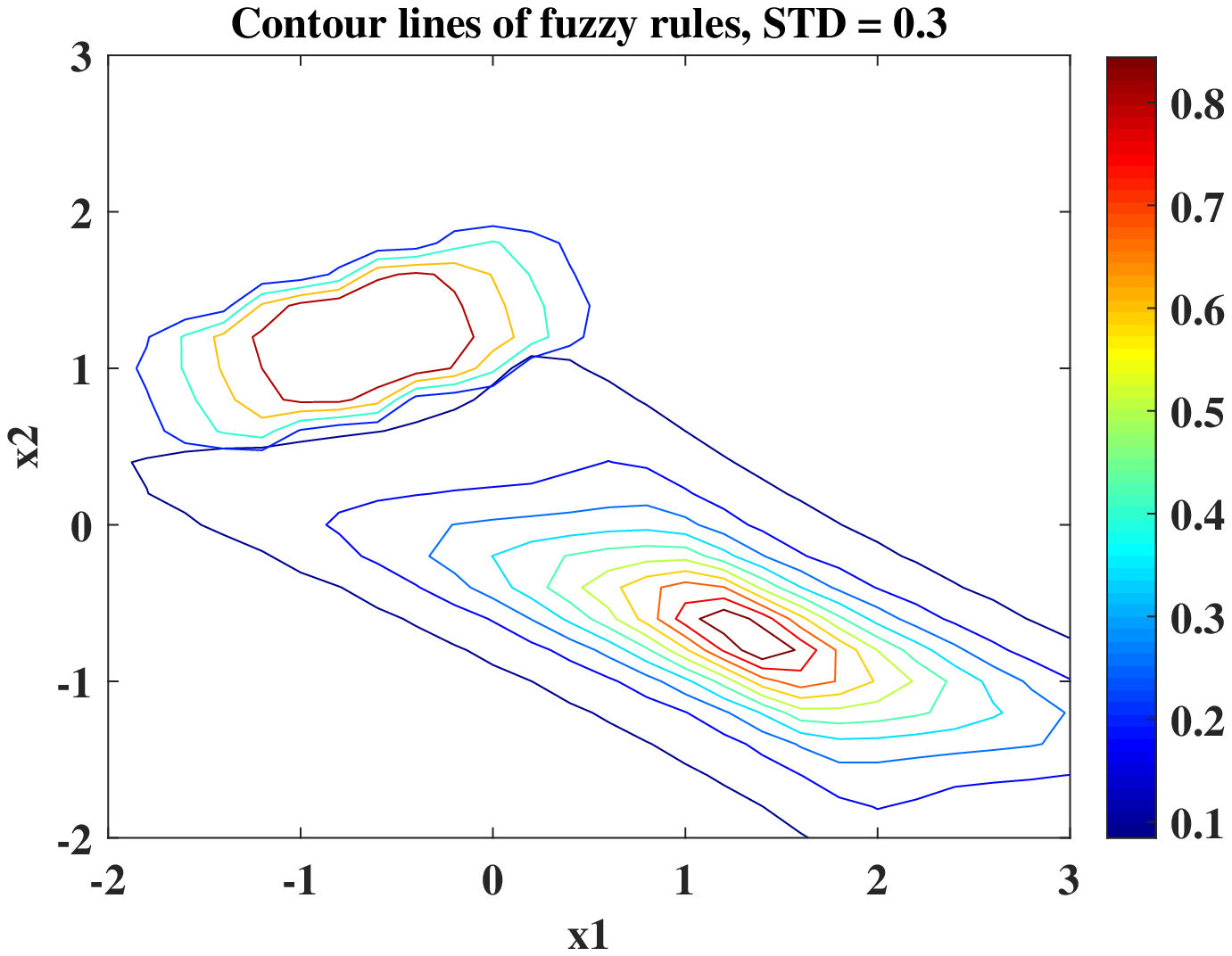}} \\
    % \subfigure[][]{\includegraphics[width = 1.5in]{rules_std_4.eps}} &
    %\subfigure[][]{\includegraphics[width = 2in]{rules_std_5.eps}} \\
    \subfigure[][]{\includegraphics[width = 2in]{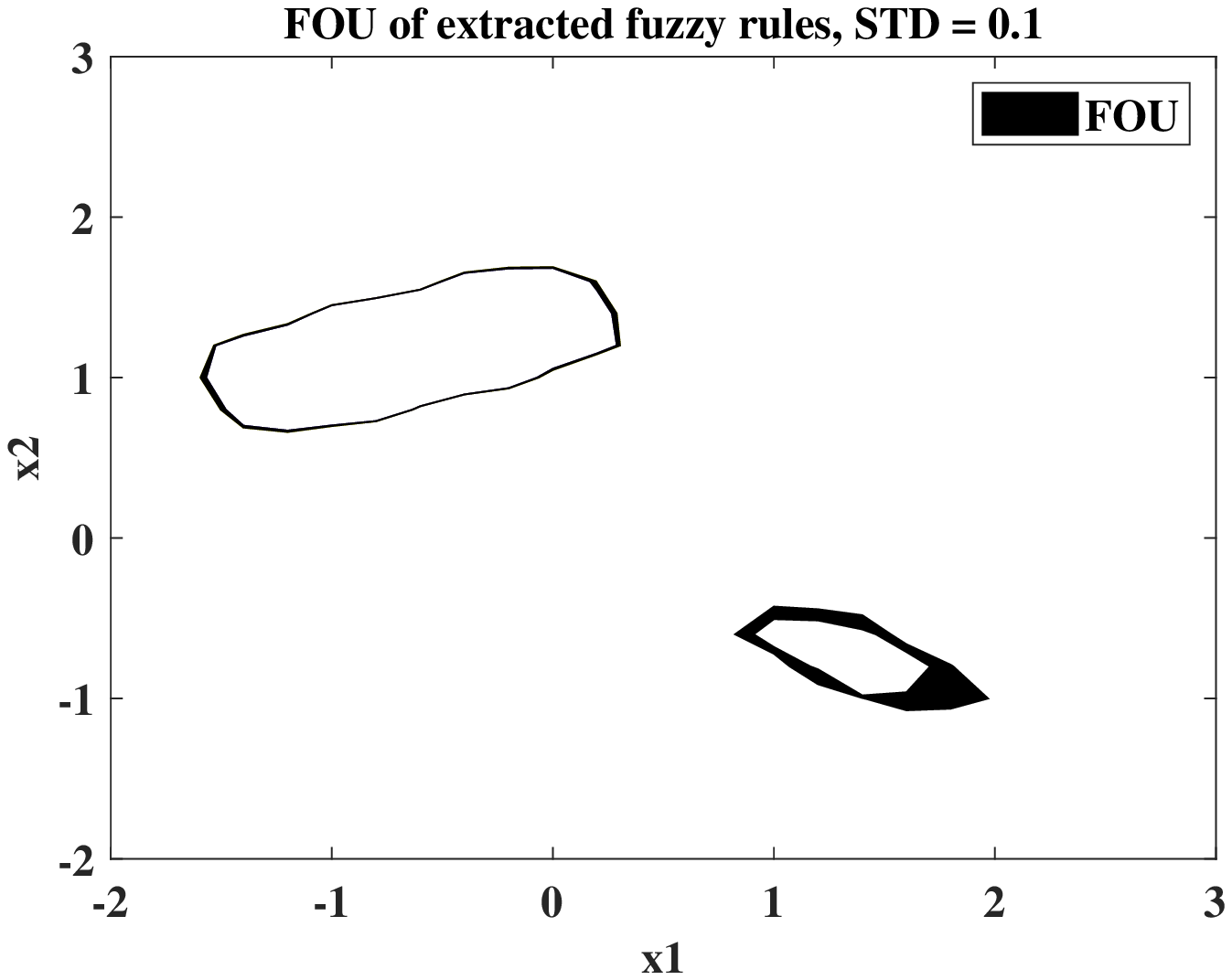}}&
    \subfigure[][]{\includegraphics[width = 2in]{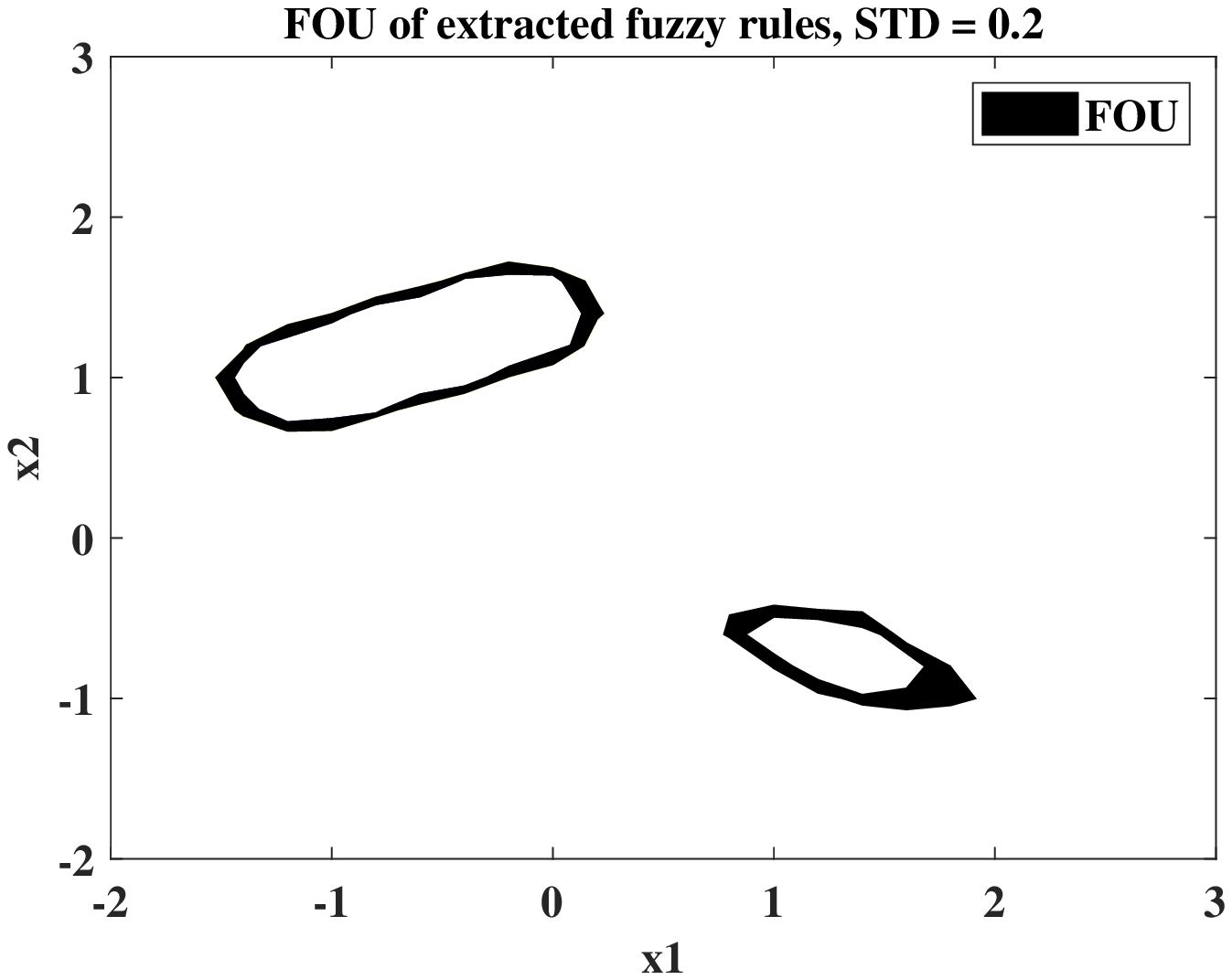}}&
    \subfigure[][]{\includegraphics[width = 2in]{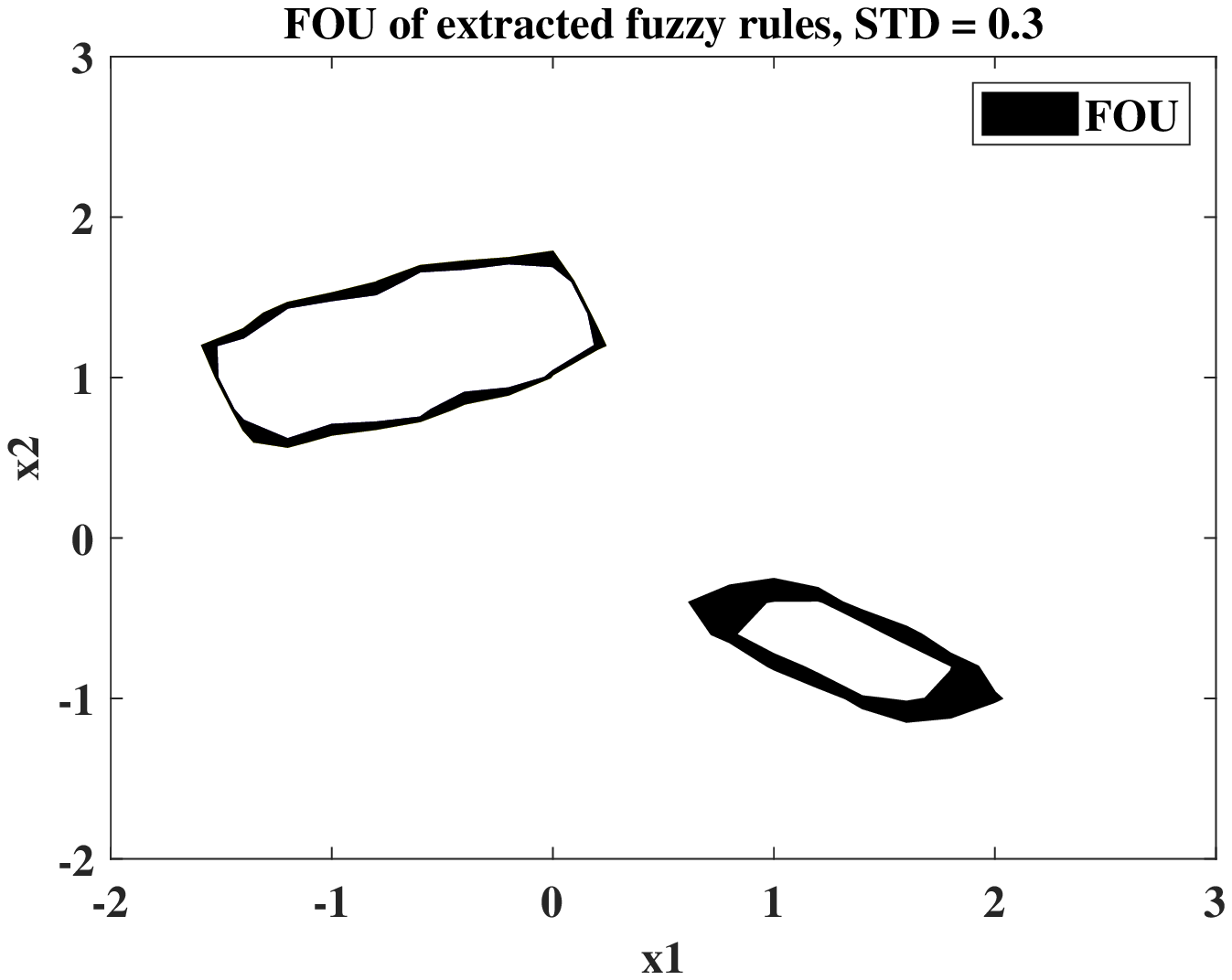}}
   % \subfigure[][]{\includegraphics[width = 1.5in]{fou_std_4.eps}}&
    %\subfigure[][]{\includegraphics[width = 2in]{fou_std_5.eps}}
\end{tabular}
\caption{Comparing contours of the target function (a) with contours of the extracted fuzzy rules trained with different levels of noise (b-d) and their footprint of uncertainty (FOU) (e-g). Only contours of one level for FOU are shown.}
\label{fig_exp2}
\end{figure}

Totally 700 samples are chosen uniformly and the number of fuzzy rules is set to two. For training 350 samples are chosen randomly and the rest used as the test data. Gaussian noise with mean equal to zero and different values of standard deviation (STD) is added to the generated data to investigate the effect of uncertainty in training data on forming interval type-2 fuzzy rules. Figure \ref{fig_exp2} compares the shape of extracted fuzzy rules' contours with the contours of the target function and shows the effect of amount of uncertainty on the footprint of uncertainty (FOU). Based on the results in Figure \ref{fig_exp2}, the extracted fuzzy rules are similar to the covered regions. Moreover, the size of FOU expands by increasing the amount of noise STD. Furthermore, the performance of the proposed model decreases by increasing the amount of noise STD, but the method is able to learn the target function efficiently.

\begin{figure}[!t]
\centering
\begin{tabular}{ccc}
    \multicolumn{3}{c}{\subfigure[][]{\includegraphics[width = 2in]{target_func.eps}}}\\
    \subfigure[][]{\includegraphics[width = 2in]{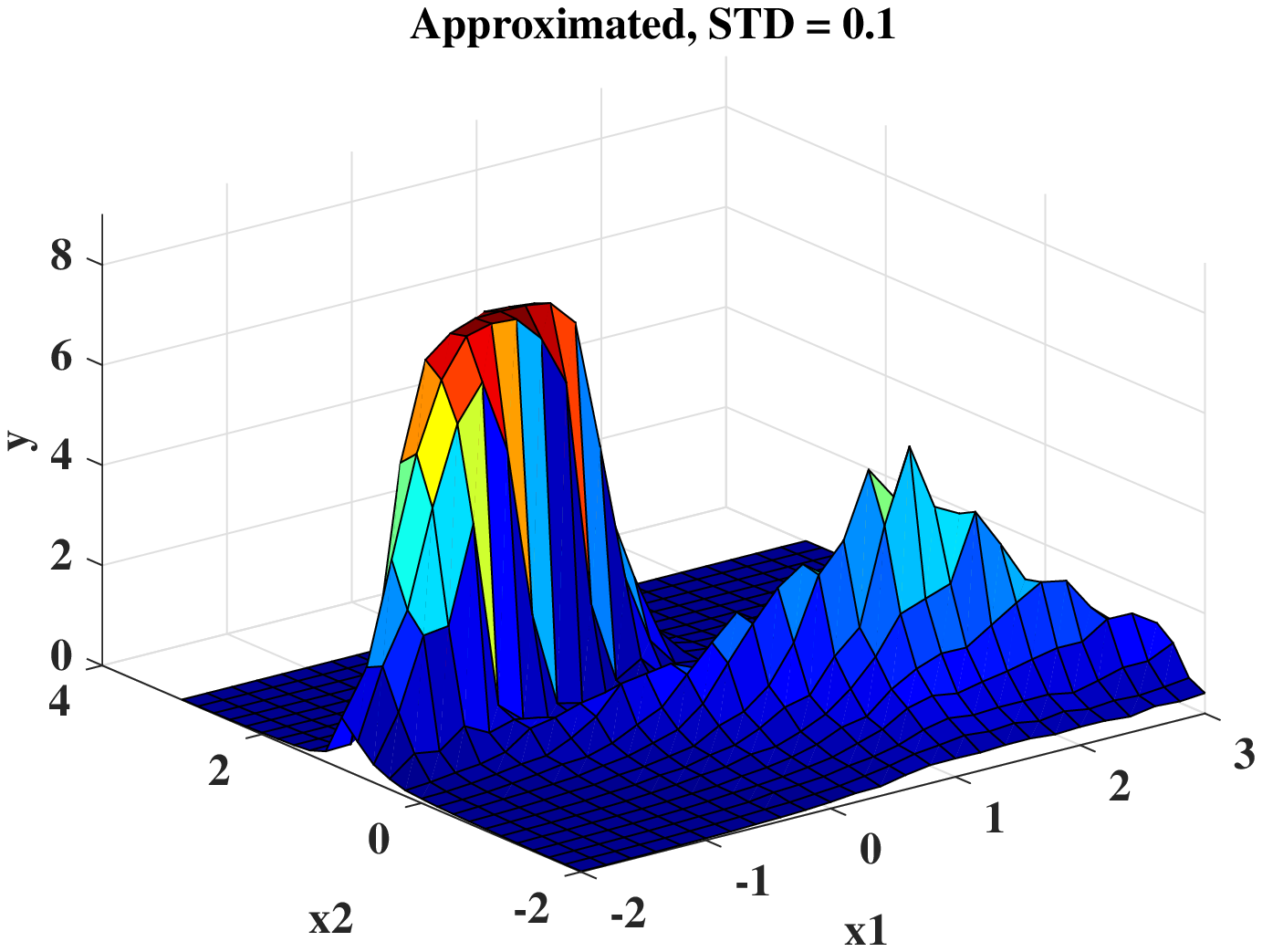}} &
    \subfigure[][]{\includegraphics[width = 2in]{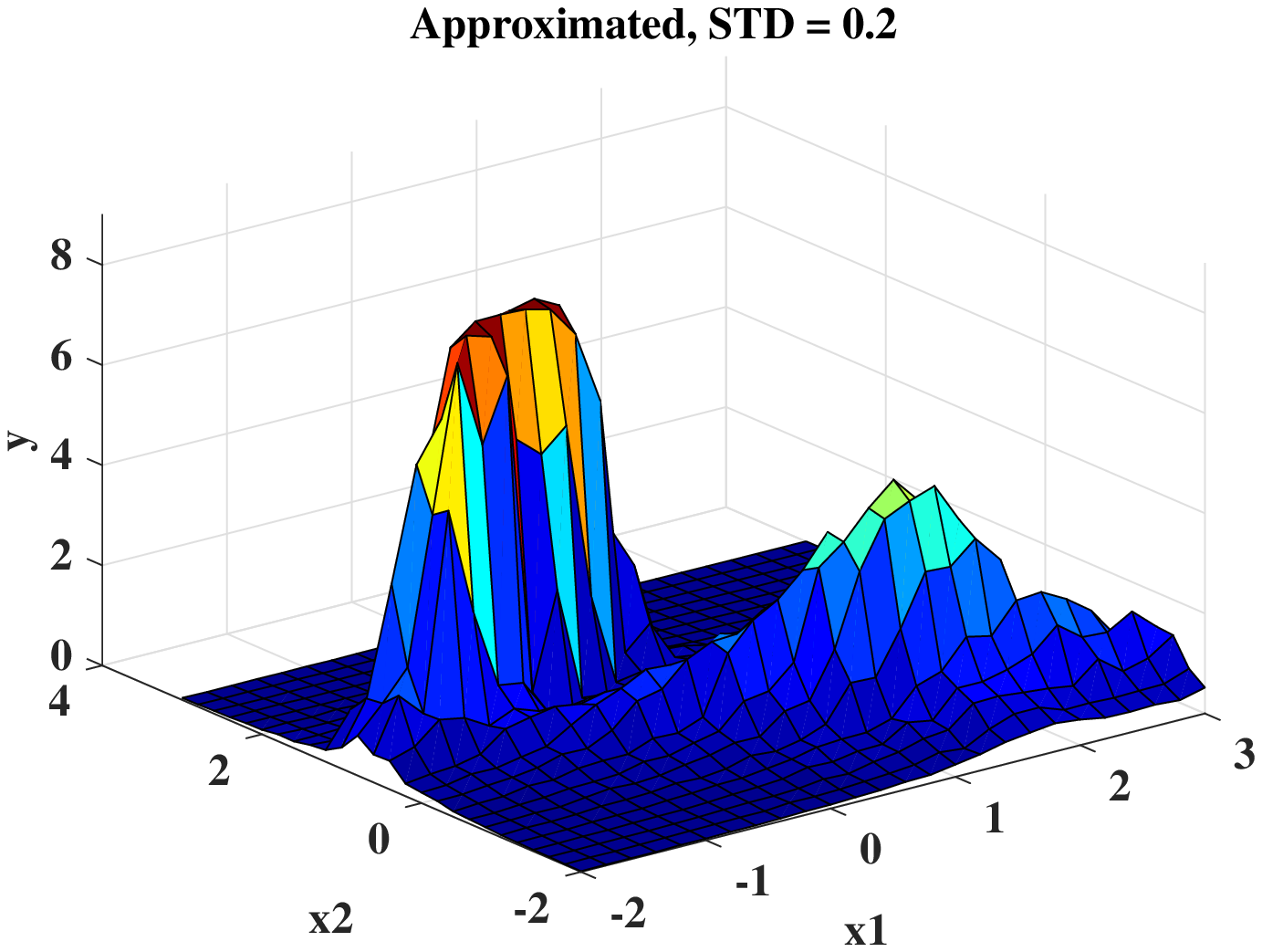}} &
    \subfigure[][]{\includegraphics[width = 2in]{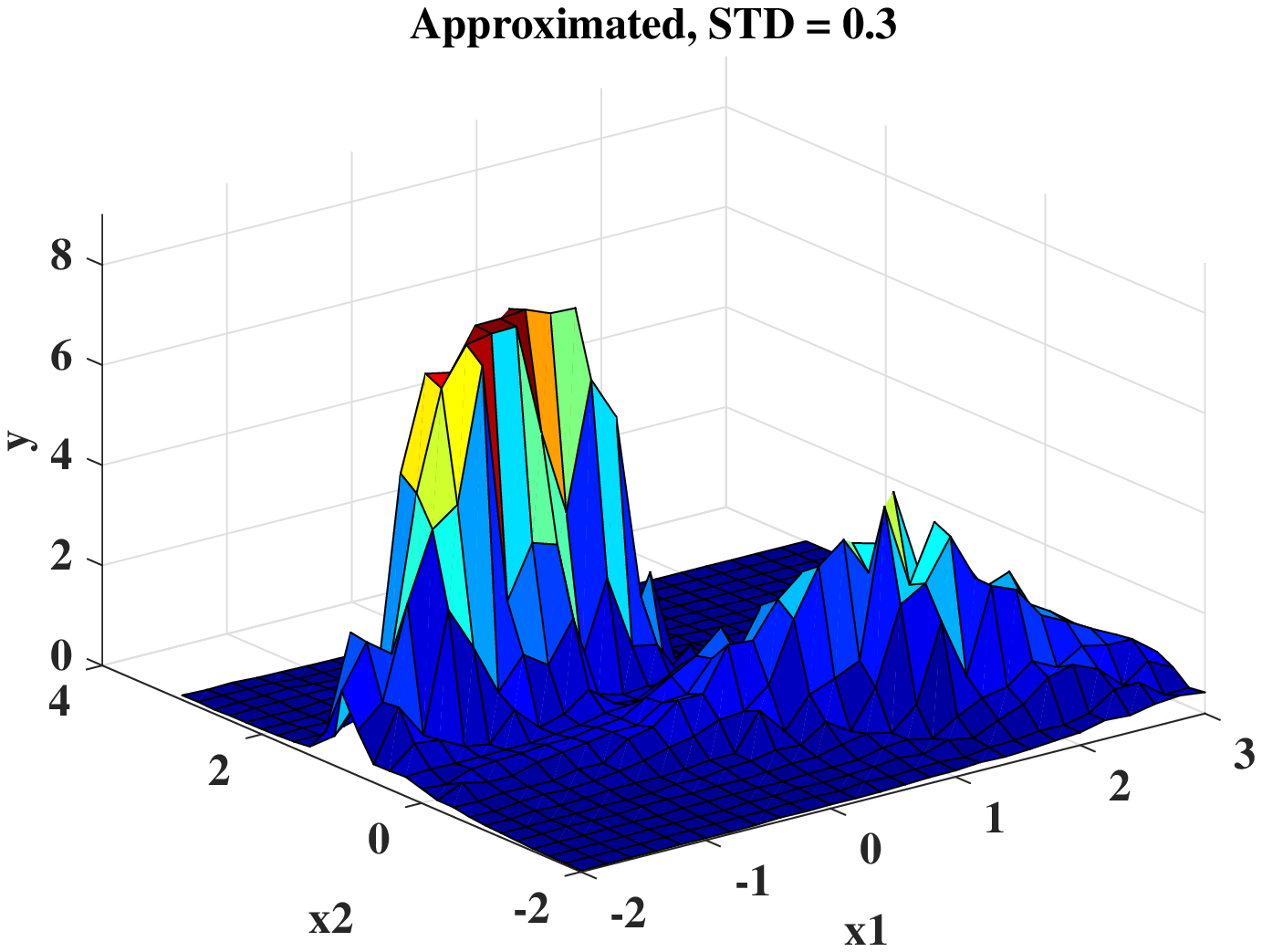}}
    % \subfigure[][]{\includegraphics[width = 1.5in]{rules_std_4.eps}} &
    %\subfigure[][]{\includegraphics[width = 2in]{rules_std_5.eps}} \\
   % \subfigure[][]{\includegraphics[width = 1.5in]{fou_std_4.eps}}&
    %\subfigure[][]{\includegraphics[width = 2in]{fou_std_5.eps}}
\end{tabular}
\caption{Comparing target function (a) with network's outputs trained with different levels of noise (b-d).}
\label{fig_exp3}
\end{figure}

Figure \ref{fig_exp3} compares the outputs of the network, trained in different levels of noise, with the target function. The same level of noise is added to both training and test samples. It is shown that the performance of the network in presence of noise remains agreeable.

\begin{figure}[t]
\centering
\includegraphics[width = 6.5in]{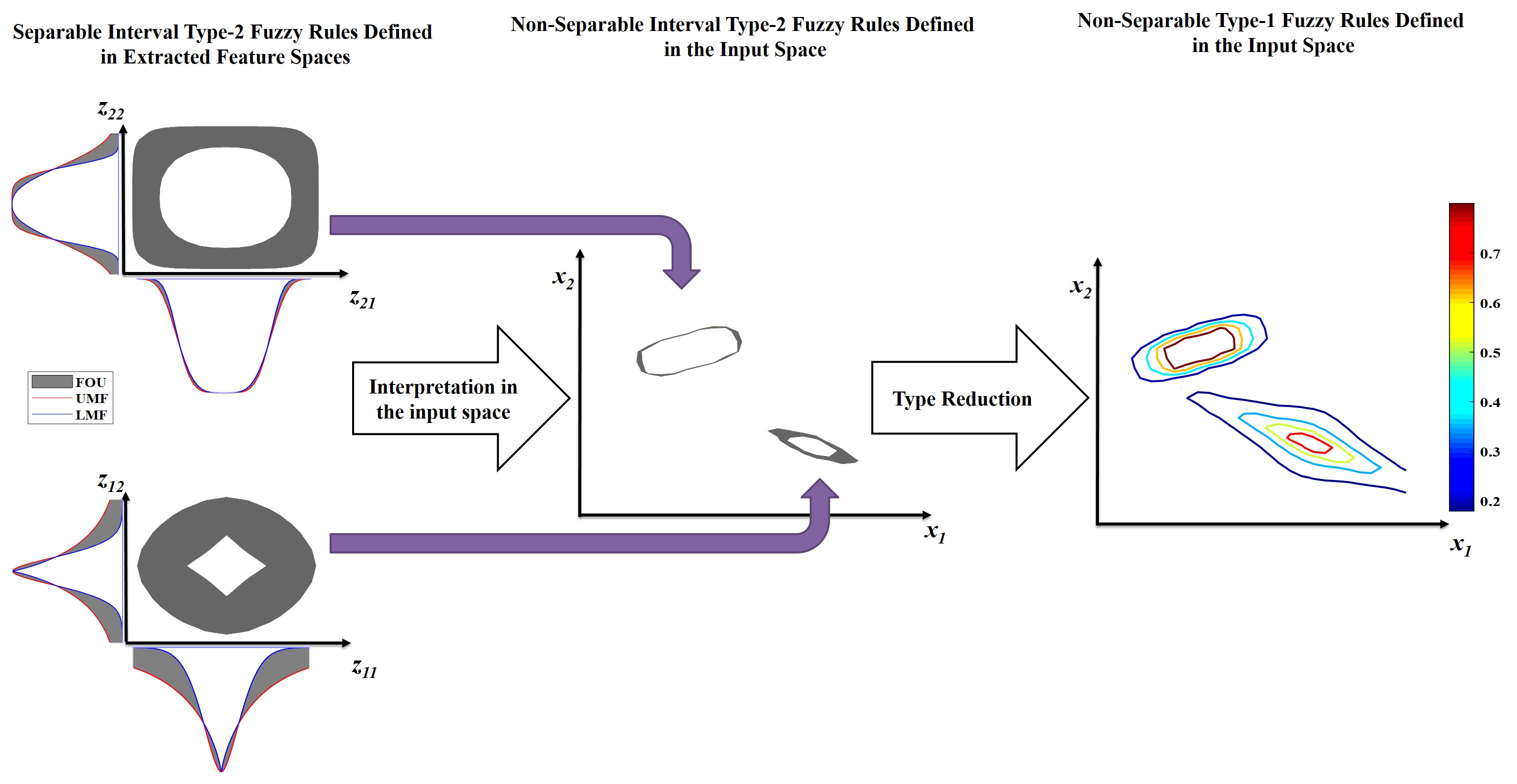}
\caption{Forming interval type-2 non-separable fuzzy rules and their type-1 counterparts in the interactive input space of experiment 1.}
\label{fig_sum}
\end{figure}

Finally, Figure \ref{fig_sum} summarizes the whole process of building fuzzy rules in this example. First, the input space is transformed to new feature spaces learned for defining each fuzzy rule properly. Next, interval type-2 separable fuzzy rules are formed in these new feature spaces based on extension principle, using interval type-2 fuzzy sets defined in these feature spaces. The separable fuzzy rules in these new feature spaces have non-separable interpretations in the initial interactive input space. By applying type reduction, type-1 fuzzy rules similar to the covered regions are extracted.

\subsection{Experiment 2: Noisy Mackey-Glass Time-Series Prediction}
\label{section32}
In this experiment the ability of the proposed method encountering noisy data is evaluated and compared with the performance of the other interval type-2 fuzzy neural networks in the literature. In this experiment, the proposed method is applied to the \textit{Mackey-Glass chaotic time-series prediction} problem which is a classic benchmark problem used in the literature \cite{Ebadzadeh15, Das15, IT2FNN-SVR, BAKLOUTI2018,CASTRO20092175,SOFMLS,GPFNN,Eyoh2018,LUO2019}. This benchmark data are generated using the following differential equation \cite{Ebadzadeh15,IT2FNN-SVR,Das15,BAKLOUTI2018,Eyoh2018}:
\begin{equation}\label{eq_mcglass}
  \frac{dx}{dt} = \frac{0.2x(t-\tau)}{1+x^{10}(t-\tau)}-0.1x(t)
\end{equation}
where $\tau > 16.5$. The aim of this problem is to predict $x(t)$ based on previous samples $\{x(t),x(t-\Delta t), \dots, x(t-n\Delta t)\}$ \cite{Ebadzadeh15,IT2FNN-SVR}. Following \cite{Ebadzadeh15,Das15,IT2FNN-SVR}, we set $\Delta t = 6$, and $n=4$. Therefore, the problem is to approximate function $f$ as follows \cite{Ebadzadeh15,Das15,IT2FNN-SVR}:
\begin{equation}\label{eq_mcglass2}
  x(t) = f\left(x(t-6),x(t-12), x(t-18), x(t-24)\right)
\end{equation}
To follow previous studies \cite{LUO2019,BAKLOUTI2018}, the initial condition $x_0$ and delay parameter $\tau$ are considered as $1.2$ and $17$, respectively.

To train the proposed network, 1000 samples are generated from $t = 124$ to $t = 1123$. For training first 500 samples are used and the remaining 500 samples are utilized for testing \cite{LUO2019,BAKLOUTI2018,Das15,IT2FNN-SVR}. Following previous studies \cite{LUO2019,BAKLOUTI2018,Das15,IT2FNN-SVR}, to evaluate the performance of the proposed network encountering noisy data, Gaussian noise with mean equal to zero and different values of standard deviation (STD) is added to the generated data.

In Table \ref{table-mcglass} the performance of the proposed network is compared with some other methods including a correlation-aware type-1 fuzzy neural network (\textit{IC-FNN} \cite{Ebadzadeh2017}), an interval type-2 fuzzy neural network using Gaussian membership function (\textit{GIT2FNN} \cite{BAKLOUTI2018}), an interval type-2 fuzzy neural network with $beta$ basis function able to from fuzzy rules with different shapes (\textit{BIT2FNN} \cite{BAKLOUTI2018}), a recurrent interval type-2 fuzzy neural network (\textit{eRIT2IFNN} \cite{LUO2019}), facing different levels of noise in training and test data, based on the number of fuzzy rules and the RMSE of the test data. The initial trust region $\lambda_0$ is set 1, and for rate of changing it ($\eta$ in equation (\ref{eq37})) 1.001 is chosen.

\begin{figure}[t]
\centering
\begin{tabular}{ccc}
    \subfigure[][]{\includegraphics[width = 2in]{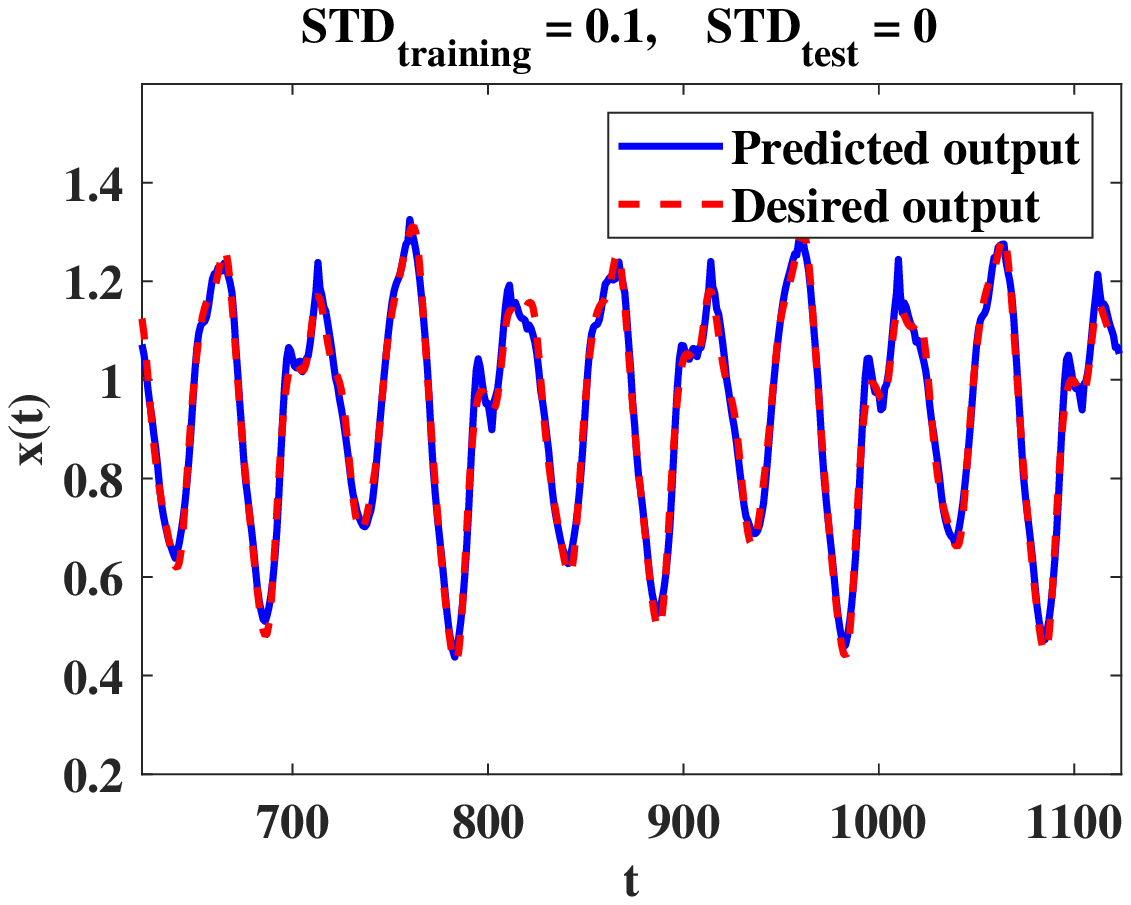}} &
    \subfigure[][]{\includegraphics[width = 2in]{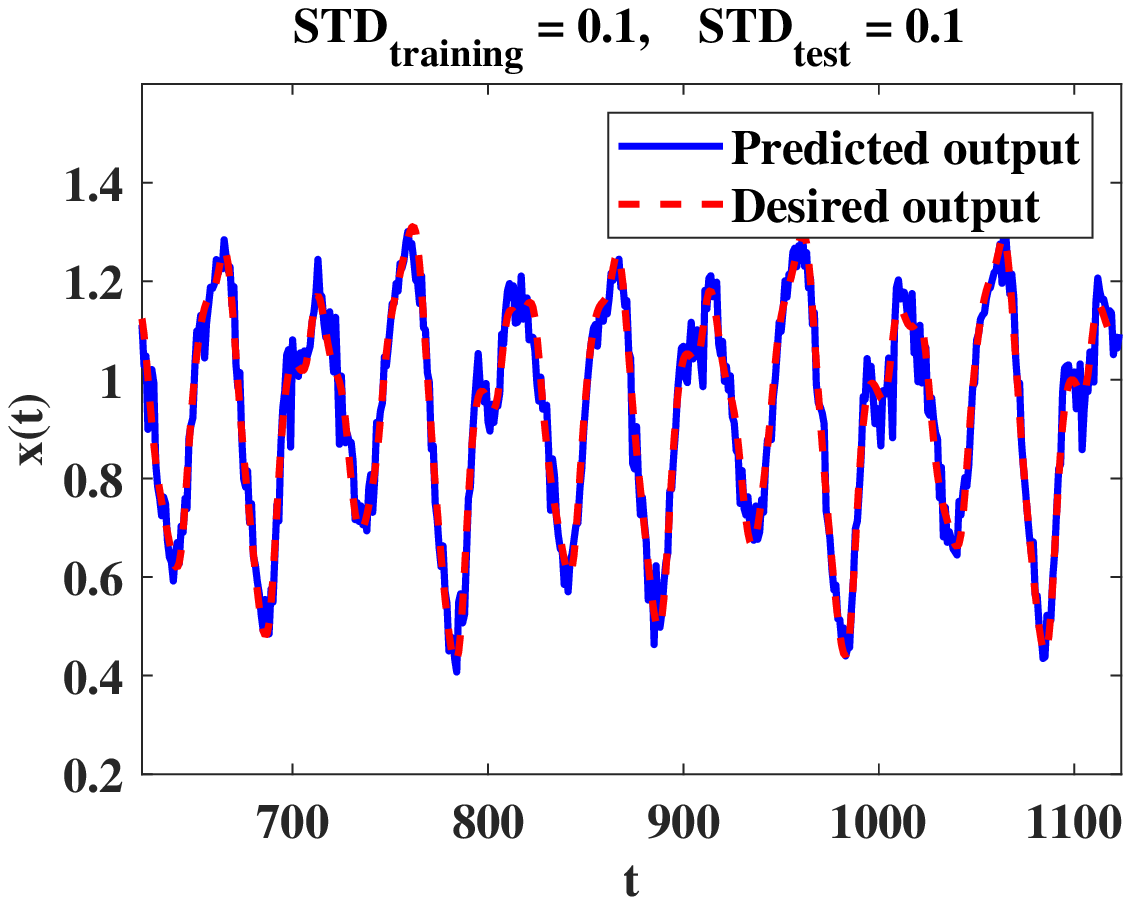}} &
    \subfigure[][]{\includegraphics[width = 2in]{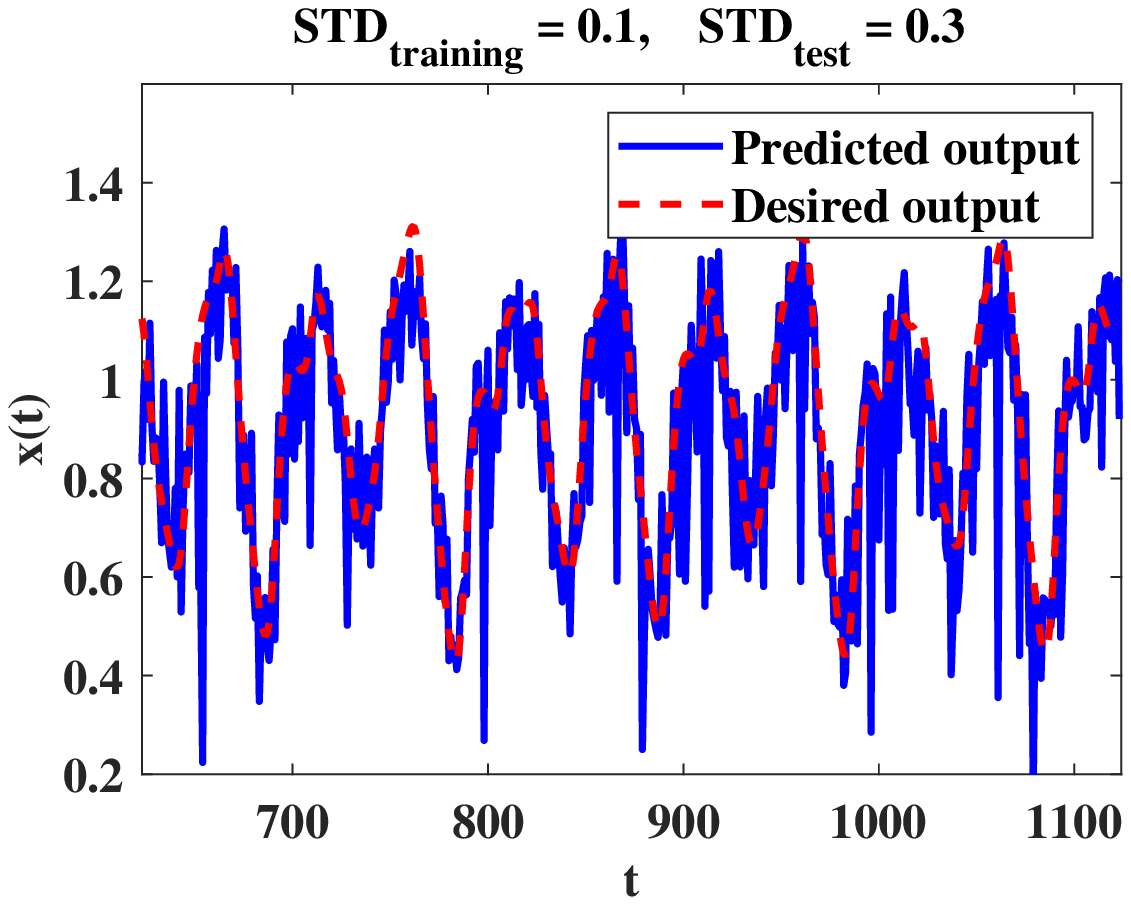}} \\
    \subfigure[][]{\includegraphics[width = 2in]{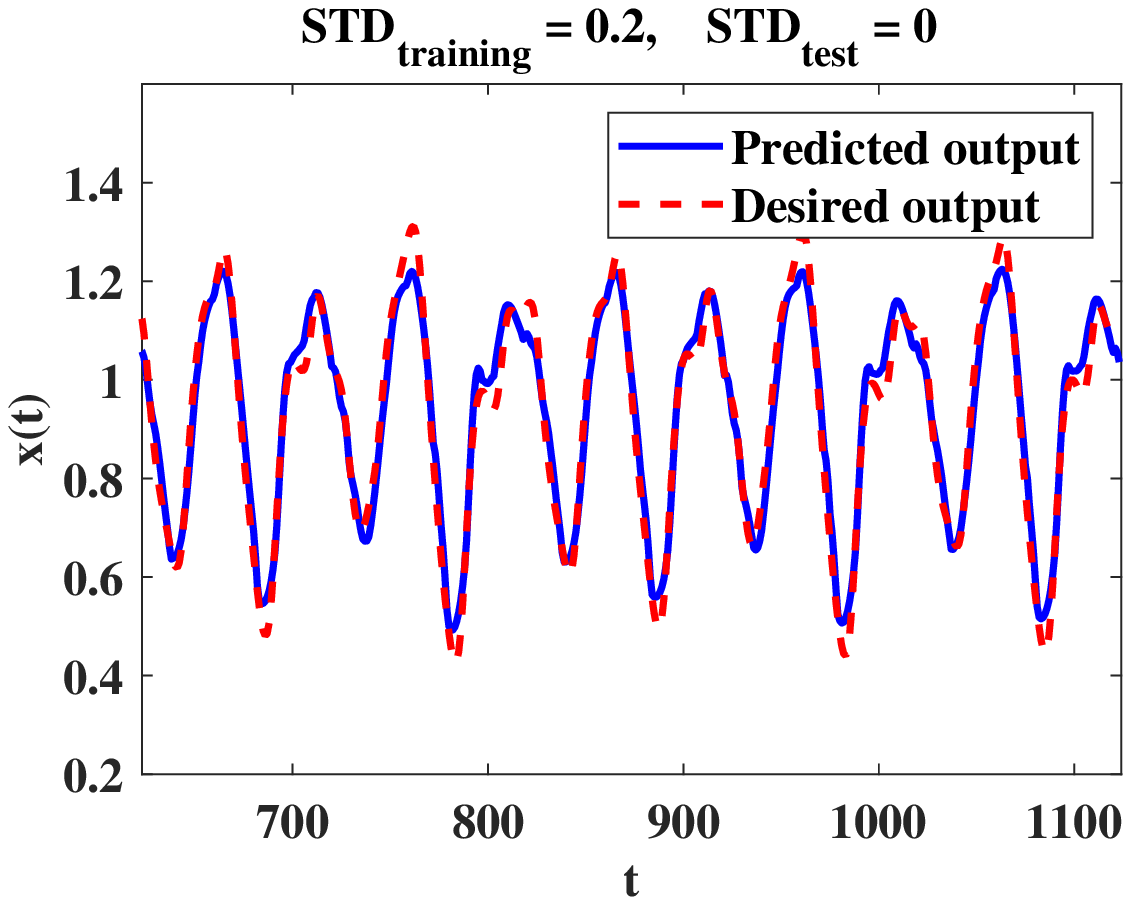}} &
    \subfigure[][]{\includegraphics[width = 2in]{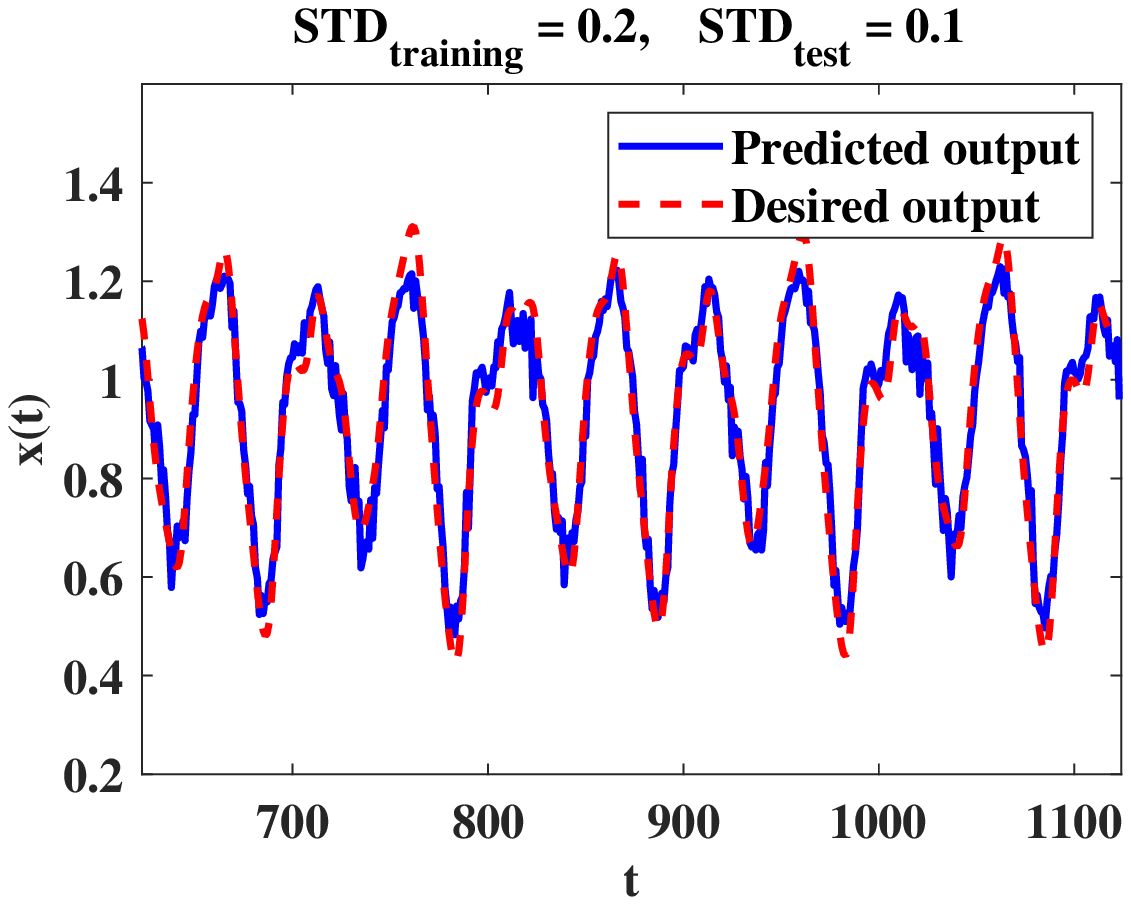}} &
    \subfigure[][]{\includegraphics[width = 2in]{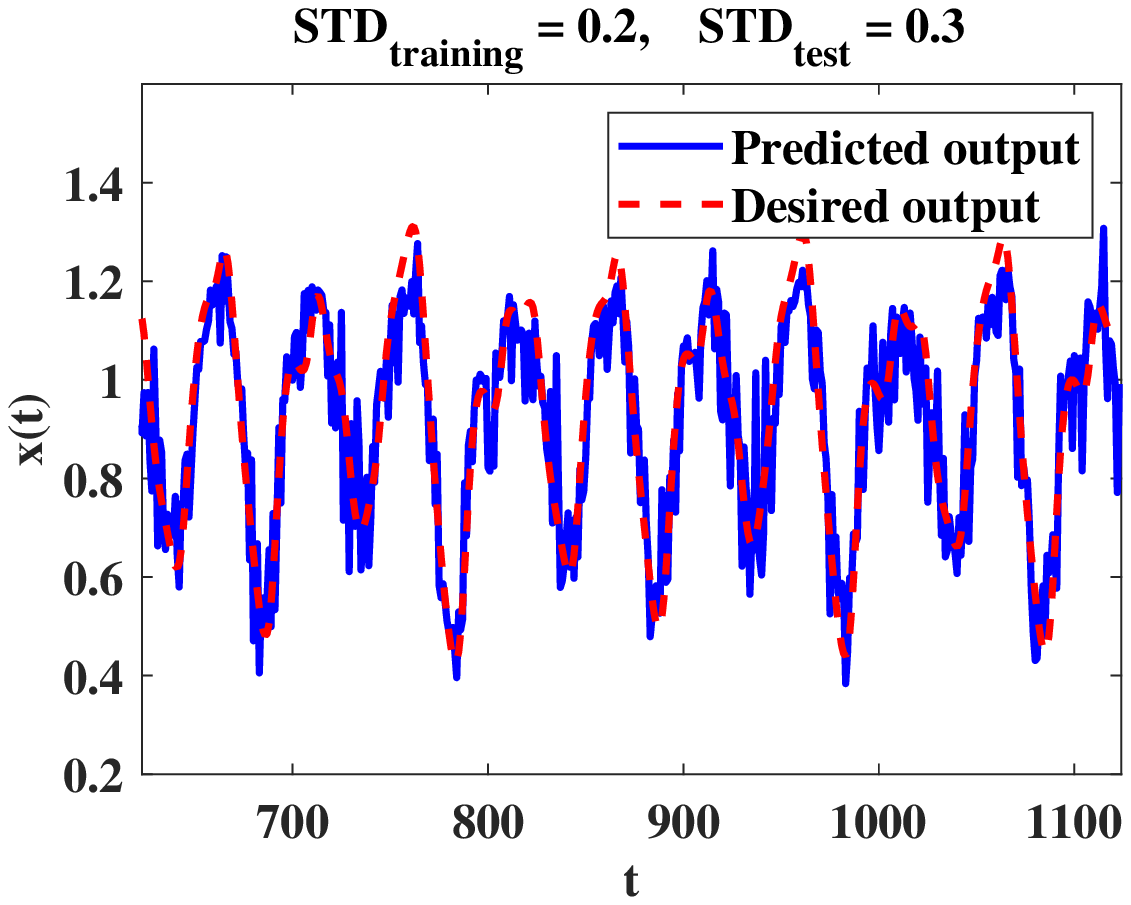}} \\
    \subfigure[][]{\includegraphics[width = 2in]{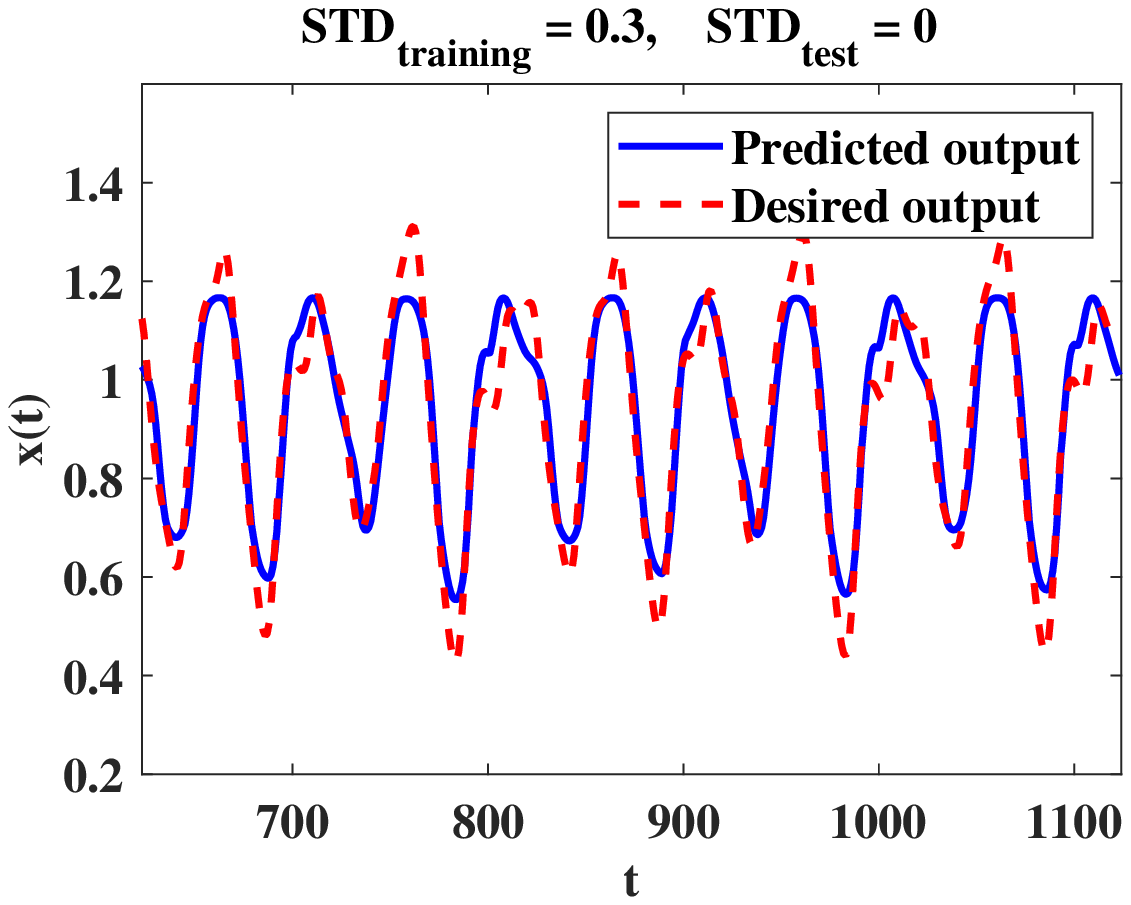}} &
    \subfigure[][]{\includegraphics[width = 2in]{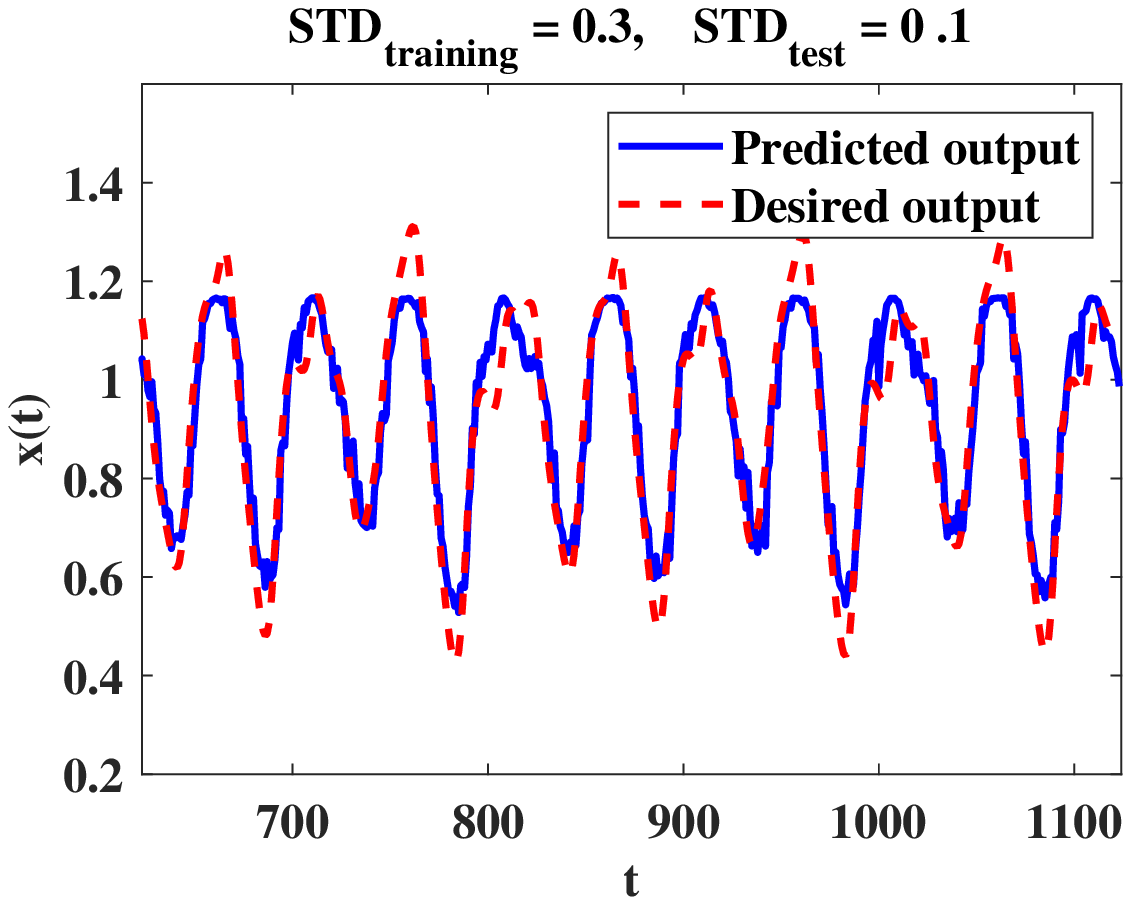}} &
    \subfigure[][]{\includegraphics[width = 2in]{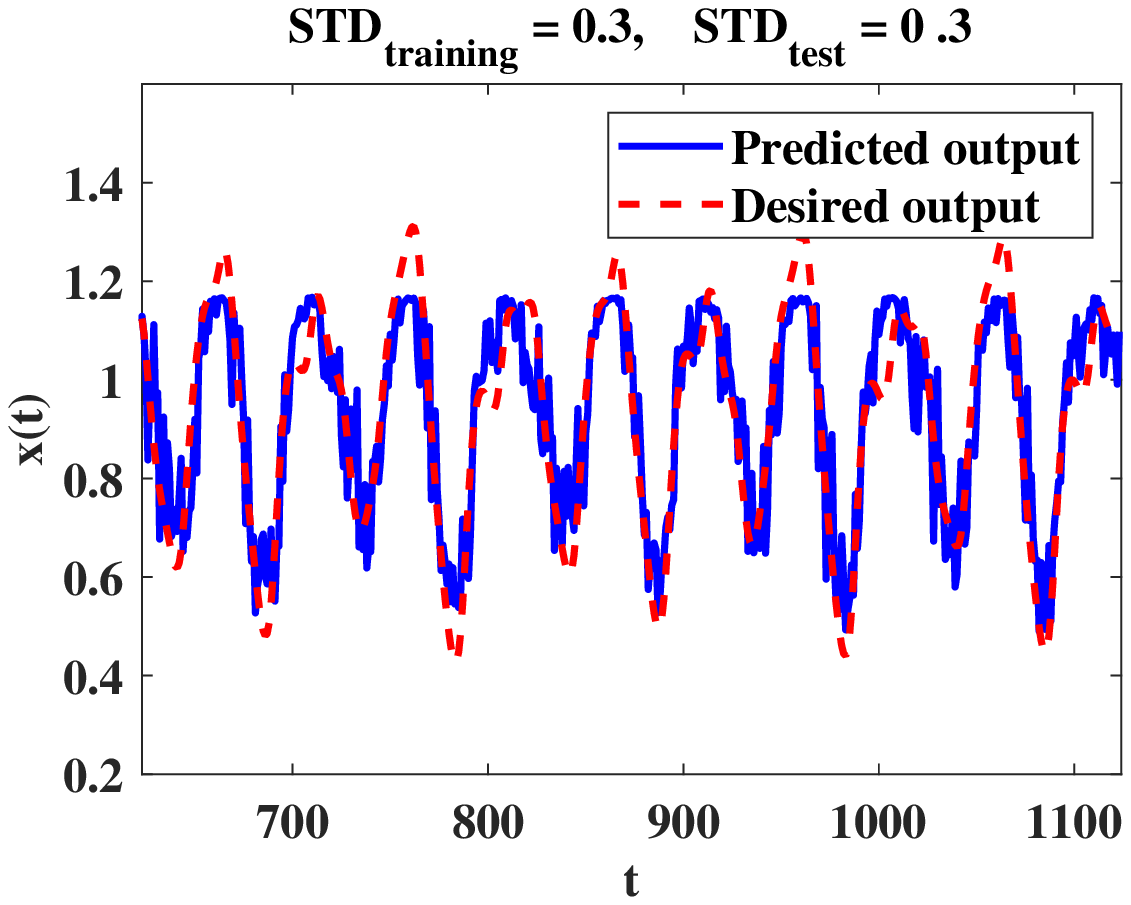}} \\
\end{tabular}
\caption{Comparison between the proposed network's outputs and desired outputs for Mackey–Glass chaotic time-series prediction problem.}
\label{fig_mcglass}
\end{figure}

\begin{figure}[t]
  \centering
  \includegraphics[width=3in]{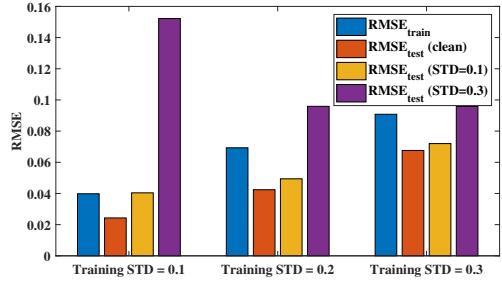}
  \caption{Comparison of training and test RMSE for different noise levels. It is shown that by increasing the level of noise added to the training data, the generalization ability of the method encountering noisy unseen data is improved. }\label{fig_mcglass2}
\end{figure}

Based on these results, the proposed method is robust encountering noisy training and test data. Furthermore, increasing the level of noise causes reducing the performance of the method. However, the performance of the method for noisy data remains agreeable (see Figure \ref{fig_mcglass} and \ref{fig_mcglass2}). Moreover, the performance of the proposed method is better than the previous studies with more parsimonious architecture (only two fuzzy rules) based on its correlation-aware nature. By comparing the proposed method with the type-1 correlation-aware structure, the advantage of type-2 for noisy data is shown. Furthermore, based on Figures \ref{fig_mcglass} and \ref{fig_mcglass2} the performance of the proposed network on test data remains higher and more reliable encountering the noisy data, in the situation of training on noisy data. Finally, Table \ref{table-mcglass2} compares the mean performance of the proposed method with some type-1 and interval type-2 fuzzy neural networks (the results of the other methods are reported from \cite{LUO2019,BAKLOUTI2018}). Based on these results, the proposed model has the best average performance.

\begin{table}[t]
\scriptsize
    \caption[c]{Predicting Mackey–Glass time-series with different noise level.}
    \centering
    \begin{tabular}{cccccccccccc}
        \hline
       \multicolumn{2}{c}{\textbf{Noise level (STD)}} &\multicolumn{2}{c}{\textbf{IT2CFNN}} & \multicolumn{2}{c}{\textbf{eRIT2IFNN} \cite{LUO2019}}& \multicolumn{2}{c}{\textbf{BIT2FNN} \cite{BAKLOUTI2018}}& \multicolumn{2}{c}{\textbf{GIT2FNN} \cite{BAKLOUTI2018}} & \multicolumn{2}{c}{\textbf{IC-FNN} \cite{Ebadzadeh2017}} \\
        \hline
        Train &  Test &  Rule &  RMSE & Rule &  RMSE &  Rule &  RMSE&Rule &  RMSE&Rule &  RMSE \\
        \hline
        0.1 & clean & \underline{\textbf{2}} & 0.0243 & 17 & 0.0441 & 16 & 0.0439 & 16 & 0.0981 & 2 & 0.1016\\
        \cline{2-12}
            & 0.1 & \underline{\textbf{2}} & 0.0404 & 17 & 0.0655 & 16 & 0.0821 & 16 & 0.1583 & 2 &  0.1144\\
         \cline{2-12}
            & 0.3 & \underline{\textbf{2}} & 0.1522 & 17 & 0.0943 & 16 & 0.1657 & 16 & 0.2145 & 2 & 0.2429 \\
         \hline
         0.2 & clean & \underline{\textbf{2}} & 0.0424 & 17 & 0.0617 & 16 & 0.0560 & 16 & 0.1325 & 2 & 0.0945 \\
         \cline{2-12}
         & 0.1 & \underline{\textbf{2}} & 0.0494 & 17 & 0.0805 & 16 & 0.0870 & 16 & 0.2025 & 2 & 0.0931 \\
         \cline{2-12}
         & 0.3 & \underline{\textbf{2}} & 0.0959 & 17 & 0.1033 & 16 & 0.1800 & 16 & 0.2984 & 2 &  0.1334\\
         \hline
         0.3 & clean & \underline{\textbf{2}} & 0.0676 & 17 & 0.0726 & 16 & 0.0920 & 16 & 0.1836 & 2 & 0.1693 \\
         \cline{2-12}
         & 0.1 & \underline{\textbf{2}} & 0.0720 & 17 & 0.0861 & 16 & 0.1270 & 16 & 0.2871 & 2 & 0.1663 \\
         \cline{2-12}
         & 0.3 & \underline{\textbf{2}} & 0.0958 & 17 & 0.1042 & 16 & 0.1820 & 16 & 0.3411 & 2 & 0.1706 \\
         \hline
    \end{tabular}
    \label{table-mcglass}
\end{table}

\begin{table}[t]
\scriptsize
    \caption[c]{Performance comparison of different models for Mackey–Glass chaotic prediction
problem with noise.}
    \centering
    \begin{tabular}{ccccc}
        \hline
       \textbf{Type} &\textbf{Algorithm} & \multicolumn{3}{c}{\textbf{RMSE (train with STD)}}\\
       & & 0.1 & 0.2 & 0.3\\
        \hline
         &  SONFIN \cite{SONFIN} &  0.139 &  0.178 & 0.236 \\
        & DENFIS \cite{DENFIS} & 0.147 & 0.161 & 0.175 \\
       Type-1 & EFuNN \cite{EFuNN} & 0.143 & 0.195 & 0.257 \\
        & ANFIS \cite{ANFIS} & 0.130 & 0.249 & 0.330\\
        & IC-FNN \cite{Ebadzadeh2017} & 0.153 & 0.107 & 0.168\\
        \hline
         & SEIT2FNN \cite{SEIT2FNN}& 0.119& 0.141&0.216 \\
         & IT2FNN-SVR \cite{IT2FNN-SVR}& 0.109& 0.125&0.151 \\
          & eT2FIS \cite{eT2FIS}& 0.127& 0.154&0.177 \\
          Type-2 & McIT2FIS-UM \cite{Das15}& 0.104& 0.117&0.126 \\
            & McIT2FIS-US \cite{Das15}& 0.104& 0.110&0.128 \\
             & GIT2FNN \cite{BAKLOUTI2018}& 0.157 & 0.211&0.270 \\
             & BIT2FNN \cite{BAKLOUTI2018}& 0.097& 0.107&0.133 \\
        &\textbf{IT2CFNN} & 0.072 & 0.062 & 0.078\\
        \hline
    \end{tabular}
    \label{table-mcglass2}
\end{table}

\subsection{Experiment 3: Santa-Fe Chaotic Laser Time Series}
\label{section331}
\begin{figure}[t]
  \centering
  \includegraphics[width=3in]{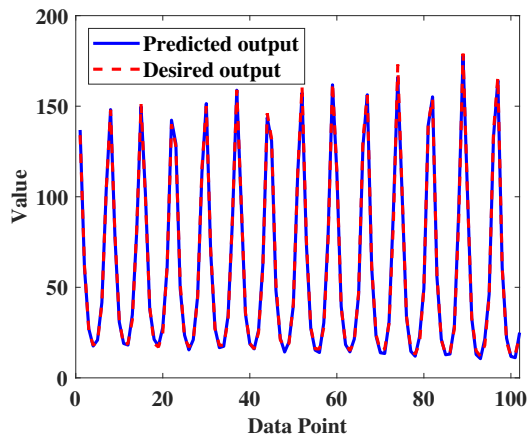}
  \caption{Comparison between the proposed network's outputs and desired outputs for Santa-Fe chaotic laser time-series prediction problem.}\label{fig-laser}
\end{figure}

The series-A from the Santa-Fe time series competition\footnote{http://www-psych.stanford.edu/~andreas/Time-Series/SantaFe.html}, recorded from a Far-Infrared-Laser in a chaotic state, is a well-known real-world time series prediction problem used in some previous studies \cite{Ebadzadeh15, Ebadzadeh2017,Eyoh2018,IT2FNN-SVR}. The purpose of this problem is to predict the current state based on a window including five steps ago. Indeed, the aim of this problem is to learn function $f$ as: $y_t = f(y_{t-1},y_{t-2},y_{t-3},y_{t-4},y_{t-5})$ \cite{Eyoh2018,IT2FNN-SVR}. Samples are normalized to the interval $[0,1]$ following \cite{IT2FNN-SVR,Eyoh2018}. Dataset composed of 1000 samples that first 900 samples are used as training dataset and the remaining 100 points are used for testing. The initial trust region $\lambda_0$ is set 1, and for rate of changing it ($\eta$ in equation (\ref{eq37})) 1.001 is chosen. Figure \ref{fig-laser} shows the network's prediction along with the desired sequence. Table \ref{table-laser} compares the performance and architecture size of the proposed method with other previous models.

\begin{table}[t]
\scriptsize
    \caption[c]{Comparison of the performance of the proposed method with previous methods in Santa-Fe Laser time-series prediction problem.}
    \centering
    \begin{tabular}{ccccc}
        \hline
       \textbf{Method} & \textbf{Number of rules} & \textbf{Number of parameters} & \textbf{Training RMSE} &\textbf{Testing RMSE}\\
       \hline
       SONFIN \cite{SONFIN} & 9 & 144 & 6.956 & 5.983 \\
       SEIT2FNN \cite{SEIT2FNN} & 5 & 135 & 7.677 & 5.766 \\
       IT2FNN-SVR(N) \cite{IT2FNN-SVR} & 5 & 106 & 13.565 & 4.337 \\
       IT2FNN-SVR(F) \cite{IT2FNN-SVR} & 5 & 106 & 9.094 & 3.474 \\
        IT2 IFLS-DEKF+GD \cite{Eyoh2018} & 32 & 414 & 6.075 &1.668 \\
        \textbf{IT2CFNN} & 3 & 123 & 7.106 &1.924 \\
        \hline
    \end{tabular}
    \label{table-laser}
\end{table}

Moreover, to study the effect of noise on the performance of the proposed method similar to experiments performed in \cite{IT2FNN-SVR}, a Gaussian noise with mean equal to zero and standard deviation (STD) equal to 0.05 is added to the training data samples. After training based on the noisy data, the performance of the method is evaluated on clean test data, and noisy test data (STD = 0.01 and STD = 0.03) \cite{IT2FNN-SVR}. Table \ref{laser-noise} compares the performance of the method with some other previous models in presence of noise. Based on these results, the proposed network has the best performance in presence of noise with lower number of rules.

\begin{table}[t]
\scriptsize
    \caption[c]{Comparison of the performance of the proposed method with previous methods in Santa-Fe Laser time-series prediction problem.}
    \centering
    \begin{tabular}{cccccccc}
        \hline
       \textbf{Method} & \textbf{Number of rules} & \textbf{Number of parameters} & \textbf{Training RMSE} &\multicolumn{4}{c}{\textbf{Testing RMSE}}\\
       \hline
        & & & STD = 0.05 &Clean&STD = 0.01&STD = 0.03&STD=0.05\\
        \hline
       SONFIN \cite{SONFIN} & 8 & 128 & 19.41 & 7.99 & 8.51 & 11.63 & 15.57 \\
       SEIT2FNN \cite{SEIT2FNN} & 5 & 135 & 19.43 & 6.93 & 7.93 & 10.54 & 14.41 \\
       IT2FNN-SVR(N) \cite{IT2FNN-SVR} & 5 & 106 & 21.12 & 6.99 & 7.27 & 9.64 & 12.33 \\
       IT2FNN-SVR(F) \cite{IT2FNN-SVR}& 5 & 106 & 20.90 & 6.68 & 6.84 & 9.29 & 12.51 \\
        \textbf{IT2CFNN} & 3 & 123 & 16.05 & 3.45 & 3.87 & 8.13 & 10.99 \\
        \hline
    \end{tabular}
    \label{laser-noise}
\end{table}

\subsection{Experiment 4: Box-Jenkins Gas Furnace Problem}
\label{section33}

Box-Jenkins gas furnace system identification \cite{Box90} is a well-known problem which frequently used for performance evaluation in the literature \cite{box,box2,Das15,Eyoh2018,LUO2019,SOFMLS,ABDOLLAHZADE2015107,anh2019interval}. The purpose in this benchmark problem us to predict the $CO_2$ concentration (indicated as y) based on the gas flow rate (indicated by u) \cite{Das15,ABDOLLAHZADE2015107,Eyoh2018,LUO2019,anh2019interval}. The problem is defined in different versions \cite{box,SOFMLS,Das15}. For ease of comparison with previous studies, the version defined in \cite{SOFMLS,Eyoh2018,ABDOLLAHZADE2015107} is considered. The dataset consists of 296 instances and the aim is to predict the $CO_2$ concentration at time step $t$ based on the input methane flow rate at time step $(t-4)$ and the amount of $CO_2$ produced at the previous time step $(t-1)$. Indeed, the purpose of this problem is to approximate function $f$ as follows: $y(t) = f(u(t-4),y(t-1))$. Considering the delay amount 4, the number of samples reduced to 292 instances. For training 200 samples are used and the remaining 92 instances are utilized as the test dataset. The initial trust region $\lambda_0$ is set 1, and for rate of changing it ($\eta$ in equation (\ref{eq37})) 1.001 is chosen.

\begin{figure}[t]
\centering
\includegraphics[width = 3in]{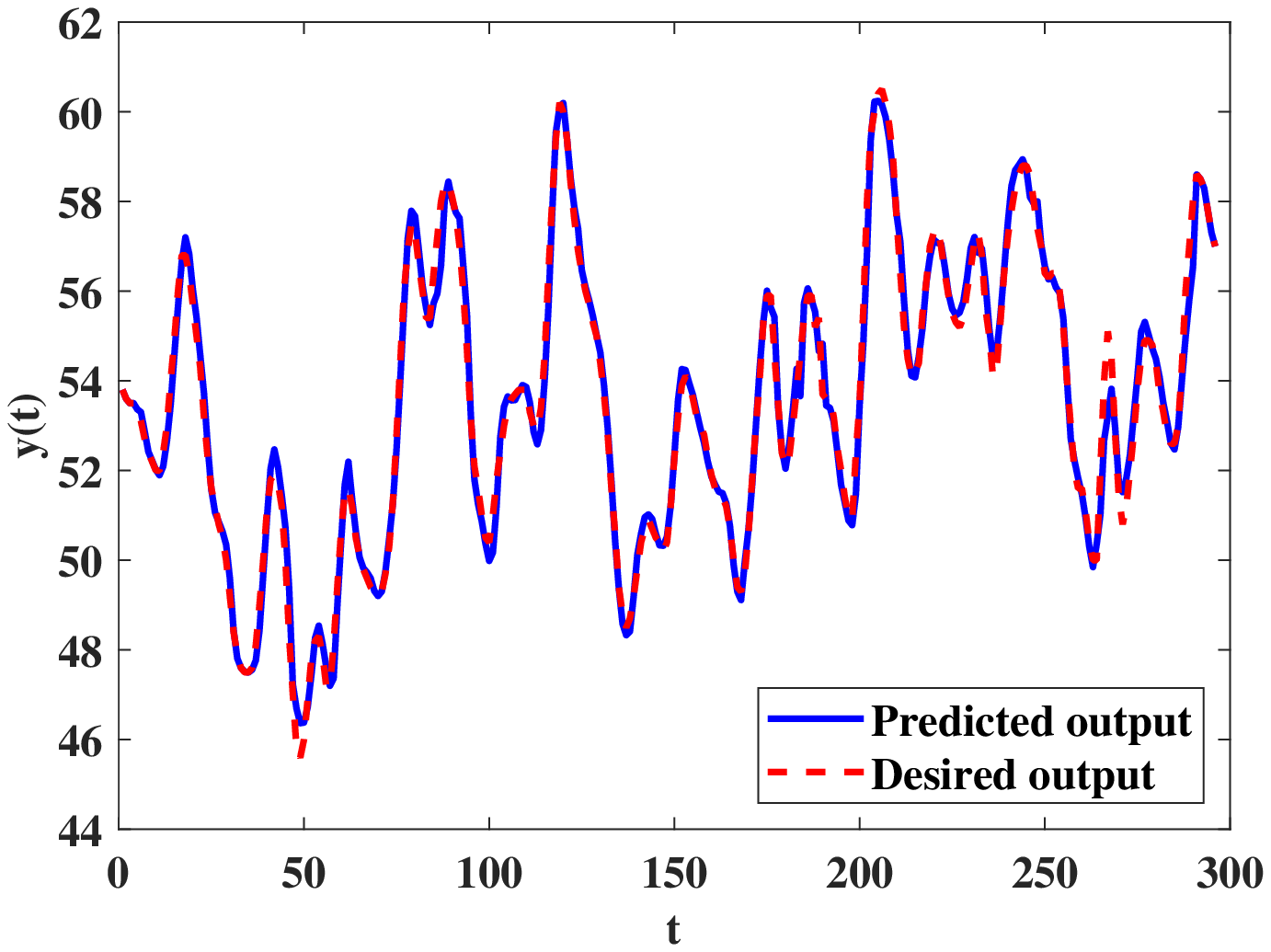}
\caption{Comparison between actual and predicted outputs for Box-Jenkins nonlinear system identification.}
\label{fig_box}
\end{figure}

Figure \ref{fig_box} shows the desired and approximated output. Table \ref{table_box} compares the performance of the method with the previous models. Based on the results reported in Table \ref{table_box}, the proposed network has a better performance than its type-1 counterpart (\textit{IC-FNN} \cite{Ebadzadeh2017}). Comparison with existing studies shows IT2FNN performing better than or comparatively with other works in the literature.

\begin{table}[t]
\scriptsize
    \caption[c]{Comparison of the performance of the proposed method with previous methods in Box-Jenkins system identification problem.}
    \centering
    \begin{tabular}{ccc}
        \hline
       \textbf{Method} & \textbf{Number of rules} & \textbf{Testing RMSE}\\
       \hline
        ARMA \cite{Box90} & -  & 0.8430 \\
        Tong's model \cite{Tong80} & 19  & 0.6850 \\
        Pedrycz's model \cite{Pedrycz84} & 81  & 0.5660 \\
        Xu's model \cite{Xu87} & 25  & 0.5730 \\
        Lee's model \cite{lee94} & 25  & 0.6380 \\
        Lin's model \cite{lin95} & 4  & 0.5110 \\
        Nie's model \cite{Nie95} & 45  & 0.4120 \\
        ANFIS \cite{ANFIS} & 4  & 0.0850 \\
        SAFIS \cite{safis} & 5  & 0.0710 \\
        eTS \cite{eTS} & 5 & 0.0490 \\
        SOFMLS \cite{SOFMLS} & 5 & 0.0470 \\
        McFIS \cite{mcfis} & 4 &  0.0360 \\
        IFLS-DEKF+GD \cite{Eyoh2018} & 4  & 0.0273  \\
        IT2 IFLS DEKF+GD \cite{Eyoh2018} & 4  & 0.0249  \\
        eRIT2IFNN \cite{LUO2019}  & 7  & 0.0197  \\
        SIT2FNN \cite{SIT2FNN} & 4  & 0.0560 \\
        SEIT2FNN \cite{SEIT2FNN} & 3  & 0.0574 \\
        LLNF \cite{ABDOLLAHZADE2015107} & -  & 0.0462 \\
        OSSA–LLNF \cite{ABDOLLAHZADE2015107} & -  & 0.0321 \\
        McIT2FIS-GD \cite{anh2019interval} & 3 & 0.0440\\
        McIT2FIS-PI \cite{anh2019interval} & 3 & 0.0420\\
        IC-FNN \cite{Ebadzadeh2017} & 3 & 0.0450 \\
        \textbf{IT2CFNN} & 3 & 0.0301 \\
        \hline
    \end{tabular}
    \label{table_box}
\end{table}

\subsection{Experiment 5: Electricity Load Forecasting}
\label{section34}
\begin{figure}[t]
\centering
\begin{tabular}{cc}
    \subfigure[][]{\includegraphics[width = 2in]{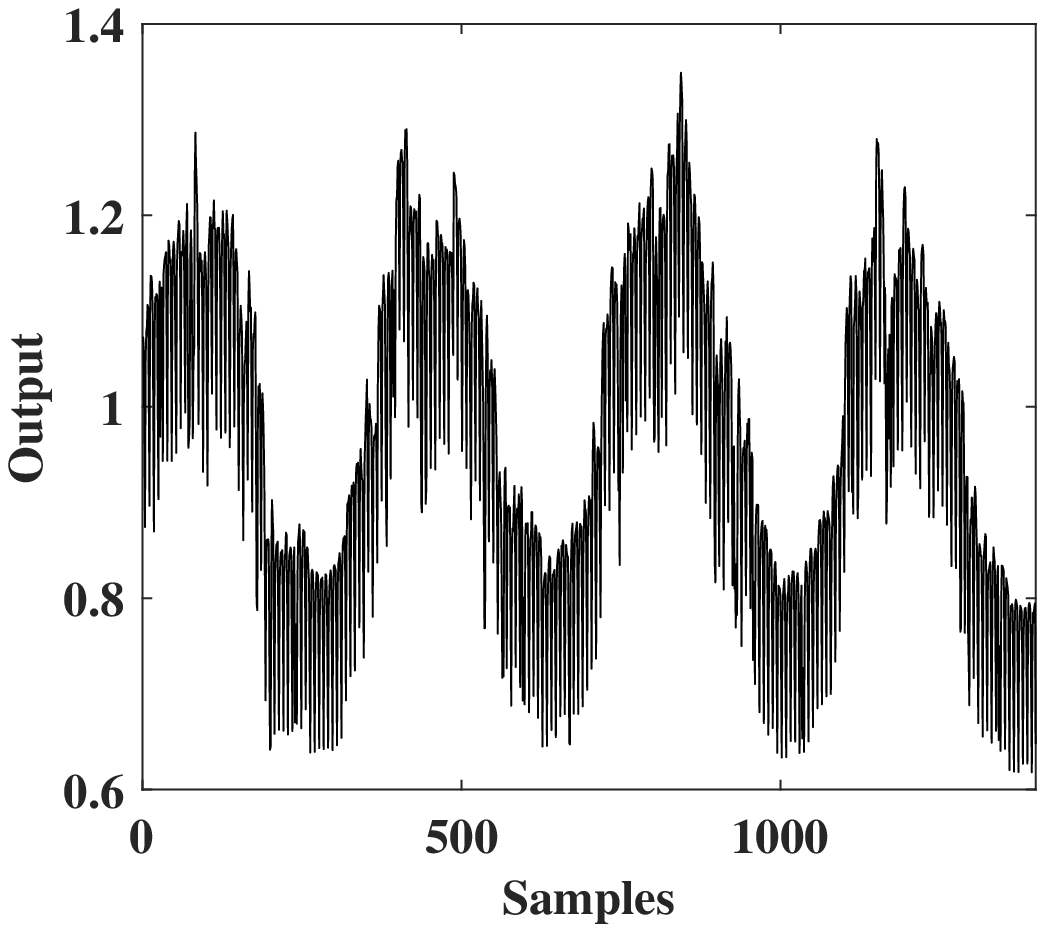}} &
    \subfigure[][]{\includegraphics[width = 2in]{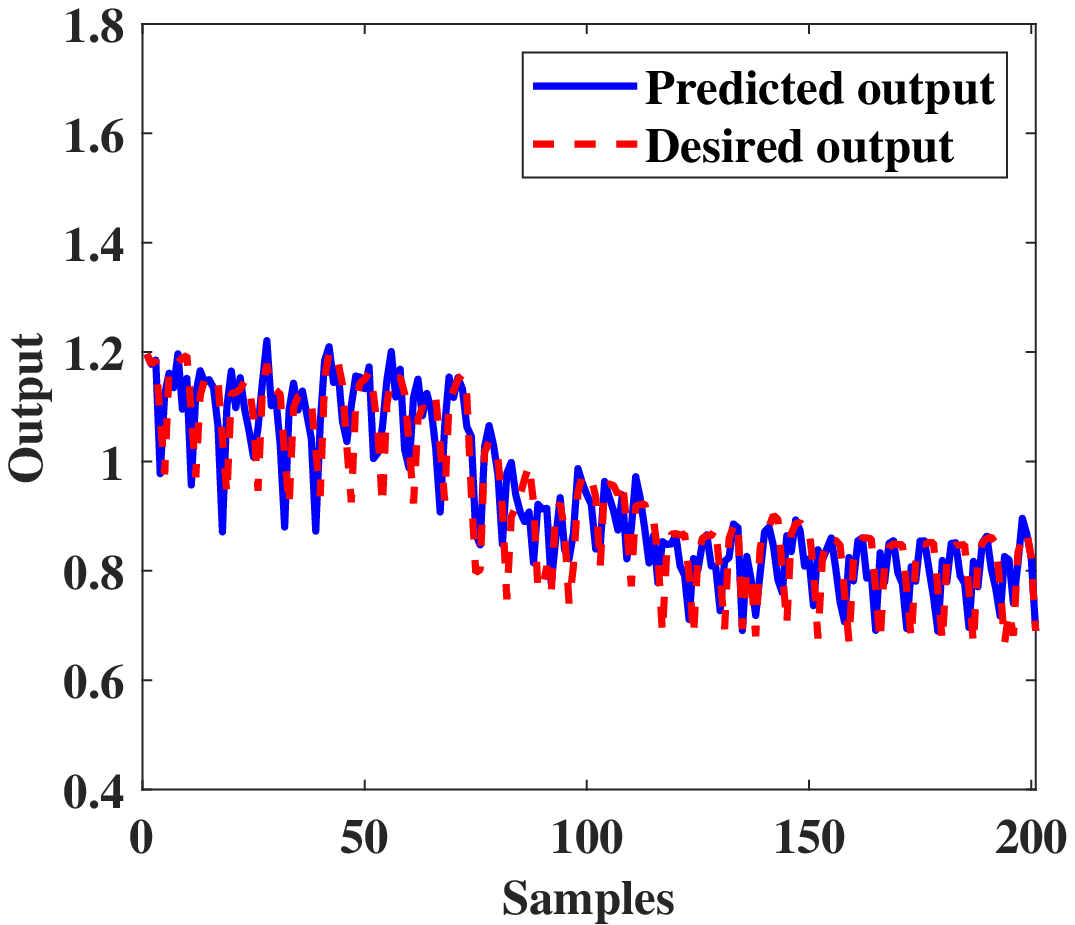}}
\end{tabular}
\caption{Training data for Poland electricity load forecasting (a) and the comparison between actual and predicted outputs for Poland electricity load forecasting (b).}
\label{fig_poland}
\end{figure}

The aim of this experiment is to perform a one-step-ahead prediction of electricity load values of Poland based on acquired data in the 1990's\footnote[1]{https://research.cs.aalto.fi/aml/datasets.shtml}. The training data contains 1400 samples and 201 instances are considered as the testing data. The training data is shown in Figure \ref{fig_poland}. The number of fuzzy rules is set 5, the initial trust region $\lambda_0$ is set 1, and for rate of changing it ($\eta$ in equation (\ref{eq37})) 1.001 is chosen. Table \ref{table_poland} compares the performance the proposed architecture with some other type-2 and type-1 models. Based on this results, the proposed network performs better than or comparatively with other works in both training and test. Moreover, Figure \ref{fig_poland} compares the desired and the network's predicted output.

\begin{table}[t]
\scriptsize
    \caption[c]{Comparison of the performance of the proposed method with previous methods in Poland electricity load forecasting.}
    \centering
    \begin{tabular}{ccc}
        \hline
       \textbf{Method} & \textbf{Testing RMSE} & \textbf{Testing RMSE}\\
       \hline
        IT2 FLS DEKF+GD \cite{Eyoh2018} & 0.0564  & 0.0595 \\
        IFLS DEKF+GD \cite{Eyoh2018} & 0.0589  & 0.0599  \\
        IT2 IFLS DEKF+GD \cite{Eyoh2018} & 0.0560  & 0.0572  \\
        \textbf{IT2CFNN} & 0.0562 & 0.0569 \\
        \hline
    \end{tabular}
    \label{table_poland}
\end{table}

\subsection{Experiment 6: Google Stock Price Tracking}
\label{section35}

\begin{figure}[t]
\centering
\begin{tabular}{cc}
    \subfigure[][]{\includegraphics[width = 2.25in]{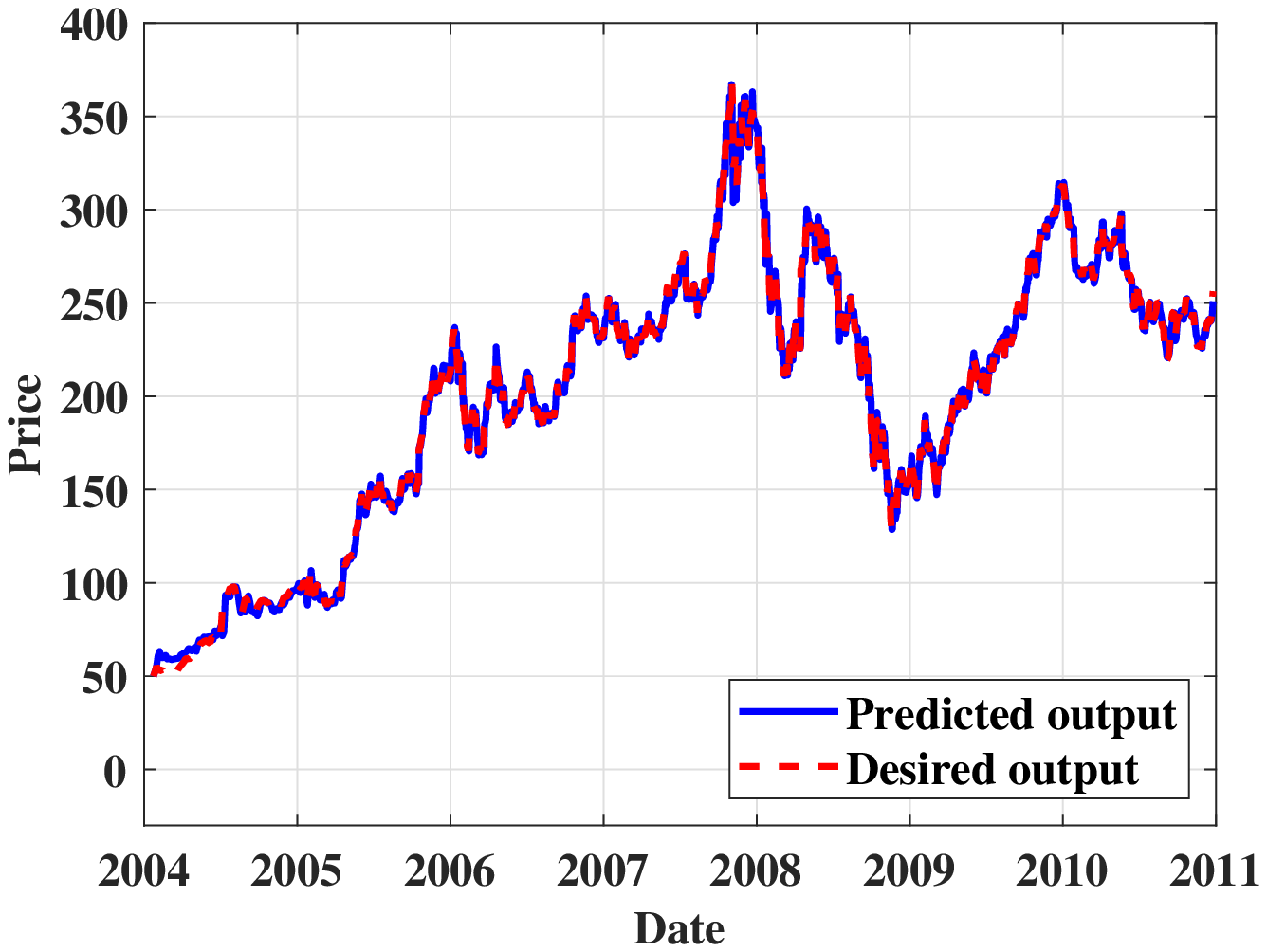}}&
    \subfigure[][]{\includegraphics[width = 2.25in]{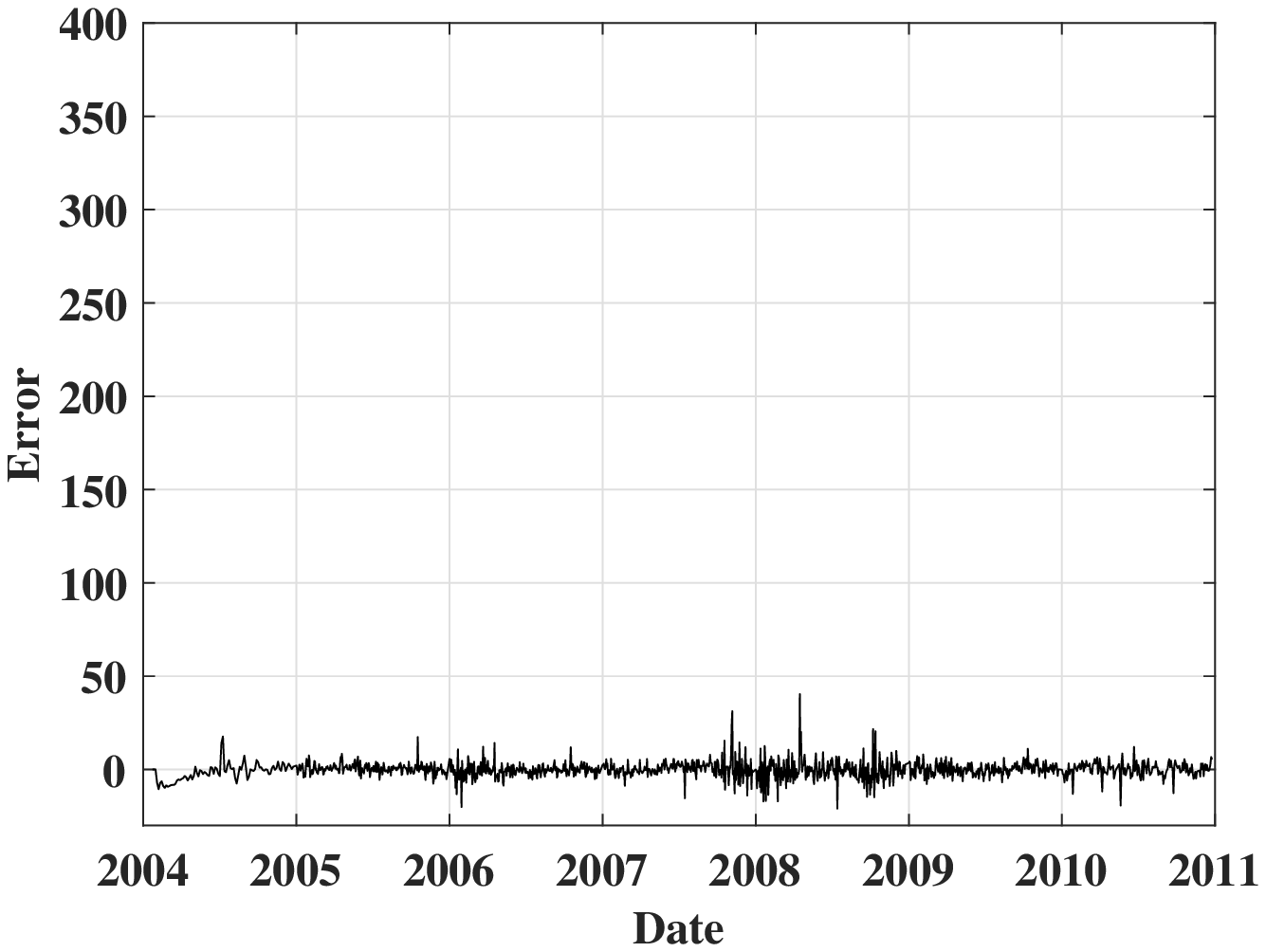}}
\end{tabular}
\caption{Comparison between actual and predicted outputs for Google stock price tracking (a) along with the error signal (b).}
\label{fig_google}
\end{figure}

In this experiment, the ability of the proposed network is studied for tracking a real-world financial dataset \cite{Das15,Ebadzadeh2017}. Data samples are obtained from Yahoo finance \footnote[1]{http://finance.yahoo.com/}. To follow the previous studies, 1534 samples of daily stock prices over six years from 19-August-2004 to 21-September-2010 are collected \cite{Das15,Ebadzadeh2017}. All samples are used for both training and testing. The initial trust region $\lambda_0$ is set 1, and for rate of changing it ($\eta$ in equation (\ref{eq37})) 1.001 is chosen. Table \ref{table_google} compares the performance of the proposed model with previous approaches. Based on the reported results, the proposed model has the best performance with the most parsimonious architecture. Figure \ref{fig_google} compares the target and predicted time-series along with the error.

\begin{table}[t]
\scriptsize
    \caption[c]{Comparison of the performance of the proposed method with previous methods in Google stock price tracking.}
    \centering
    \begin{tabular}{ccc}
        \hline
       \textbf{Method} & \textbf{Number of rules} & \textbf{RMSE}\\
       \hline
        McIT2FIS-UM \cite{Das15} & 7  & 11.10 \\
        McIT2FIS-UM \cite{Das15} & 11  & 9.90  \\
        eT2FIS \cite{eT2FIS} & 34  & 18.46  \\
        CFNN \cite{Ebadzadeh15} & 3  & 15.74  \\
        IC-FNN \cite{Ebadzadeh2017} & 3  & 6.38  \\
        PCA-FNN \cite{Ebadzadeh2017} & 3  & 65.01  \\
        \textbf{IT2CFNN} & 3 & 4.22 \\
        \hline
    \end{tabular}
    \label{table_google}
\end{table}

\subsection{Experiment 7: Sydney Stock Price Tracking}
\label{section36}

\begin{figure}[t]
\centering
\begin{tabular}{cc}
    \subfigure[][]{\includegraphics[width = 2.25in]{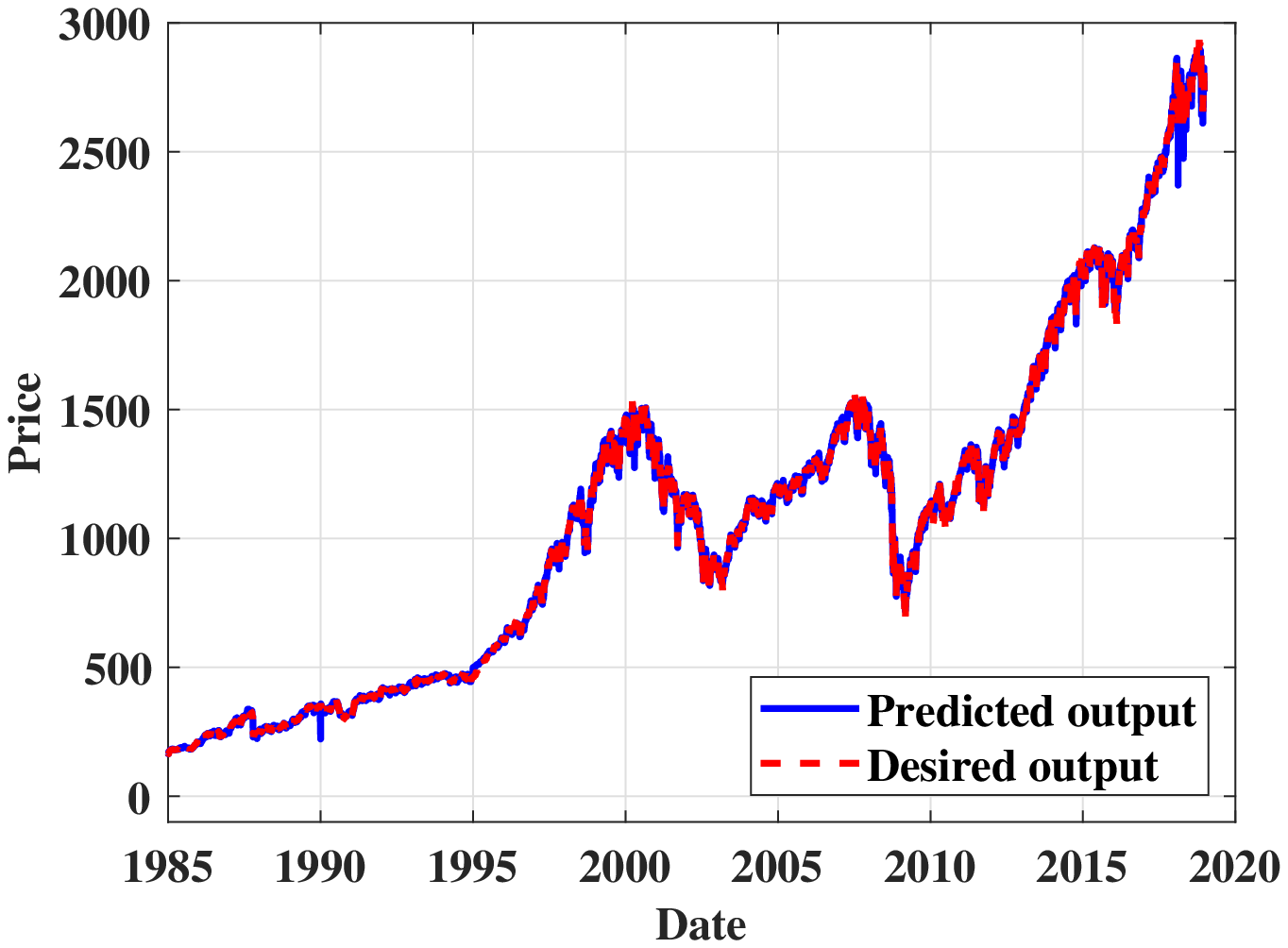}}&
    \subfigure[][]{\includegraphics[width = 2.25in]{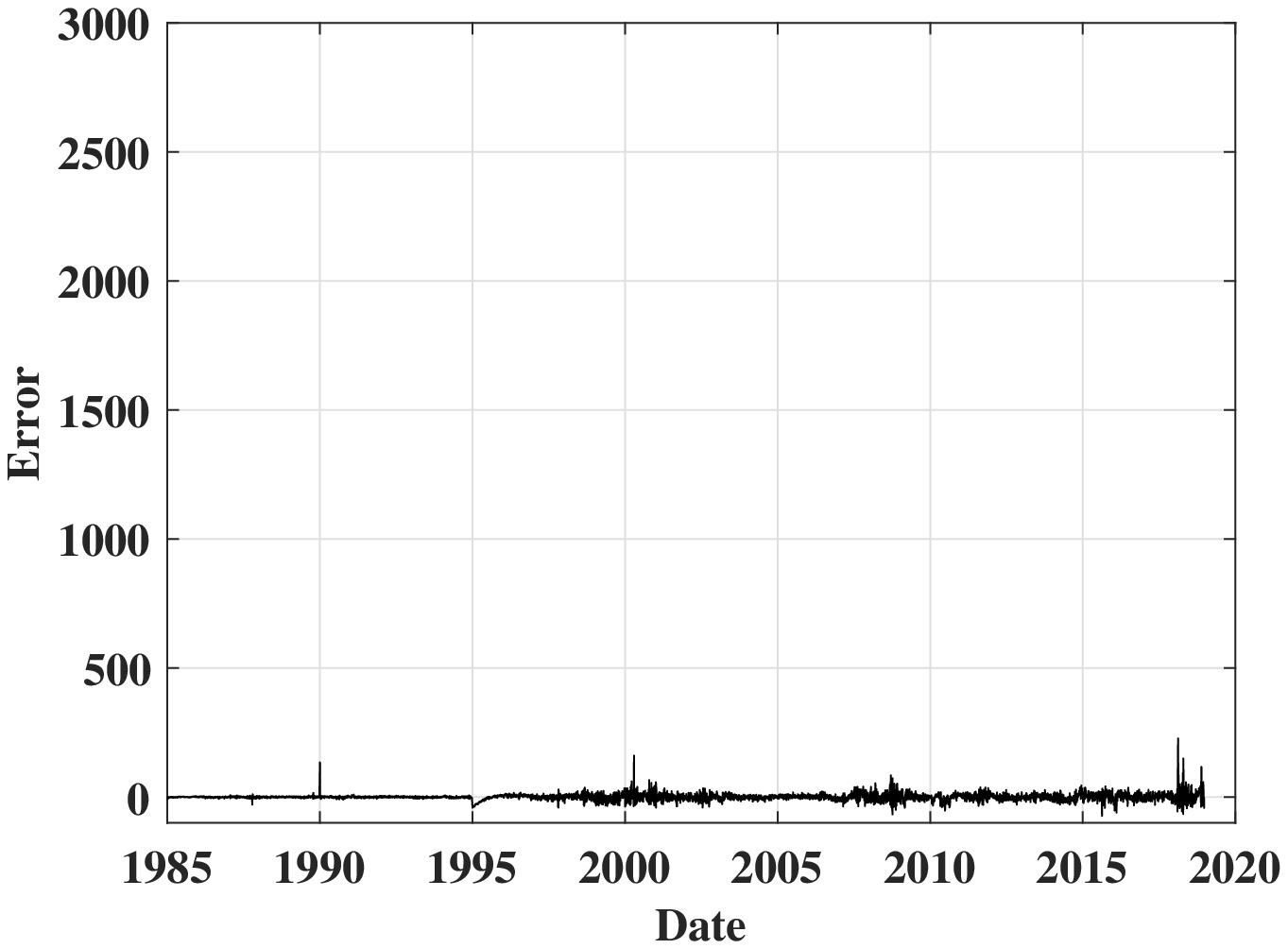}}\\
    \subfigure[][]{\includegraphics[width = 2.25in]{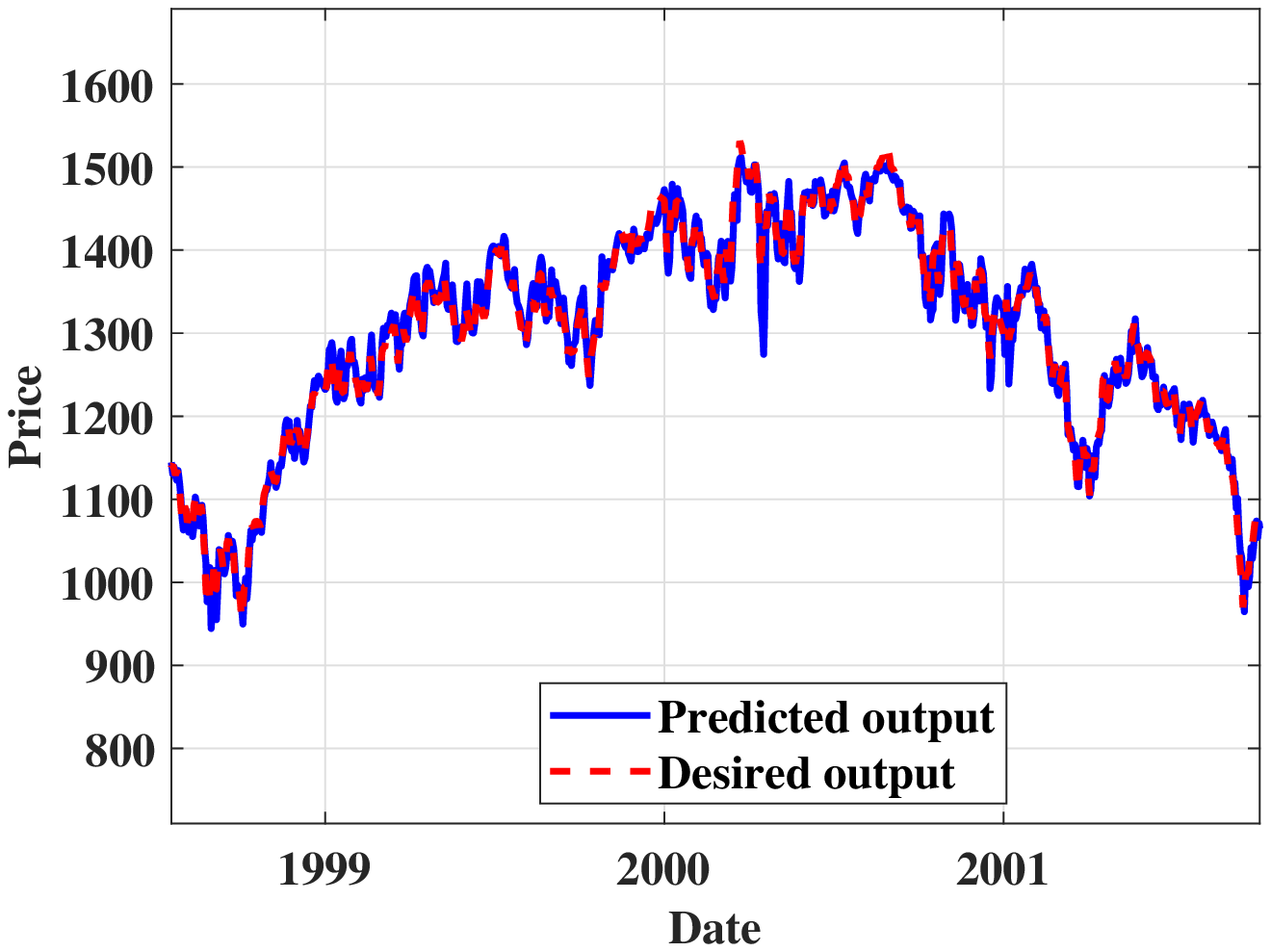}}&
    \subfigure[][]{\includegraphics[width = 2.25in]{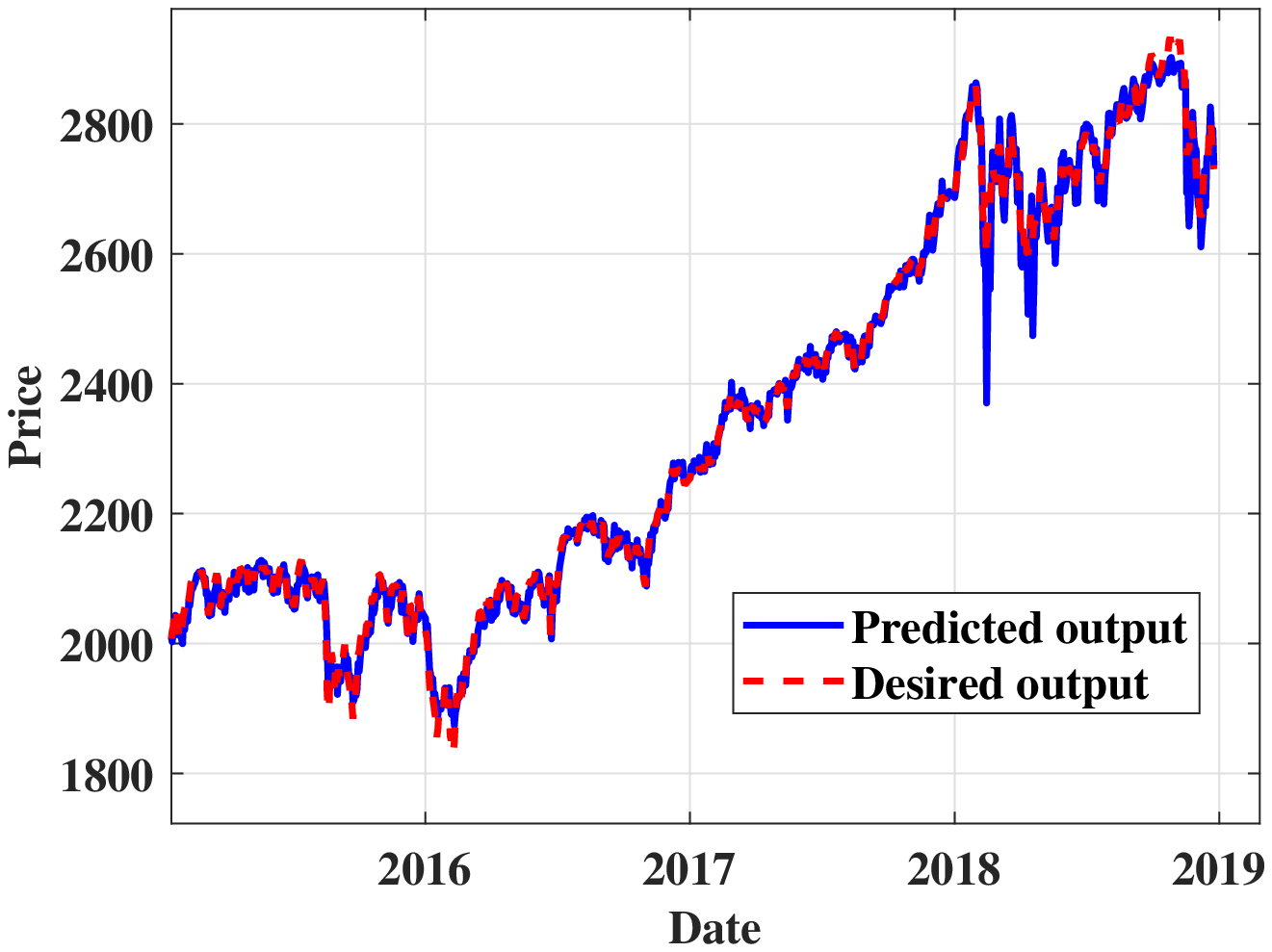}}
\end{tabular}
\caption{Comparison between actual and predicted outputs for Sydney stock price tracking (a) along with the error signal (b). To better show the performance of the network, two intervals with higher error are presented in (c) and (d).}
\label{fig_sp}
\end{figure}

A large dataset containing 8540 data samples of Sydney stock market ($S\&P$-$500$) from 3rd January 1985 to 14th November 2018 is used in this experiment \cite{ashrafi2020it2}. Data samples related to the first 5 years are selected as the training dataset and the remaining 28 years considered as the test dataset \cite{ashrafi2020it2}. The purpose of this experiment is to predict the price in the next time ($t+1$) using the current price, one, two and three steps ago modeled as: $y_{t+1}=f(y_t,y_{t-1},y_{t-2},y_{t-3})$. The initial trust region $\lambda_0$ is set 1, and for rate of changing it ($\eta$ in equation (\ref{eq37})) 1.001 is chosen. Table \ref{table_sp} compares the performance of the proposed model with previous approaches. Based on the reported results, the proposed model performs better or comparatively to the other models with the most parsimonious structure. Figure \ref{fig_sp} compares the target and predicted time-series along with the error.

\begin{table}[t]
\scriptsize
    \caption[c]{Comparison of the performance of the proposed method with previous methods in Sydney stock price tracking.}
    \centering
    \begin{tabular}{ccc}
        \hline
       \textbf{Method} & \textbf{Number of rules} & \textbf{RMSE}\\
       \hline
        IT2GSETSK \cite{ashrafi2020it2} & 5  & 13.30 \\
        GSETSK \cite{nguyen2015gsetsk} & 2  & 12.30  \\
        EFuNN \cite{EFuNN} & 5  & 217.10  \\
        DENFIS \cite{DENFIS} & 46  & 14.70  \\
        IC-FNN \cite{Ebadzadeh2017} & 2  & 34.37  \\
        \textbf{IT2CFNN} & 2 & 13.22 \\
        \hline
    \end{tabular}
    \label{table_sp}
\end{table}

\section{Conclusions}
\label{section4}
In this paper, considering the function approximation application, a new concept of uncertainty regarding the shape of fuzzy membership function is introduced. To realize this new concept, a novel form of interval type-2 fuzzy membership function is proposed. The new interval type-2 fuzzy membership function can form different shapes and its shape is uncertain. Afterward, the new form of interval type-2 fuzzy set is utilized to introduce an Interval type-2 Correlation-aware Fuzzy Neural Network (IT2CFNN). The proposed network can shape fuzzy rules similar to the covered region of the target function considering the presence of uncertainty. Next, the proposed network is learned based on Lvenberg-Marquadt (LM) optimization method and applied in some benchmark problems.

In the first experiment, the ability of the proposed model to form fuzzy rule similar to the covered region in presence of noise and uncertainty is studied. Moreover, the effect of noise level in forming footprint of uncertainty is shown.

Afterward, the ability of the proposed network for predicting well-known Mackey-Glass time-series in presence of different levels of noise is investigated. Furthermore, the performance of the network is compared with the reported performance of previous models.

Next, the performance of the model in some real-world problems including \textit{Santa-Fe Laser Time-Series Prediction}, \textit{Box-Jenkins Gas Furnace Nonlinear System Identification}, \textit{Poland  Electricity Load Forecasting}, \textit{Google Stock Price Tracking}, and \textit{Sydney Stock Price Tracking} is studied and compared to previous type-1 and type-2 models. In all problems the proposed model performs better than or comparatively with other works.

Considering the static structure of the proposed model based on Mamdani's fuzzy inference system, proposing and evolving version of the model based on TSK fuzzy inference system is proposed as the future study.

\clearpage
\bibliographystyle{plain}
\bibliography{ref}

\end{document}